\documentclass[a4paper,11pt,DIV=11,
abstract=on
]{scrartcl}
\usepackage{mathtools}
\mathtoolsset{showonlyrefs}
\usepackage{amsmath, amsthm, amssymb}
\usepackage{bm}
\usepackage{fontenc}
\usepackage[utf8]{inputenc}
\usepackage[english]{babel}
\usepackage{graphicx}
\usepackage{subcaption,xargs}
\usepackage{algorithm, booktabs}
\usepackage[noend]{algpseudocode}
\usepackage{listings}
\usepackage{hyperref}
\usepackage{environ,ifthen}
\usepackage{multirow}
\usepackage{enumerate}
\usepackage{enumitem}
\usepackage{color}
\usepackage{placeins}

\newtheorem{theorem}{Theorem}[section]

\newtheorem{proposition}[theorem]{Proposition}

\newtheorem{remark}[theorem]{Remark}
\newtheorem{example}[theorem]{Example}

\setkomafont{sectioning}{\rmfamily\bfseries}
\setkomafont{title}{\rmfamily}
\setkomafont{descriptionlabel}{\rmfamily\bfseries}

\newcommand{\E}{\mathbb{E}}
\newcommand{\R}{\mathbb{R}}
\newcommand{\N}{\mathbb{N}}
\newcommand{\Z}{\mathbb{Z}}

\renewcommand{\P}{\mathbb{P}}
\newcommand{\GG}{\mathcal{G}}

\newcommand{\SP}{\mathbb{S}}
\newcommand{\NN}{\mathcal{N}}
\newcommand{\MM}{\mathcal{M}}
\renewcommand{\AA}{\mathcal{A}}

\renewcommand{\SS}{\mathcal{S}}

\newcommand{\dx}{{\,\mathrm{d}}}
\newcommand{\e}{{\mathrm{e}}}
\newcommand{\vecOne}{\mathbf 1} 

\newcommand{\tT}{\mathrm{T}}
\DeclareMathOperator*{\argmin}{arg\,min}
\DeclareMathOperator*{\argmax}{arg\,max}

\DeclareMathOperator{\arcosh}{arcosh}

\DeclareMathOperator{\SPD}{SPD}
\DeclareMathOperator{\tr}{tr}

\DeclareMathOperator{\dist}{dist}
\DeclareMathOperator{\diag}{diag}
\DeclareMathOperator{\TV}{TV}

\DeclareMathOperator{\Log}{Log}
\DeclareMathOperator{\Exp}{Exp}

\newcommand{\abs}[1]{\left| #1 \right|}
\newcommand{\norm}[2]{\left\| #1 \right\|_{#2}}

\DeclareCaptionLabelSeparator{periodspace}{.\ }
\hypersetup{pdfauthor={Friederike Laus, Johannes Persch, Gabriele Steidl},
	pdftitle={A Nonlocal Denoising Algorithm for Manifold-Valued Images
		Using Second Order Statistics},
	pdfsubject={manuscript},%
	pdfcreator = {pdflatex and TextMate},%
	pdfkeywords={},	
	colorlinks=true,linkcolor=black,citecolor=black,urlcolor=black%
}
\begin{document}
	\title{
		A Nonlocal Denoising Algorithm for Manifold-Valued Images
		Using Second Order Statistics}
	\subtitle{Extended Version}
	\date{\today}
	
	\author{
		Friederike Laus\footnote{Department of Mathematics,
			Technische Universität Kaiserslautern,
			Paul-Ehrlich-Str.~31, 67663 Kaisers\-lautern, Germany,
			\{friederike.laus, persch, steidl\}@mathematik.uni-kl.de.
		}, 
		Mila Nikolova\footnote{CMLA -- CNRS, ENS Cachan, 61 av.
			President  Wilson, 94235 Cachan Cedex, France, nikolova@cmla.ens-cachan.fr
		},
		Johannes Persch\footnotemark[1], and Gabriele Steidl\footnotemark[1]
	}	
	\maketitle%

	\begin{abstract}
		\noindent\small
		Nonlocal patch-based methods, in particular the Bayes' approach of Lebrun, Buades and Morel~\cite{LBM13}, 
		are considered as state-of-the-art methods for denoising (color) 
		images corrupted by white Gaussian noise of moderate variance.
		This paper is the first attempt to generalize this technique to manifold-valued images.
		Such images, for example images with phase or directional entries or
		with values in the manifold of symmetric positive definite matrices, 
		are frequently encountered in real-world applications.
		Generalizing the normal law to manifolds is not canonical and different attempts have been considered.
		Here we focus on a straightforward intrinsic model and discuss the 
		relation to other approaches for specific manifolds.
		We  reinterpret the Bayesian approach of Lebrun et al.~\cite{LBM13} 
		in terms of minimum mean squared error estimation, 
		which motivates our definition of a corresponding estimator on the manifold. 
		With this estimator at hand we present a nonlocal patch-based method 
		for the restoration of manifold-valued images.
		Various proof of concept examples demonstrate the potential of the proposed algorithm.
	\end{abstract}
	
	\section{Introduction} \label{sec:intro}
	
	In many situations where measurements are taken the obtained data are corrupted by noise, 
	and typically one uses a stochastic model to describe the recorded data. 
	If there are several, independent factors that may have an influence on the data acquisition, 
	the central limit theorem suggests to model the noise as additive white Gaussian noise. 
	This is also the standard noise model one encounters in image analysis, see, e.g., \cite{GW2008}. 
	One might of course wonder whether this noise modeling is realistic and in fact, 
	in many situations the image formation process already suggests a non-Gaussian model, 
	e.g.\ Poisson noise in the case where images are obtained based on photon counting with a CCD device. 
	But also in these cases, in order to benefit from the rich knowledge and all the appealing properties of the normal distribution, 
	one often tries to transform the image in such a way that the assumption of Gaussian white noise is at least approximately fulfilled. 
	For instance, for the Poisson noise this can be achieved by the so called Anscombe transform \cite{An48}. 
	\\
	Much effort has been spent on the denoising of images corrupted with white Gaussian noise
	and a huge amount of methods have been proposed in the literature.
	Among others we  mention variational models with total variation regularizers~\cite{ROF92} and many extensions thereof, 
	denoising based on sparse representations over learned dictionaries~\cite{EA06},
	nonlocal means~\cite{GO08,S2010,YSM2012,WPCMB07} and their generalizations \cite{CM2012,DDT2009,Kerv2014,S2010}, 
	the piecewise linear estimator from Gaussian mixture models (PLE, E-PLE)~\cite{YSM2012, Wan13} 
	and SURE guided Gaussian mixture models~\cite{WM2013}, 
	patch-ordering based wavelet methods~\cite{REC2014},
	the expected patch log-likelihood (EPLL) algorithm \cite{ZW2011} 
	or better 
	its multiscale variant \cite{PE2016},
	BM3D~\cite{DFKE08} and BM3D-SAPCA~\cite{DFKE2009}, 
	and 
	the nonlocal Bayes' algorithm of Lebrun et al.~\cite{LBM13b,LBM13}.
	The latter can be viewed as an optimized reinterpretation of the two step image denoising method (TSID)~\cite{SDD2011,YEA2001}
	in a Bayesian framework.
	For a recent review of the denoising problem and the different denoising principles 
	we refer to~\cite{LCBM2012}. 
	Currently, nonlocal patch-based methods 
	achieve the best results and the quality of the denoised images 
	has become excellent for moderate noise levels.
	Even more, based on experiments with a set of 20.000 images containing about $10^{10}$ patches 
	the authors of~\cite{LN11} conjecture that for natural images, 
	the recent patch-based denoising methods might already be close to optimality. 
	Their conjecture points in the same direction as the paper of Chatterjee et al.~\cite{CM10}, who raised the question \glqq Is denoising dead?\grqq.\\
	The situation described above completely changes when dealing with manifold-valued images instead of real-valued ones, 
	a situation which is frequently encountered in applications. 
	For instance, images with values on the circle (periodic data) 
	appear in interferometric synthetic aperture radar~\cite{BKAE08,DDT11}, 
	in applications involving the phase of Fourier transformed data \cite{BLSW14},
	or when working with the hue-component of an image in the HSV color space.  
	Spherical data play a role when dealing with 3D directional information~\cite{KS2002, VO2002}  
	or 
	in the representation of a color image  in the chromaticity-brightness (CB) color space~\cite{CKS01}.    
	SO(3)-valued data appear in electron backscattered tomography \cite{BaHiSc11,BCHPS15}.
	Finally, to mention only a few examples,  
	images  with symmetric positive definite matrices as values are handled  in 
	DT-MRI imaging~\cite{chefd2004regularizing,FJ2004,VBK13,WFWBB06,WPCMB07} 
	or when covariance matrices are associated to image pixels~\cite{TPM08}. 
	Recently, some methods for the denoising of manifold-valued images have been suggested, 
	among them variational approaches using embeddings in higher dimensional spaces~\cite{rosman2012group} or 
	based on (generalized) TV-regularization~\cite{BBSW2015,BW15b,LSKC13,CS13,WDS2014}. \\
	In this paper, we aim at  generalizing the nonlocal patch-based denoising of Lebrun et al.~\cite{LBM13b,LBM13} to manifold-valued images. 
	However, for general manifolds, already the question of how to define Gaussian white noise (or, more general, a normal distribution) 
	is not canonically solved. 
	Different approaches have been proposed in the literature, 
	either by making use of characterizing properties of the real-valued normal distribution as for instance in~\cite{OC95,Pen06} 
	or by restricting to particular manifolds such as spheres, see, e.g.~\cite{MJ2000}, 
	the simplex~\cite{MPE2013}, or symmetric positive definite matrices~\cite{SBBM15}. 
	In this paper, we adopt a simple model for a normal distribution and in particular for Gaussian white noise on a manifold 
	and discuss its relationship to existing models.
	We  review the minimum mean squared error estimator in the Euclidean setting,
	which coincides with those of the Bayesian approach in~\cite{LBM13} 
	under the normal distribution assumption.
	This motivates our definition of a corresponding estimator on the manifold and gives rise 
	to a nonlocal patch-based method for the restoration of manifold-valued images.
	\\
	The outline of this paper is as follows: 
	in Section~\ref{sec:model_real} we reinterprete the nonlocal Bayes algorithm of Lebrun et al.~\cite{LBM13b,LBM13} 
	in a minimum mean square error estimation setting. This review in the Euclidean setting is  necessary to understand
	its generalization to manifold-valued images.
	In Section~\ref{sec:model_random} we introduce the notation on  manifolds. 
	Then, in Section~\ref{sec:model_manifold} we detail the nonlocal patch-based denoising algorithm for manifold valued images.
	This requires to precise what we mean by the normal law on the manifolds of interest. 
	We discuss the relation between this model and other existing ones.
	In Section~\ref{sec:numerics} we provide  several numerical examples to demonstrate that our denoising approach 
	is indeed computationally manageable.
	Examples, yet academical, for cyclic and directional data, and for images with values in the manifold of symmetric positive definite matrices 
	show the potential of nonlocal techniques for manifold-valued images. 
	Specific real-world applications are not within the scope of this paper.
	Finally, we  draw conclusions 
	and initiate further directions of research in Section \ref{sec:conclusions}.

	\section{Nonlocal Patch-Based Denoising of Real-Valued Images} \label{sec:model_real}
	%
	In this section we consider the nonlocal Bayesian image denoising method of Lebrun et al.~\cite{LBM13b,LBM13}.
	In contrast to these authors we prefer to motivate the method by a minimum mean square estimation approach.
	One reason is that the best \emph{linear} unbiased estimator in \eqref{blue_est} has a similar form as the MMSE,
	but does not rely on the assumption that the random variables are jointly normally distributed.
	This leaves potential for future work, e.g.\ when extending the model to other distributions than the normal distribution.
	
	\subsection{Minimum Mean-Square Estimator}\label{Sec:MMSE}
	
	Let $(\Omega,\AA,\P)$ be a probability space and 
	$X\colon\Omega\to  \R^n$ and $Y\colon \Omega\to \R^n$ two random vectors.
	We wish to estimate $X$ given $Y$, i.e., we seek an estimator $T\colon \R^n\to \R^n$ such that
	$\hat{X} = T(Y)$ approximates $X$. 
	A common quality measure for this task is the \emph{mean square error} 
	$\E\norm{X-T(Y)}{2}^2$, which gives rise to the definition of the \emph{minimum mean square estimator} (MMSE)
	\begin{align*}
	T_{\text{MMSE}} (Y) &= \argmin_{T}\E\norm{X-T(Y)}{2}^2\\
	& = \argmin_{Z\in \sigma(Y)} \E\norm{X-Z}{2}^2,
	\end{align*}
	where $\sigma(Y)$ denotes the $\sigma$-algebra generated by $Y$ 
	and $Z\in \sigma(Y)$ stands for all $\sigma(Y)$-measurable random variables $Z$, see, e.g.,~\cite{LC06}.
	Under weak additional regularity assumptions on the estimator $T$, 
	the Lehmann-Scheff\'e theorem~\cite{LS50,LS55}
	states that the general solution of the minimization problem is determined by  
	$$T_{\text{MMSE}}(Y)= \E(X|Y). $$ 
	In general it is not possible to give an analytical expression of the MMSE. 
	One exception constitutes of the normal distribution. 
	Recall that a random vector $X$ is normally distributed with mean $\mu \in \R^n	$ 
	and covariance matrix $\Sigma\in \R^{n \times n}$, $X\sim \NN(\mu,\Sigma)$, 
	if and only if there exists a random vector $Z\in \R^l$, whose components 
	are independent real-valued standard normally distributed  random variables 
	and a $p\times l$ matrix $A$, such that $X = AZ + \mu_X$, where  $l$ 
	is the rank of the covariance matrix $\Sigma_X = AA^\tT$. 
	If $\Sigma_X$ has full rank, then the probability density function (pdf) 
	of $X \sim \NN(\mu,\Sigma)$ with respect to the Lebesgue measure is given by
	\begin{equation} \label{gaussian_density}
	p_X(x|\mu,\Sigma) = \frac{1}{(2\pi)^{\frac{n}{2}}}\frac{1}{{\lvert\Sigma\rvert}^{\frac{1}{2}}} 
	\e^{-\frac{1}{2}(x-\mu)^{\tT}\Sigma^{-1}(x-\mu)},
	\end{equation}
	where $\lvert\Sigma\rvert$ denotes the determinant of $\Sigma$.
	In view of the next section it is useful to recall some properties of the normal distribution.
	\vspace{0.2cm}
	
	\begin{remark} {\rm (Properties of Gaussian distribution on $\mathbb R^n$)} \label{rem:prop_gaussian}
		\begin{enumerate}[label = {\upshape(\roman*)}]
			\item The Gaussian density function in \eqref{gaussian_density}  maximizes the {\rm entropy} 
			\\
			$
			H(X) \coloneqq \mathbb E \left[ -\log \left( p_X (X|\mu,\Sigma) \right) \right] 
			$
			over all density functions on $\mathbb R^n$ with fixed mean $\mu$ and covariance matrix $\Sigma$.
			\item
			Let $x_1,\ldots,x_K \in \mathbb R^n$, $K\in \N$, be i.i.d.\ 
			realizations of an absolutely continuous distribution having first and second moments, denoted by $\mu$ and $\Sigma$. 
			Then the likelihood function reads as
			$L(\mu,\Sigma|x_1,\ldots,x_K) = \prod_{k=1}^K p_X(x_k|\mu,\Sigma)$ and 
			the {\rm maximum likelihood (ML) estimator} is defined as
			$$\hat \mu \coloneqq\argmax_{\mu}  L(\mu,\Sigma |x_1,\ldots,x_K).$$
			It holds that 
			\begin{equation} \label{sample_mean}
			\hat \mu = \frac{1}{K} \sum_{k=1}^K x_k = \argmin_{x \in \mathbb R^n} \sum_{k=1}^K \lVert x-x_k\rVert_2^2
			\end{equation}
			if and only if the density function is of the form~\eqref{gaussian_density}, 
			see, e.g., {\rm \cite{DLS14,Sta93}}.
			For the normal distribution the ML estimator of the covariance matrix  reads as
			\begin{equation} \label{covar}
			\hat{\Sigma} = \frac{1}{K}\sum_{k=1}^K (x_k-\hat{\mu}){(x_k-\hat{\mu})^\tT}.
			\end{equation}
			\item The density function of the standard normal distribution $\NN(0,\sigma^2I_n)$
			with the $n \times n$ identity matrix $I_n$ is the kernel
			of the heat equation.
		\end{enumerate}
	\end{remark}
	\vspace{0.2cm}
	
	In order to compute the MMSE estimator for Gaussian random variables 
	we need to determine the conditional distribution of $X$ given $Y$. 
	It is well known that, if $X \sim \NN(\mu_X,\Sigma_X)$ and $Y\sim \NN(\mu_Y,\Sigma_Y)$ 
	are jointly normally distributed, i.e.,
	\begin{equation*}
	\begin{pmatrix}
	X\\Y
	\end{pmatrix}\sim \NN\Biggl(\begin{pmatrix} \mu_X\\ \mu_Y\end{pmatrix},\begin{pmatrix}
	\Sigma_X & \Sigma_{XY}\\
	\Sigma_{YX} & \Sigma_Y
	\end{pmatrix}\Biggr),
	\end{equation*}
	then
	the conditional distribution of $X$ given $Y=a$ is normally distributed as well and reads as
	\begin{equation*}
	(X|Y=a) \sim \NN\bigl(\mu_{X|Y},\Sigma_{X|Y}  \bigr), 
	\end{equation*}
	where
	\begin{equation*}
	\mu_{X|Y} = \mu_X + \Sigma_{XY} \Sigma^{-1}_Y (a-\mu_Y),\qquad \Sigma_{X|Y}= \Sigma_X-\Sigma_{XY}\Sigma^{-1}_Y\Sigma_{YX}.
	\end{equation*}
	As a consequence we obtain for normally distributed random vectors the MMSE estimator 
	\begin{equation} \label{mmse_est}
	T_{\mathrm{MMSE}}(Y)= \E(X|Y) = \mu_X + \Sigma_{XY} \Sigma^{-1}_Y (Y-\mu_Y) .
	\end{equation}
	In our situation (denoising) fits into the above framework if we set
	\begin{equation} \label{gaussian_setting}
	Y = X + \eta, \qquad X\sim \NN(\mu_X,\Sigma_X), \quad \eta\sim \NN(0,\sigma^2I_n),
	\end{equation}
	where we assume that $X$ and $\eta$ are independent and $\sigma^2>0$ is known.
	Then  
	$\mu_X = \mu_Y$,  and by the independence of $X$ and $\eta$ 
	further $\Sigma_{XY} = \Sigma_X$ and
	\begin{equation} \label{hidden}
	\Sigma_Y = \E\left( (X+ \eta -\mu_X) (X+\eta - \mu_X)^\tT \right) = \Sigma_X + \sigma^2 I_n.
	\end{equation} 
	Now, the MMSE of $X$ given $Y$ in \eqref{mmse_est} becomes
	\begin{align} 
	T_{\text{MMSE}}(Y) 
	&= \mu_X + \Sigma_X (\Sigma_X + \sigma^2 I_d)^{-1} (Y - \mu_X) \\
	&= \mu_Y + (\Sigma_Y - \sigma^2 I_n) \Sigma_Y^{-1}(Y-\mu_Y).\label{mmse_next}
	\end{align}
	Two remarks may be useful to see the relation to other estimators.
	\vspace{0.2cm}
	
	\begin{remark} {\rm (Relation between MMSE and BLUE)}\\
		The estimator of the general form
		\begin{equation} \label{blue_est}
		T_{\mathrm{BLUE}}(Y) = \E(X) + \Sigma_{XY} \Sigma^{-1}_Y \bigl(Y-\E(Y)\bigr)
		\end{equation}
		makes also sense for more general distributions. It is known as
		\emph{best linear unbiased estimator} {\rm (BLUE)}, 
		as it is an unbiased estimator which has minimum mean-square error  among all affine estimators.
		For jointly normally distributed $X$ and $Y$ it coincides with $T_{\text{\rm MMSE}}$.
	\end{remark}
	\vspace{0.2cm}

	\begin{remark} {\rm (Relation between MMSE and MAP)}\\
		The MMSE can also be derived in a Bayesian framework under a Gaussian prior (see, e.g.~{\rm \cite{Fes16}}), which is detailed in the following.
		Let $Y = X + \eta$, $X\sim \NN(\mu_X,\Sigma_X)$, $\eta\sim \NN(0,\sigma^2 I_d)$, where $X$ and $\eta$ are independent. 
		This implies $Y\sim \NN(\mu_X,\Sigma_X + \sigma^2 I_d)$ and $(Y|X=x)\sim \NN(x,\sigma^2 I_d)$, so that the respective densities are given by
		\begin{equation*}
		p_Y(y|X=x) = \frac{1}{(2\pi \sigma^2)^{\frac{d}{2}}} \e^{-\frac{1}{2\sigma^2}\lVert y-x\rVert_2^2}
		\end{equation*}
		and
		\begin{equation*}
		p_X(x) = \frac{1}{(2\pi)^{\frac{d}{2}}} \frac{1}{|\Sigma_X|^{\frac{1}{2}}}\e^{-\frac{1}{2}(x-\mu_X)^{\tT}\Sigma_X^{-1}(x-\mu_X)}.
		\end{equation*}
		By Bayes' formula we have
		\begin{equation*}
		p_X(x|Y=y) = \frac{p_Y(y|X=x) p_X(x)}{p_Y(y)}\propto p_Y(y|X=x) p_X(x),
		\end{equation*}
		and therewith, the maximum a posteriori (MAP) estimate reads as
		\begin{align*}
		\hat x &= \argmax_{x} \{p_X(x|Y=y)\}  = \argmax_{x} \{ p_Y(Y|X=x) p_X(x)\}\\
		& =  \argmax_{x} \big\{ \log(p_Y(Y|X=x)) + \log( p_X(x)) \big\}\\
		& =  \argmin_{x} \left\{ \frac{1}{2\sigma^2}\lVert x-y\rVert_2^2 + \frac{1}{2}(x-\mu_X)^{\tT}\Sigma_X^{-1}(x-\mu_X) \right\}.
		\end{align*}
		Setting the gradient to zero results in
		\begin{align*}
		(I_d  + \sigma^2 \Sigma_X^{-1}) \hat x &= \sigma^2 \Sigma_X^{-1} \mu_X + y.		 
		\end{align*}
		Observing that 
		$I_d +  \sigma^2 \Sigma_X^{-1} = \Sigma_X^{-1}(\Sigma_X + \sigma^2 I_d)$ 
		and
		$ \sigma^2 (\Sigma_X + \sigma^2 I_d)^{-1}=I_d - \Sigma_X (\Sigma_X + \sigma^2 I_d)^{-1} $,
		we obtain finally
		\begin{align*}
		\hat{x} & = \sigma^2 (\Sigma_X + \sigma^2 I_d)^{-1}\mu_X +(\Sigma_X + \sigma^2 I_d)^{-1} \Sigma_X y \\
		& = \mu_X + \Sigma_X(\Sigma_X + \sigma^2 I_d)^{-1} (y-\mu_X).
		\end{align*}	
	\end{remark}
	\vspace{0.2cm}
	
	In practical applications, the parameters $\mu_Y$ and $\Sigma_Y$ 
	are unknown and need to be estimated using realizations (observations) 
	$y_1,\ldots,y_K$ of $Y$. 
	Here we use the ML estimators given in \eqref{sample_mean} and \eqref{covar}.
	Note that the ML estimator for the covariance matrix is slightly biased.
	Instead we could also deal with an unbiased estimator by replacing the averaging factor by 
	$\frac{1}{K-1}$. However, the numerical difference is  negligible for large $K$.
	
	Summarizing our findings, we obtain the following empirical estimator
	\begin{equation} \label{wichtig}
	\hat T_{\text{MMSE}}(y) = \hat \mu_Y +( \hat \Sigma_Y - \sigma^2 I_n ) \hat \Sigma_Y^{-1} (y- \hat \mu_Y).
	\end{equation}
	
	\begin{remark}[Positive definiteness of the empirical covariance matrix] \label{neg_cov}
				Equation \eqref{wichtig} contains via \eqref{hidden} the hidden assumption that
		$
		\hat{\Sigma}_X = \hat{\Sigma}_Y - \sigma^2 I_n.
		$
		However, based on the empirical covariance  matrix $\hat{\Sigma}_Y$ it is not necessarily ensured that 
		$\hat{\Sigma}_Y - \sigma^2 I_n$ is positive semi-definite and thus  a valid covariance matrix.
		There are different ways to overcome this problem, e.g.\ replacing negative eigenvalues by a small positive value as for instance proposed in~\cite{RJ11}, 
		compare also the discussion in~\cite[Section 3.5]{LBM13b} or~\cite[page 406]{Yar12}. In our numerical experiments we did not observe that this issue had negative impacts on the results.
	\end{remark}
	
	\subsection{Denoising Using the MMSE Approach}\label{subsec:denoise_real}
	Next we describe how the results of the previous section can be used for image denoising. 
	To this aim, let  ${x\colon \GG\to \R}$ be a discrete gray-value image, 
	defined on a grid $\GG = \{1,\ldots,N_1\}\times \{1,\ldots,N_2\}$. 
	By a slight abuse of notation we  also write $x\in \R^{N}$, where $N = N_1N_2$
	for the  columnwise reshaped version of the image. 
	It will be always clear from the context to which notation we refer. 
	We assume that the image is corrupted with white Gaussian noise, i.e.,
	\begin{equation*}
	y = x  + \eta, 
	\end{equation*}
	where
	$\eta$ is now a realization of $ \NN(0,\sigma^2 I_{N})$. Based on $y$ we wish to reconstruct the original image $x$. \\
	We use  the fact that natural images are to some extend self-similar, 
	i.e., small similar patches may be found several times in the image, and that for these patches locally
	a normality assumption holds approximately true, see, e.g.~\cite{ZW2011}.
	To formalize this idea,  consider 
	an $s \times s$
	neighborhood (patch) $y_i$ centered at $i = (i_1,i_2)\in \mathcal{G}$, where  $s = 2\kappa+1$,  $\kappa\in\N$.
	After vectorization this corresponds  to a realization of an \mbox{$n$-dimensional} 
	normally distributed random vector $Y_i\sim \NN(\mu_i,\Sigma_i)$, where 
	$n = s^2$.
	This patch is referred to as a reference patch in the following.
	Similar patches are interpreted as other realizations of 
	$\NN(\mu_i,\Sigma_i)$.  
	There are several strategies to define similar patches. 
	Take for example, for a fixed $K\in \N$, 
	the $K$ nearest patches with respect to the Euclidean distance 
	in a  $w \times w$ search window around $i$, where
	$w = 2\nu + 1 \gg s$, $\nu \in\N$. 
	Let $\SS(i)$ denote the set of centers of patches similar to $y_i$. 
	Then the estimates of the expectation value \eqref{sample_mean} and the covariance \eqref{covar} become
	\begin{equation*}
	\hat{\mu}_i = \frac{1}{K} \sum_{k\in\SS(i)} y_k
	\qquad {\rm and} \qquad
	\hat{\Sigma}_i = \frac{1}{K}\sum_{k\in \SS(i)} (y_k-\hat{\mu}_i)(y_k-\hat{\mu}_i)^\tT.
	\end{equation*}
	The obtained estimates are then used to restore the reference patch and all its similar patches with \eqref{wichtig} as: 
	\begin{equation} \label{3a}
	\hat{y}_j = \hat \mu_i + (\hat \Sigma_i-\sigma^2 I_n) \hat \Sigma_i^{-1}(y_j - \hat \mu_i),\qquad j\in\SS(i). 
	\end{equation}
	Proceeding as above for all pixels $i\in {\mathcal G}$ yields  a variable number of estimates for each pixel. 
	Therewith, the final estimate at pixel $i$ is obtained as an average over all patches containing the pixel $i$ (aggregation). 
	There  are some fine-tuning steps that were partly 
	also considered in~\cite{LBM13b,LBM13}. This is summarized in the following remark.
	\vspace{0.2cm}
	
	\begin{remark}  {\rm (Fine tuning steps)} \label{details}
		\begin{enumerate}[label = {\upshape(\roman*)}]
			\item \emph{Boundaries}:
			Special attention has to be paid to patches at the boundaries of an image. 
			There are at least two possibilities: Either, one extends the image, e.g.\ by mirroring, or one considers only patches 
			lying completely inside the image together with appropriately smaller search windows. To our opinion the second strategy is preferable since it does not introduce 
			artificial information. 
			However, it leads to less estimates at the boundaries of the image, but we observed that this does not yield visible artifacts in practice. 
			\item \emph{Flat areas}:
			Flat areas, where differences between  patches are only caused by noise, require a special consideration, as it 
			is very likely that the estimated covariance matrix will not have full rank. In this case, the patches are better denoised by only using their mean. 
			Flat areas might be detected using the empirical variance of the patches, which is close to $\sigma^2$. 
			\item \emph{Second step}:
			The similarity of patches and the covariance structure of the patches can be better estimated using the first step denoised image as an oracle image
			for a second step.
			\item \emph{Acceleration}: To speed up the denoising procedure, each patch that has been used (and therefore denoised
			at least once) in a group of similar patches is not considered as reference patch anymore. 
			Nevertheless, it may be denoised several times by being potentially chosen in other groups.
		\end{enumerate}
	\end{remark}
	\vspace{0.2cm}
	
	The whole denoising procedure is given in Algorithm~\ref{Alg:NL_real}.
	We would like to point out the differences between the two steps, which look at first glance very similar: Step 2 uses the  
	denoised image from Step 1 in order to find similar patches and to estimate the covariance matrix, but reuses the original noisy image for the other computations, i.e, for the mean patch and the restored image.
	\\
	
	\begin{algorithm}[t]
		{
			\caption{Nonlocal MMSE Denoising Algorithm on $\R^d$, Step 1}
			\label{Alg:NL_real}
			\begin{algorithmic}
				\State \textbf{Input:}
				noisy image $y\in \R^{N,d}$, variance $\sigma^2$ of the noise
				\State \textbf{Output:}	first step denoised image $\hat{y}$ and final image $\tilde{y}$
				\State \textbf{Parameters:} $s_1,s_2$ sizes of patches, 
				$K_1,K_2$ numbers of similar patches, $\gamma$ homogeneous area parameter, $w_1,w_2$ sizes of search areas
				\State \textbf{Step 1:}
				\For{all patches $y_i \in \R^{s_1^2,d}$ of the noisy image $y$ not considered before}
				\State
				Determine the set $\SS_1(i)$ of centers of $K_1$ patches similar to $y_i$ in a $w_1\times w_1$ window around $i$
				\State	Compute the empirical mean patch, $\hat{\mu}_i = (\hat{\mu}_{i,j})_{j=1}^{s_1^2}$, 
				\vspace{-0.5\baselineskip}
				\begin{equation*}
				\hat{\mu}_i =\frac{1}{K_1} \sum_{k\in\SS_1(i)} y_k 
				\end{equation*}
				\vspace{-0.5\baselineskip}
				\State \textbf{Homogeneous area test:} Compute the 
				mean value  
				$\hat{m}_i =\frac{1}{s_1^2} \sum_{j=1}^{s_1^2} \hat{\mu}_{i,j}$
				\State and the empirical variance of the patches
				\begin{equation*}
				\hat{\sigma}^2_i =\frac{1}{d K_1 s_1^2} \sum_{k\in\SS_1(i)} \bigl(y_k- \vecOne_{s_1^2}
				\otimes\hat{m}_i\bigr)^\tT \bigl(y_k- \vecOne_{s_1^2}\otimes\hat{m}_i\bigr)
				\end{equation*}
				\If{$\hat{\sigma}^2_i \leq \gamma \sigma^2$} 
				\State Compute the restored patches  $\hat{y}_k = 	\vecOne_{s_1^2}\otimes\hat{m}_i$, $k\in \SS_1(i)$
				\Else
				\State Compute the empirical covariance matrix\vspace{-0.5\baselineskip}
				\begin{equation*}
				\hat{\Sigma}_i =\frac{1}{K_1}\sum_{k\in \SS_1(i)} (y_k -\hat{\mu}_i)(y_k -\hat{\mu}_i)^\tT  
				\end{equation*}\vspace{-0.5\baselineskip}
				\State Compute the restored patches  
				$\hat{y}_j = \hat{\mu}_i+ (\hat{\Sigma}_i - \sigma^2 I_{s_1^2}) \hat{\Sigma}_i^{-1} (y_j-	\hat{\mu}_i)$, $j\in \SS_1(i)$
				\EndIf
				\State	\textbf{Aggregation:} Obtain the first estimate $\hat{y}$ at each pixel by computing the average over all restored patches containing the pixel	
				\EndFor		
				\algstore{realalg}
			\end{algorithmic} 	}
		\end{algorithm}

		\begin{algorithm}[p]
			\ContinuedFloat
			\caption{Nonlocal MMSE Denoising Algorithm on $\R^d$, Step 2}
			\begin{algorithmic}
				\algrestore{realalg}
				\State \textbf{Step 2:} 
				\For{all patches $y_i$ of the noisy image $y$ not considered before}
				\State
				Determine in a $w_2\times w_2$ window around $i$
				the set ${\SS_2}(i)$ of centers of $K_2$ patches which are similar 
				to patch~$\hat{y}_i$ of the denoised image in Step 1.  
				\State	Compute the empirical mean patch, $\hat{\mu}_i = (\hat{\mu}_{i,j})_{j=1}^{s_1^2}$,  \vspace{-0.5\baselineskip}
				\begin{equation*}
				\tilde{\mu}_i =\frac{1}{K_2} \sum_{k\in\SS_2(i)} y_k 
				\end{equation*}\vspace{-0.5\baselineskip}
				\State  \textbf{Homogeneous area test:} Compute the 
				mean value by
				$\tilde{m}_i =\frac{1}{s_2^2} \sum_{j=1}^{s_2^2} \tilde{\mu}_{i,j}$
				\State 	and the empirical variance of the patches
				\State \begin{equation*}
				\tilde{\sigma}^2_i =\frac{1}{d K_2 s_2^2} \sum_{k\in\SS_2(i)} \bigl(y_k- \vecOne_{s_2^2}\otimes\tilde{m}_i\bigr)^\tT \bigl(y_k- \vecOne_{s_2^2}\otimes\tilde{m}_i\bigr)
				\end{equation*}
				\If{$\tilde{\sigma}^2_i \leq \gamma \sigma^2$} 
				\State Compute the restored patches $\tilde{y}_j = \vecOne_{s_2^2}	\otimes\tilde{m}_i$, $j\in \SS_2(i)$
				\Else
				\State Compute the empirical covariance matrix\vspace{-0.5\baselineskip}
				\begin{equation*}
				\widetilde{\Sigma}_i =\frac{1}{K_2}\sum_{k\in \SS_2(i)}(\hat{y}_k -\tilde{\mu}_i)(\hat{y}_k -\tilde{\mu}_i)^\tT  + \sigma^2 I_{s_2^2}
				\end{equation*}\vspace{-0.5\baselineskip}
				\State 
				Compute the restored patches 
				$\tilde{y}_j = \tilde{\mu}_i+(\widetilde{\Sigma}_i- \sigma^2 I_{s_2^2})\widetilde{\Sigma}_i^{-1} (y_j-	\tilde{\mu}_i)$, $j\in \SS_2(i)$
				\EndIf
				\State	\textbf{Aggregation:} 
				Obtain the final estimate $\tilde{y}$ at each pixel by computing the average over all restored patches containing the pixel	
				\EndFor	
			\end{algorithmic} 	
		\end{algorithm}
		\FloatBarrier
		
		The overall approach can be generalized to images with values in $\R^d$, $d > 1$, in a straightforward way,
		dealing now with $n$-dimensional random vectors, where 
		$n = s^2 d$. In particular, RGB-color images ($d=3$) can be denoised in this way. At this point, considering  the three color channels independently does usually not 
		yield good results as there is a significant correlation between the red, the green, 
		and the blue color channel. This correlation is correctly taken into account in the three-variate setting. 
		As an alternative, Lebrun et al.~{\rm \cite{LBM13}} suggested to work in the so-called
		$Y_o U_o V_o$ color space~{\rm \cite{OKS1980}},
		which is a variant of the YUV space where  transform from the RGB space is orthogonal and thus does not change the noise statistics. 
		This color system separates geometric from chromatic information and thereby decorrelates the color channels, so that
		treating them independently  does not create noticeable color artifacts as it would be the case in the RGB space.	
		
		\section{Random Points on  Manifolds} \label{sec:model_random}
		%
		Instead of $\mathbb R^d$-valued images we are  now interested in images 
		having values in a $d$-dimensional manifold $M$.
		We start by introducing the necessary notation in Riemannian manifolds.
		In our numerical examples we will deal with images having components
		on the $d$-sphere $\mathbb S^d$ equipped with the Euclidean metric
		of the embedding spaces $\R^{d+1}$, $d=1,2$, 
		and the manifold of positive definite $r \times r$-matrices ${\SPD}(r)$, $r=2,3$, 
		with the affine invariant metric.
		For these manifolds the specific expressions of the following quantities are given in Appendix \ref{app:ex_mani}.
		Further, we will consider the open probability simplex
		$\Delta_{d} \subset \R_{>0}^{d+1}$, $d=1$,
		with the Rao-Fisher metric obtained from the categorial distribution 
		and 
		the hyperbolic manifold  $\mathbb H^d$, $d=2$, equipped with the Minkowski metric.
		Besides many textbooks on differential geometry the reader may have a look into Pennec's paper
		\cite{Pen06} to get an overview. We adapted our notation to this paper.

		\paragraph{Manifolds} 
		If not stated otherwise, let $\MM$ be a complete, connected $n$-dimensional Riemannian manifold. 
		All of the previously mentioned manifolds are complete, except for the probability simplex.
		Observe that we will work with $s \times s$ patches of $d$-dimensional manifolds $M$
		such that we finally deal with product manifolds $\MM = M^{s^2}$
		of dimension $n = s^2 d$ with the usual product metric.
		By $T_{\bm x} \MM$ we denote the tangent space of $\MM$ at $\bm x \in \MM$ 
		and by
		$\langle \cdot,\cdot \rangle_{\bm x} \colon T_{\bm x} \MM \times T_{\bm x} \MM \rightarrow \mathbb R$ 
		the Riemannian metric.
		Let 
		\(\gamma_{{\bm x},v}(t)\), \({\bm x} \in\mathcal M\), \(v\in T_{\bm x} \MM\),
		be the geodesic starting from 
		\(\gamma_{{\bm x},v}(0) = {\bm x}\) with
		\(\dot\gamma_{{\bm x},v} (0) = v\). 
		Since $\MM$ is complete, 
		the exponential map \(\exp_{\bm x}\colon T_{\bm x} \MM \rightarrow \mathcal M\) with
		$$
		\exp_{\bm x}(v) \coloneqq \gamma_{\bm x,v}(1)
		$$
		is well-defined for every $\bm x \in \MM$. 
		The exponential map realizes a local diffeomorphism (exponential chart) 
		from a ``sufficiently small neighborhood'' of the origin $0_{\bm x}$ of $T_{\bm x}\MM$ 
		into a neighborhood of ${\bm x} \in \MM$.
		To precise how large this ``small neighborhood'' can be chosen,
		we follow the geodesic $\gamma_{{\bm x},v}$ from $t=0$ to infinity. 
		It is either minimizing all along or up to a finite time $t_0$ and not any longer afterwards.
		In the latter case, $\gamma_{{\bm x},v}(t_0)$ is called \emph{cut point} 
		and the corresponding tangent vector $t_0v$ is called \emph{tangential cut point}.
		The set of all cut points of all geodesics starting from $\bm x$ 
		is the \emph{cut locus} $\mathcal{C}({\bm x})$ and the set of corresponding vectors $\mathcal{C}_T(0_{\bm x})$
		the \emph{tangential cut locus}.
		Then the open domain $\mathcal{D}_T(0_{\bm x})$ around $0_{\bm x}$ bounded by the tangential cut locus is
		the maximal domain for which the exponential chart at $\bm x$ is injective.
		It is connected and star-shaped with respect to $0_{\bm x}$
		and 
		$ \exp_{\bm x} \mathcal{D}_T (0_{\bm x}) = \MM \backslash {\cal C} (\bm \mu) $.
		This allows to define the inverse exponential map as
		$$
		\log_{\bm x} \coloneqq \exp_{\bm x}^{-1}\colon \MM \backslash {\cal C} (\bm \mu) \to T_{\bm x}\mathcal M.
		$$
		For the $d$-sphere $\mathbb S^d$ the cut locus of ${\bm x}$ is just its antipodal point $-\bm x$. 
		Thus the tangential cut locus $\mathcal{D}_T(0_{\bm x})$ is the ball  with radius $\pi$ around $0_{\bm x}$
		and  $\mathcal{C}_T(0_{\bm x})$ its boundary.
		For Hadamard manifolds which are complete, simply-connected manifolds 
		with  non-positive sectional curvature~\cite{B2014},
		as ${\SPD}(r)$ or $\mathbb H^d$, 
		we have that 
		$\mathcal{D}_T(0_{\bm x}) = T_{\bm \mu} \MM$.
		
		The Riemannian metric yields a distance function $\dist_{\MM}\colon \MM \times \MM \rightarrow \mathbb R_{\ge 0}$
		on the manifold by
		$ 
		\dist_{\cal M}(\bm x,\bm y) = \langle \log_{\bm x} (\bm y), \log_{\bm x} (\bm y) \rangle_{\bm x}
		$ 
		and a measure ${\rm d}_{\MM} (\bm x)$ written 
		in local coordinates $x = (x^1,\ldots,x^n)$ by
		${\rm d}_{\MM} (\bm x)  = \sqrt{ G(x) } \, \dx x$, where
		$
		G(x) \coloneqq 
		\big( \big\langle \frac{\partial}{\partial x ^i}, \frac{\partial}{\partial x^j} \big\rangle_{\bm x} 
		\big)_{i,j = 1}^n 
		$
		and $\dx x \coloneqq \dx x^1 \ldots \dx x^n$.
		\\
		
		\paragraph{Random Points}
		Let $(\Omega, \AA, \mathbb P)$ be a probability space and 
		${\cal B} (\MM)$  the Borel $\sigma$-algebra on $\MM$ (with respect to $\dist_{\MM}$).
		A measurable map $\bm X\colon \Omega \rightarrow \MM$
		is called a \emph{random point} on $\MM$.
		We consider only absolutely continuous random points $\bm X$ with probability density $p_{\bm X}$, i.e.,
		$\mathbb P(\bm X \in B) = \int_B p_{\bm X} ({\bm x}) \, {\rm d}_{\MM}({\bm x})$ for all $B \in {\cal B} (\MM)$ 
		and
		$\mathbb P (\bm X \in \MM) = 1$.
		The \emph{variance} of  $\bm X$ with respect to a given point $\bm y$ is defined as
		\begin{equation} \label{variance}
		\sigma^2_{\bm X} (\bm y) \coloneqq \E\bigl( \dist_{\MM}(\bm X,\bm y)^2\bigr) 
		= \int_{\MM} \dist_{\MM} (\bm x, \bm y)^2 \, p_{\bm X}(\bm x) \dx_{\MM} (\bm x),
		\end{equation}
		and  local minimizers of $\bm y \mapsto \sigma^2_{\bm X} (\bm y)$  are called   \emph{Riemannian centers of mass}~\cite{Karcher1977}.
		For a discussion of the existence and uniqueness of global minimizers, 
		known as \emph{Fr{\'e}chet expectation} or \emph{means} $\E(\bm X)$ of $\bm X$ see, 
		e.g., \cite{AR11,Karcher1977,Kendall1990}.
		For Hadamard manifolds with curvature bounded from below 
		the Riemannian center of mass exists and is unique.
		For the spheres ${\mathbb S}^d$, if
		the support of $p_{\bm X}$ is contained in a geodesic ball of radius $r < \pi/2$, then
		the Riemannian center of mass is unique within this ball and it is the global minimizer of \eqref{variance}.
		In the following we assume that the variance is finite and the cut locus has a probability measure zero 
		at any point $\bm y \in \MM$. Then a necessary condition for $\bm \mu$ to be a Riemannian center of mass is
		\begin{equation} \label{karcher_mean}
		\int_{\MM} \, \log_{\bm \mu} (\bm x) \,  \dx_{\MM} (\bm x) = 0.
		\end{equation}
		For Hadamard manifolds with curvature bounded from below this condition is also sufficient.
		Assuming that the mean $\bm \mu = \E(\bm X)$ is known, 
		we define the covariance matrix $\Sigma$ of $\bm X$ (with respect to $\bm \mu$) by\\[-\baselineskip]
		\begin{equation} \label{cov}
		\Sigma = \E \bigl(\log_{\bm \mu} (\bm X) \log_{\bm \mu} (\bm X)^\tT \bigr) 
		= 
		\int_{\MM} \log_{\bm \mu} (\bm x) \log_{\bm \mu} (\bm x)^\tT \, p_{\bm X}(\bm x) \, \dx_{\MM} (\bm x).
		\end{equation}
		In practice, typically $\E(\bm X)$ and $\Sigma$ are unknown and need to be estimated. 
		Given observations $\bm x_1,\ldots, \bm x_K \in \MM$ of a random point $\bm X$, we estimate the mean point by
		\begin{equation} \label{karcher_disc}
		\hat{\bm \mu} \in \argmin_{\bm x \in \MM} \frac{1}{K} \sum_{k=1}^K \dist_{\MM}(\bm x,\bm x_k)^2,
		\end{equation}
		which is according to~\cite{BP03} a consistent estimator of $\mathbb E(\bm X)$ and can be computed by a gradient descent algorithm, see, e.g.~\cite{ATV13}. 
		An estimator for the covariance matrix reads as
		\begin{equation} \label{cov_empir}
		\hat{\Sigma} =\frac{1}{K} \sum_{k=1}^K \log_{\hat {\bm \mu}} (\bm x_k) \log_{\hat {\bm \mu}}(\bm x_k)^\tT.
		\end{equation}
		%
		\section{Nonlocal Patch-Based Denoising of Manifold-Valued Images} \label{sec:model_manifold}
		%
		In this section we propose an NL-MMSE denoising algorithm for manifold-valued images.
		To this end, we have to specify what we mean by ``normally  distributed'' random points on manifolds. 
		In contrast to the vector space setting, there does not exist a canonical definition 
		of a normally distributed random vector on a manifold 
		since various properties characterizing the normal distribution 
		on $\R^n$ as those in Remark \ref{rem:prop_gaussian},
		cannot be generalized to the manifold setting in a straightforward way. 
		Here we rely on a simple approach which transfers normally distributed zero mean random vectors 
		on tangent spaces via the exponential map to the manifold.
		Based on this definition we will see how the NL-MMSE from Section~\ref{sec:model_real} 
		carries over to manifold-valued images. 
		\\
		
		\subsection{Gaussian Random Points}\label{gaussian_random_points}
		In the following, we describe the Gaussian model used in this paper 
		for Hadamard manifolds and spheres and
		discuss its relation to other models for small variances.
		
		For each tangent space $T_{\bm x} \MM$ with fixed orthonormal basis $\{e_{{\bm x},i}\}_{i=1}^{n}$ 
		we can identify the element
		$
		\sum_{i=1}^n x^i e_{{\bm x},i} \in T_{\bm x}\MM
		$
		with the local coordinate vector $x = (x^i)_{i=1}^n \in \R^n$, 
		which establishes an isometry between  $T_{\bm x} \MM$ and $\R^n$.  
		Note that  also the expressions in~\eqref{cov} and~\eqref{cov_empir} are basis dependent,
		but assuming a fixed basis the relation skipped for simplicity of notation.
		Now, let $\bm \mu \in \MM$ and let $h\colon \mathbb R^n \rightarrow T_{\bm \mu} \MM$
		be the  linear isometric mapping 
		\begin{equation} \label{def_h}
		h(x) \coloneqq \sum_{i=1}^n x^i e_{\bm \mu,i} \in T_{\bm \mu}\MM.
		\end{equation}
		Let ${\cal D}_{\bm \mu} \coloneqq h^{-1} \left( {\cal D}_T ( 0_{\bm \mu} ) \right)  \subseteq \R^n$. 
		Since $\exp_{\bm \mu}$ is continuous, we have for any 
		$B \in {\cal B} ( \MM )$ 
		that 
		$B_n \coloneqq h^{-1} (\log_{\bm \mu} (B)) \subseteq {\cal D}_{\bm \mu}$ is a Borel set 
		and for any integrable function $F$ it holds
		\begin{align}\label{rueck}
		\int_{B} F(\bm x) \dx_{\MM} (\bm x )
		&= 
		\int_{B_n} 
		F \left( \exp_{\bm \mu} ( h (x))  \right) \,
		\big| G(x) \big|^\frac12 \, \dx x,
		\end{align}
		where
		$
		G(x) =
		\big(
		\langle 
		{\rm d} (\exp_{\bm \mu})_{h(x)} [e_{\bm \mu,i}],  {\rm d} (\exp_{\bm \mu})_{h(x)} [e_{\bm \mu,j}] 
		\rangle
		\big)_{i,j=1}^n
		$. 
		Conversely, for any Borel set $B_n \subseteq {\cal D}_{\bm \mu}$ and any integrable function $f$
		we see that $B \coloneqq \exp_{\bm \mu}(h(B_n)) \in {\cal B}(\MM)$ 
		and
		\begin{align} \label{transform}
		\int_{B_n} f(x) \dx x
		&= 
		\int_B f \big( h^{-1} (\log_{\bm \mu}\bm (x)) \big) \,
		\big| \tilde G(\bm x) \big|^\frac12 
		\dx_{\MM} \, (\bm x),
		\end{align}
		where
		$
		\tilde G(\bm x) =
		\big(
		\langle {\rm d} (\log_{\bm \mu})_{\bm x} [e_{\bm x,i}], {\rm d} (\log_{\bm \mu})_{\bm x} [e_{\bm x,j}] \rangle_{\bm \mu}
		\big)_{i,j=1}^n.
		$
		If $Z \sim {\cal N}(0,I_n)$ is standard normally distributed on $\R^n$ with pdf $p_Z$, then
		\begin{equation} \label{definition}
		\bm Z \coloneqq \exp_{\bm \mu} (h(Z))
		\end{equation} 
		is a random point on $\MM$.
		For Hadamard manifolds, we have $D_{\bm \mu} = \R^n$ 
		so that for $\bm x \coloneqq \exp_{\bm \mu}(h(x))$, 
		\begin{equation} \label{radial}
		\|x\|_2^2 = \langle h(x), h(x) \rangle_{\bm \mu} 
		=
		\langle \log_{\bm \mu} (\bm x), \log_{\bm \mu} (\bm x) \rangle_{\bm \mu}
		= {\rm dist}_{\MM} (\bm \mu,\bm x)^2.
		\end{equation}
		Thus, $\bm Z$
		has the pdf
		\begin{equation} \label{stand_normal}
		\begin{split}
		p_{\bm Z} (\bm z) 
		&= 
		p_Z (h^{-1} \left(\log_{\bm \mu} (\bm z)) \right) \lvert\tilde G (\bm z)\rvert^{\frac{1}{2}}\\
		&=
		\frac{1}{(2 \pi)^{n/2}} {\rm e}^{-\frac12 {\rm dist}_{\MM} (\mu, \bm z)^2 }
		\lvert\tilde G (\bm z)\rvert^{\frac{1}{2}}. 
		\end{split}
		\end{equation}  
		Note that by incorporating the factor $\lvert\tilde G (\bm z)\rvert^{\frac{1}{2}}$ 
		into the density function we avoid problems as discussed in~\cite{Jer05}.
		By construction and \eqref{karcher_mean} it follows directly that the mean of $\bm Z$ is $\bm \mu$
		and the covariance \eqref{cov}  is $I_n$. 
		We consider $\bm Z$ as normally distributed on $\MM$ and write
		$\bm Z \sim \NN_{\MM}(\bm \mu,I_n)$.
		In other words, $\bm Z$ is  normally distributed on $\MM$ with mean $\bm \mu$ and covariance $I_n$
		if $Z \coloneqq h^{-1} (\log_{\bm \mu} (\bm Z))$ is standard normally distributed on $\mathbb R^n$.
		
		If $D_{\bm \mu} \not = \R^n$ as it is the case for $d$-spheres, 
		we  assume that up to a set of Lebesgue measure zero 
		$\R^n =  {\dot \bigcup}_{j \in \mathcal{J}} {\cal D}_{\bm \mu,j}$,
		where $\mathcal{J} \subseteq \mathbb Z$ is an index set and 
		${\cal D}_{\bm \mu,0} \coloneqq {\cal D}_{\bm \mu}$.
		Further, we suppose that there are diffeomorphisms 
		$\varphi_j\colon   {\cal D}_{\bm \mu,j} \rightarrow  {\cal D}_{\bm \mu}$
		such that for 
		$x \in {\cal D}_{\bm \mu,j}$ 
		it holds
		$\exp_{\bm \mu} \left( h (x) \right) = \exp_{\bm \mu} \left( (h \circ \varphi_j) (x) \right)$.
		Then, in order to obtain the pdf of $\bm Z$ in \eqref{definition}, we have to replace $p_Z$ in \eqref{stand_normal} by the wrapped function
		\begin{align} \label{stand_normal_sphere}
		\tilde p_Z(z) \coloneqq \frac{1}{(2 \pi)^{n/2}} \sum_{j \in {\cal J}} 
		{\rm e}^{-\frac12 \|\varphi_j^{-1} (z)\|_2^2 } \, |{\rm d} \varphi_j^{-1} (z)|,
		\qquad z \in {\cal D}_{\bf \mu}.
		\end{align}
		
		
		Now, we follow the same lines as in the Euclidean setting and 
		agree that $\bm X$ is normally distributed with mean $\bm \mu$ and positive definite covariance $\Sigma = A A^\tT$
		if $X =  h^{-1} (\log_\mu(\bm X)) = A Z \sim {\cal N} (0, \Sigma)$, respectively,
		\begin{equation} \label{definition_1}
		\bm X \coloneqq \exp_{\bm \mu} (h(X)), \qquad X \sim {\cal N} (0, \Sigma),
		\end{equation} 
		and write $\bm X \sim {\cal N}_{\MM} (\bm \mu, \Sigma)$.

		The following proposition shows how the pdf 
		of a normally distributed random point \eqref{definition} looks for
		various one-dimensional manifolds. 
		\\
		
		\begin{proposition} \label{1d_manifolds}
			The pdf of a random point  $\bm X \sim {\cal N}_\MM (\bm \mu,\sigma^2 I_n)$   
			is given by
			\begin{enumerate}[label = {\upshape(\roman*)}]
				\item  
				the {\emph{log-normal distribution}} for  $\MM= \R_{>0} = \SPD(1)$, 
				\begin{equation*}
				p_{\bm X}(\bm x)  
				=\frac{1}{\sqrt{2\pi \sigma^2}} \e^{-\frac{1}{2\sigma^2} ( \ln(\bm x)-\ln(\bm \mu))^2}
				\end{equation*}
				with respect to the measure $\dx_{\R_{>0}}(\bm x) = \tfrac{1}{ \bm x} \dx  \bm x$ on $\R_{>0}$;
				\item 
				the $2\pi$-\emph{wrapped Gaussian distribution} for $\MM = \SP^1$,
				\begin{align} \label{circle}
				p_{\bm X}(\bm x(t)) &=  
				\frac{1}{\sqrt{2\pi \sigma^2}}\sum_{j\in \Z} \e^{-\frac{1}{2\sigma^2} (t - t_\mu + 2j\pi)^2}
				\end{align}
				with respect to the parameterization
				\\
				$\bm \mu \coloneqq (\cos (t_\mu), \sin (t_\mu))^ \tT$
				and the Lebesgue measure $\!\dx t$;
				\item  
				the $2\pi$-\emph{wrapped, even shifted Gaussian distribution} for $\MM = \Delta_{1}$, 
				\begin{equation} \label{simplex}
				p_{\bm X}(\bm x(t))  =\frac{1}{\sqrt{2\pi \sigma^2}}  \sum_{j \in \mathbb Z} \Big(
				\e^{-\frac{1}{2\sigma^2} (t + t_\mu + 2j\pi)^2}
				+
				\e^{-\frac{1}{2\sigma^2} (t - t_\mu + 2j\pi)^2} \Big)
				\end{equation}
				with respect to the parameterization
				$
				\bm x(t) = \frac12 
				(
				1 + \cos (t), 1 - \cos (t)
				)^\tT
				$, $ t \in (0,\pi)$, 
				$
				\bm \mu = \frac12 
				(
				1 + \cos (t_\mu),
				1 - \cos (t_\mu)
				)^\tT
				$
				and Lebesgue measure 
				$\dx t$.
			\end{enumerate}
		\end{proposition}
		\vspace{0.2cm}
		The proof of the proposition is given in the Appendix \ref{app:prop}.
		
		The above definition \eqref{definition_1} of  normally distributed random points
		has the advantage that it adopts
		the affine invariance of the Gaussian distribution  known from the Euclidean setting
		via the tangent space.
		Moreover, it is easy to sample from the distribution.
		\vspace{0.1cm}
		
		\begin{remark}[Sampling from ${\cal N}_\MM (\bm \mu, \sigma^2 I_n)$]\label{sampling}
			Sampling of a ${\cal N}_\MM (\bm \mu, \sigma^2 I_n)$ distributed random variable can be performed as follows:
			i) sample from ${\cal N}(0,\sigma^2 I_n)$ in $\mathbb R^n$, 
			ii)
			apply $h$ which by \eqref{def_h} requires only the knowledge of 
			an orthogonal basis in $T_{\bm \mu} \MM$, and 
			iii) map the result by $\exp_{\bm \mu}$
			to $\MM$.
		\end{remark}
		\vspace{0.2cm}
		
		For the one-dimensional manifolds in Proposition \ref{1d_manifolds},
		the pdfs in (i) - (iii) are the kernels of
		the heat equations with the corresponding Laplace-Beltrami operators. 
		This is in general not true for higher dimensions \cite{Grig2009}. However, numerical experiments show that samples from the Gauss-Weierstrass kernel on $\mathbb S^2$ \cite[p. 112]{FGS1998}
		and from the heat kernel on ${\rm SPD}(r)$~\cite[p. 107]{terras1988} are very similar.
		For kernel density estimations on special Hadamard spaces we refer also to \cite{CBA2015}
		and for kernels in connection with dithering on the sphere we refer to \cite{GPS2012}.
		
\begin{figure}
			\centering
			\includegraphics[width = 0.45\textwidth]{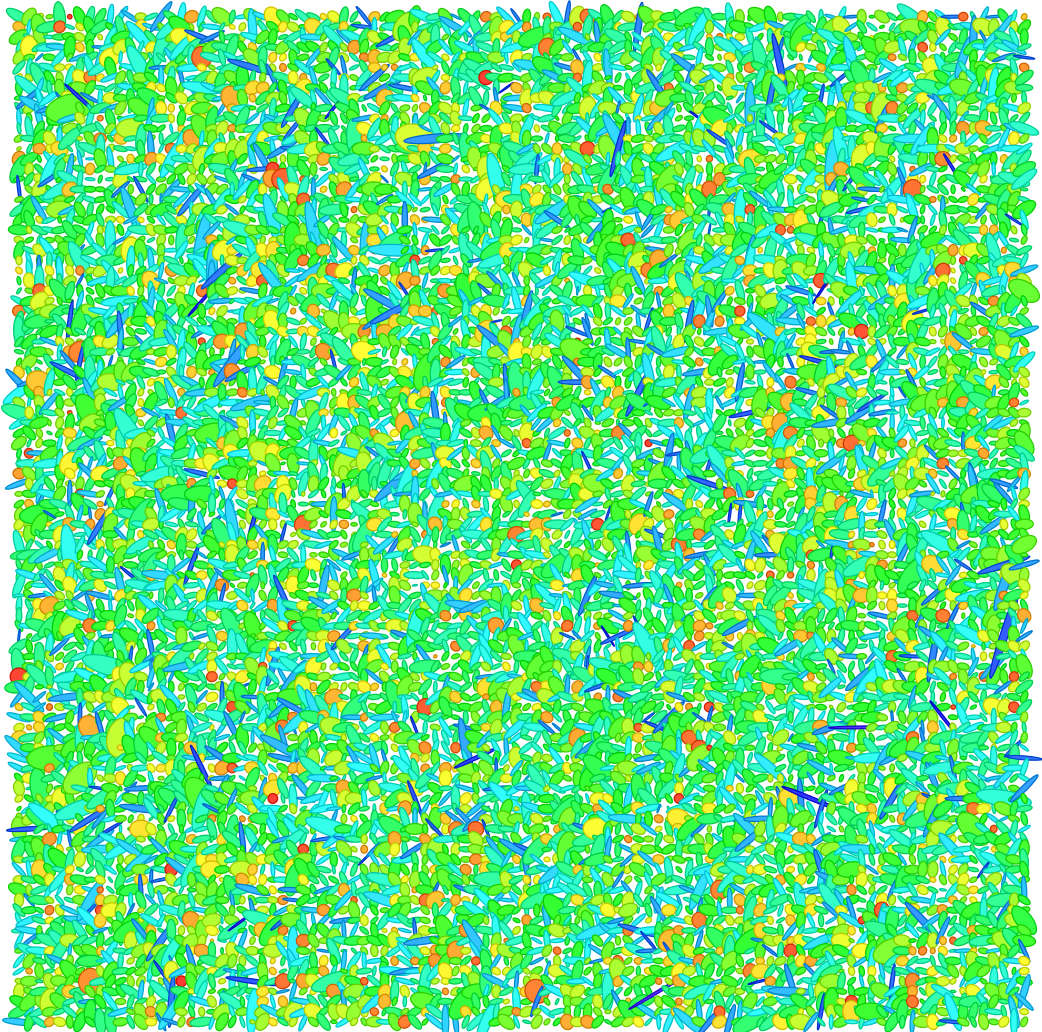}	
			\hspace{0.2cm}
			\includegraphics[width = 0.45\textwidth]{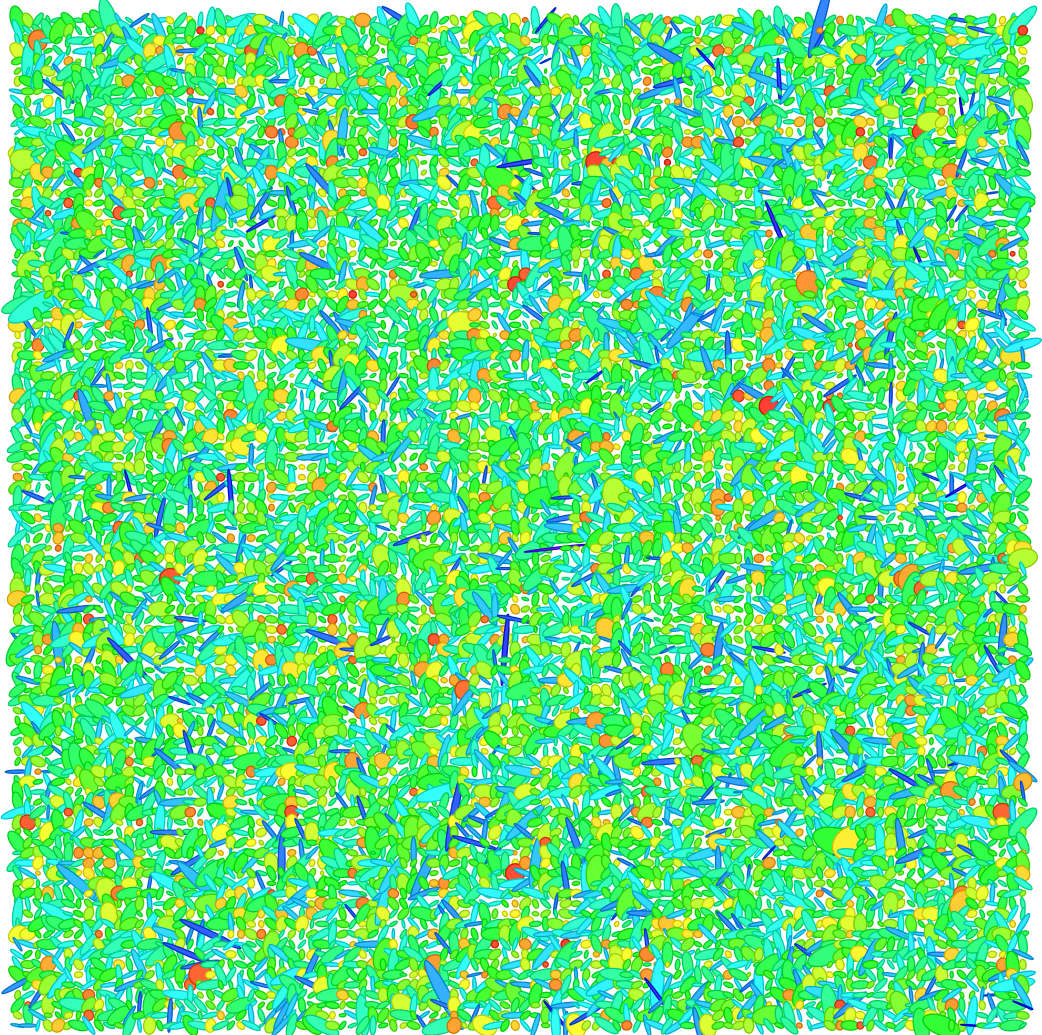}		
			\caption[]{Sampling with respect to our noise model  and that of Said et al.~\cite{SBBM15} for $100\times100$ samples 
				and  $\NN(I_2, \sigma^2 I_2)$ with $\sigma = 0.5$.}
			\label{fig:spd_noise}
		\end{figure}

		Neither the maximizing entropy nor the ML estimation property from Remark \ref{rem:prop_gaussian} 
		generalize to the above setting.
		In \cite{Pen06} Pennec showed that under certain conditions on ${\cal D}_T (0_{\bm \mu})$
		the pdf of a random point on $\MM$ that maximizes the entropy given  prescribed mean value $\bm \mu$ and covariance $\Sigma$ 
		is of the form $\frac{1}{\psi} \e^{- \frac12 (\log_{\bm \mu} (\bm x))^\tT \tilde{\Sigma} (\log_{\bm \mu} (\bm x))}$
		with normalization constant $\psi$. 
		Said and co-workers made use of the ML-estimator property in order to generalize the (isotropic) normal distribution to $\MM = {\rm SPD}(r)$ in \cite{SBBM15} 
		and to symmetric spaces of non-compact type in \cite{SHBV16}.
		They proposed the following density function for a normal distribution with mean $\bm \mu$ and covariance $\sigma^2 I_n$:
		\begin{equation} \label{entropy_min}
		p_{\bm X}(\bm x) = \frac{1}{\psi(\sigma)}\e^{-\frac{1}{2\sigma^2}\dist_{\MM}(\bm \mu, \bm x)^2}.
		\end{equation}
		For the concrete definition of $\psi$ and the noise simulation according to this model in the case of $\MM = \SPD(r)$ see~\cite{SBBM15} and
		the Appendix \ref{sec:said}. For $n=1$ the models coincide by Proposition \ref{1d_manifolds}.
		For this distribution it is clear that the  Riemannian center of mass is given by $\bm \mu$ 
		and that the ML estimator for $\bm \mu$ is the empirical Karcher mean.
		Figure \ref{fig:spd_noise}, left shows a $100 \times 100$ image of realizations of normally distributed noise on 
		${\SPD}(2)$ by our model $\NN(I_2, \sigma^2 I_2)$, where $\sigma = 0.5$.
		For comparison,
		in Figure \ref{fig:spd_noise}, right the same is shown for realizations of $\NN_{Said}(I_d,\sigma^2 I_d)$, also for $\sigma = 0.5$. The noise looks visually very similar, which is also confirmed by Table~\ref{tab:noise_parameters}.
		The first two rows of Table~\ref{tab:noise_parameters} present the estimated mean $\bm \mu$ based on \eqref{karcher_disc},
		the covariance matrix $\Sigma$ based on~\eqref{cov_empir} and the estimated standard deviation  $\sigma$
		for our noise model and those of Said et al..
		{\small
			\begin{table}[h]
				\centering
				\begin{tabular}{lccc} 		
					\toprule
					& $\bm \mu$ & $\sigma$ & $\Sigma$ \\
					\midrule 
					$\NN_{\mathrm Said}$ &$\begin{pmatrix} 0.9960& 0.0044\\0.0044 &0.9944
					\end{pmatrix}$ & $0.5051$ & $\begin{pmatrix*}[r]
					0.2513 & 0.0025 & -0.0044 \\ 0.0025 & 0.2533 & -0.0017 \\ -0.0044 & -0.0017 & 0.2608
					\end{pmatrix*}$ \\ \addlinespace
					
					$\NN$ &$\begin{pmatrix}0.9980 & 0.0036\\0.0036 & 0.9943
					\end{pmatrix}$ & $0.5026$ & $\begin{pmatrix*}[r]
					\phantom{-}0.2567 & 0.0017 & 0.0001 \\ 0.0017 & 0.2487 & -0.0008 \\ 0.0001 & -0.0008 & 0.2523
					\end{pmatrix*}$  \\			
					\bottomrule
				\end{tabular}
				\caption{Estimated parameters for the noise in Figure~\ref{fig:spd_noise}.}\label{tab:noise_parameters}
			\end{table}
		}
		
		Below we give an example for the above pdf~\eqref{entropy_min} and those in \eqref{stand_normal}  for the manifold $\MM= {\mathbb H}^2$.		
		\vspace{0.2cm}
		
		\begin{example}
			Let 
			${\mathbb H}^2 \coloneqq \{\bm x \in \mathbb R^3: x_1^2 + x_2 ^2 - x_3^2 = -1, \; x_3 >0\}$ 
			be the hyperbolic manifold equipped with the Minkowski metric
			$\langle \bm x, \bm y \rangle_{{\mathbb H}^2} \coloneqq x_1y_1 + x_2y_2 - x_3y_3$.
			The distance reads as
			$
			{\rm dist}_{{\mathbb H}^2} (\bm x, \bm y) = \arcosh \left(-\langle \bm x, \bm y \rangle_{{\mathbb H}^2} \right)
			$
			and
			\begin{align}
			\exp_{\bm x} (v) &= 
			\cosh \left( \sqrt{\langle v,v \rangle_{{\mathbb H}^2}} \right) \bm x 
			+ 
			\sinh \left( \sqrt{\langle v,v \rangle_{{\mathbb H}^2}}  \right) \frac{v}{\sqrt{\langle v,v \rangle_{{\mathbb H}^2}}},\\
			\log_{\bm x} (\bm y) &= 
			\frac{\arcosh \left(- \langle \bm x, \bm y \rangle_{{\mathbb H}^2} \right)}
			{\left(\langle \bm x, \bm y \rangle_{{\mathbb H}^2}^2 -1\right)^\frac12}
			\left(\bm x +  \langle \bm x, \bm y \rangle_{{\mathbb H}^2} \bm x\right).
			\end{align}
			We parametrize $\bm x\in {\mathbb H}^2$ as
			\begin{equation}
			\bm x(\alpha, r) =
			\begin{pmatrix}
			\cos(\alpha)\sinh(r)\\
			\sin(\alpha)\sinh(r)\\
			\cosh(r)
			\end{pmatrix},
			\quad \alpha \in [0,2\pi),\ r \in [0,\infty).
			\end{equation}
			First, we compute the pdf \eqref{stand_normal} of an ${\cal N}_{{\mathbb H}^2}(\bm \mu, \sigma^2 I_2)$ 
			distributed random point, where $\mu \coloneqq  (0,0,1)^\tT$.
			We obtain
			$
			{\rm dist}_{{\mathbb H}^2} (\bm \mu, \bm x) = r
			$ 
			and
			$\{ e_{\bm \mu,1} = (1,0,0)^\tT, \, e_{\bm \mu,2} = (0,1,0)^\tT \}$
			and for the other points ($r \not = 0$)
			$$
			e_{\bm x,1} 
			\coloneqq
			\frac{1}{\sinh(r)} \frac{\partial }{\partial \alpha} \bm x(\alpha,r) 
			= \begin{pmatrix}
			-\sin(\alpha)\\
			\cos(\alpha)\\
			0
			\end{pmatrix},
			\quad 
			e_{\bm x,2} 
			\coloneqq
			\frac{\partial }{\partial r} \bm x(\alpha,r) 
			= \begin{pmatrix}
			\cos(\alpha) \cosh( r)\\
			\sin(\alpha) \cosh( r)\\
			\sinh (r)
			\end{pmatrix}.
			$$
			Then the measure on ${\mathbb H}^2$ reads 
			${\rm d}_{{\mathbb H}^2} (\bm x) = \sinh( r ) \dx \alpha \dx r$.
			Straightforward computation gives
			$$
			\dx (\log_{\bm \mu})_{\bm x} [e_{\bm x,1}] = 
			\frac{r}{\sinh (r)}
			\begin{pmatrix}
			-\sin(\alpha) \\
			\cos(\alpha) \\
			0
			\end{pmatrix},
			\quad
			\dx (\log_{\bm \mu})_{\bm x} [e_{\bm x,2}] =
			\begin{pmatrix}
			\cos(\alpha) \\
			\sin(\alpha)\\
			0
			\end{pmatrix}
			$$
			so that
			$
			|\tilde G(\bm x)|^{\frac{1}{2}} = r/\sinh (r)
			$.
			Consequently, the density \eqref{stand_normal} is
			$$
			p_{\bm X} (\bm x(\alpha,r)) = \frac{1}{2 \pi\sigma^2} \e^{-\frac{r^2}{2\sigma^2}} \, \frac{r}{\sinh (r)}.
			$$
			In contrast, the entropy minimizing pdf \eqref{entropy_min}
			is given by
			\begin{equation}
			p_{\bm X} \bigl(\bm x(\alpha,r)\bigr) = 
			\frac{1}{\psi}
			\e^{-\frac{r^2}{2 \sigma^2}},
			\quad
			\psi 
			\coloneqq
			2\pi\int_0^\infty \e^{-\frac{r^2}{2\sigma^2}} \sinh(r) \dx r\\
			=
			2 \pi \sigma \e^{\frac{\sigma^2}{2}} \int_0^{\sigma}\e^{-\frac{t^2}{2}} \dx t.
			\end{equation}
		\end{example}
		\vspace{0.1cm}
		
		Besides the kernels of the heat equation,  the {\rm von Mises-Fisher distribution}
		is frequently considered as  ``spherical normal distribution'' on $\SP^d$.
		We briefly comment on this distribution.
		\vspace{0.1cm}
		\begin{remark}[Fisher-Mises distribution on $\SP^d$]\label{relation_FM}
			For $\SP^1$ it is well-known that
			the wrapped Gaussian distribution is closely related to the von Mises distribution $M(\bm \mu,\kappa)$ \emph{\cite{GGD1953,langevin1905, mises1918}}
			whose density function reads as
			\begin{equation}
			p_{{\mathrm{MF}}} (\bm x|\bm \mu,\kappa) =  \frac{1}{2\pi I_0(\kappa)} \,\e^{\kappa \cos(\bm x- \pi -\bm\mu)}, \qquad \bm x \in [-\pi,\pi),	\label{pdf_MF}
			\end{equation}
			where $I_n$ denotes is the modified Bessel function of first kind and order $n$. 
			The parameter $\bm\mu$ is referred to as \emph{mean direction}, 
			$\kappa > 0 $ is the \emph{concentration parameter}. The von Mises distribution is the distribution that maximizes the entropy under the constraint that the real and imaginary parts of the first {\rm circular} moment 
			(or, equivalently, the circular mean and circular variance) are specified. The maximum likelihood characterization 
			is analogously to the one given in Remark~\ref{rem:prop_gaussian}, where the sample mean is replaced by the {\rm sample mean direction}.
			A good matching between the pdfs of the wrapped Gaussian and those of the von Mises 
			for high concentration (i.e.\ large $\kappa$ respective small  $\sigma^2$) can be found
			by taking the  same center $\bm\mu$  and $\sigma^2  =-2\log\bigl(A(\kappa)\bigr)$
			with $A(\kappa) = \frac{I_1(\kappa)}{I_0(\kappa)}$, see \emph{\cite{MJ2000}}.\\
			The von Mises distribution on $\SP^1$ can be generalized to $\SP^{d}$, 
			leading to the \emph{von Mises-Fisher distribution} given by
			\begin{equation}
				p_{{\mathrm{MF}}} (\bm x|\bm \mu,\kappa) = \left(\frac{\kappa}{2}\right)^{\frac{d-1}{2}} \frac{1}{\Gamma\bigl(\tfrac{d+1}{2}\bigr) I_{\frac{d-1}{2}}(\kappa) } \,\e^{\kappa \bm\mu^\tT \bm x},\label{pdf_vonMises}
			\end{equation}
			where $\kappa >0$, $\lVert \bm\mu \rVert = 1$ and $\Gamma$ denotes the gamma function. 
			For $d = 2$, the von Mises-Fisher distribution is also known as \emph{Fisher distribution} and the pdf simplifies to 
			\begin{equation*}
				p_{{\mathrm{MF}}} (\bm x|\bm \mu,\kappa)= \frac{\kappa }{\sinh(\kappa)}\,\e^{\kappa \bm\mu^\tT \bm x}.
			\end{equation*}
			In Figure~\ref{Fig:noise_S2} we compare samples of our Gaussian noise model $\NN(0,\sigma^2)$ with the von Mises-Fisher distribution 
			$M(\bm\mu,\kappa)$ on $\SP^2$ for $\bm\mu = (0,0,1)$ and $\sigma^2 = \frac{1}{\kappa}$.
				\begin{figure}
			\centering
			\includegraphics[width = 0.3\textwidth]{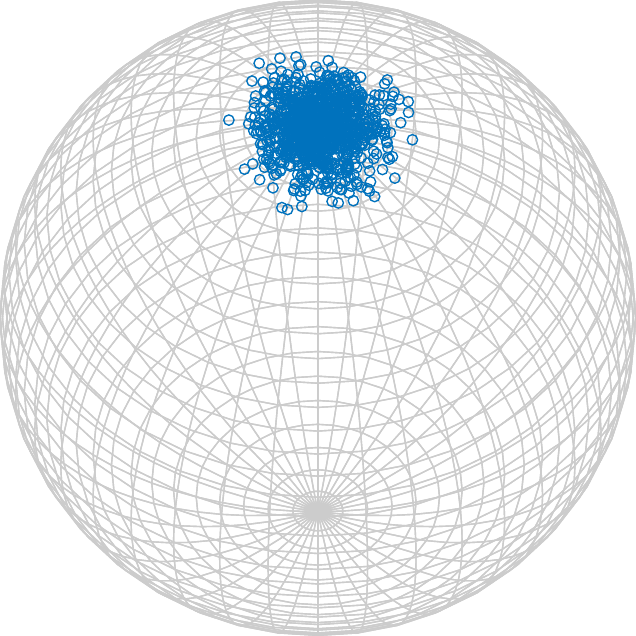}	
			\hspace{0.2cm}
			\includegraphics[width = 0.3\textwidth]{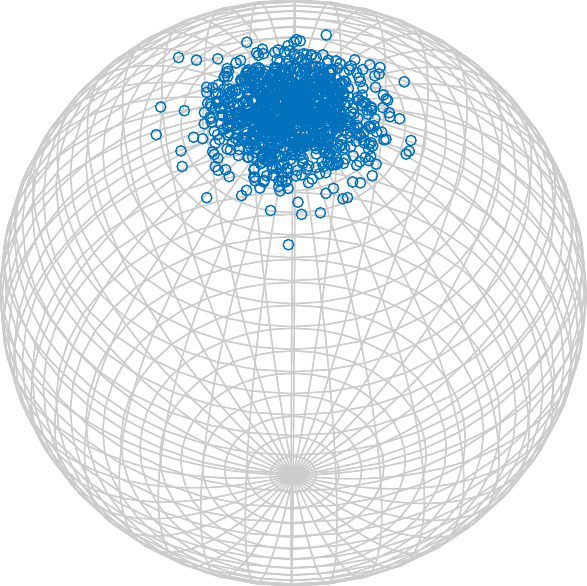}
			\hspace{0.2cm}
			\includegraphics[width = 0.3\textwidth]{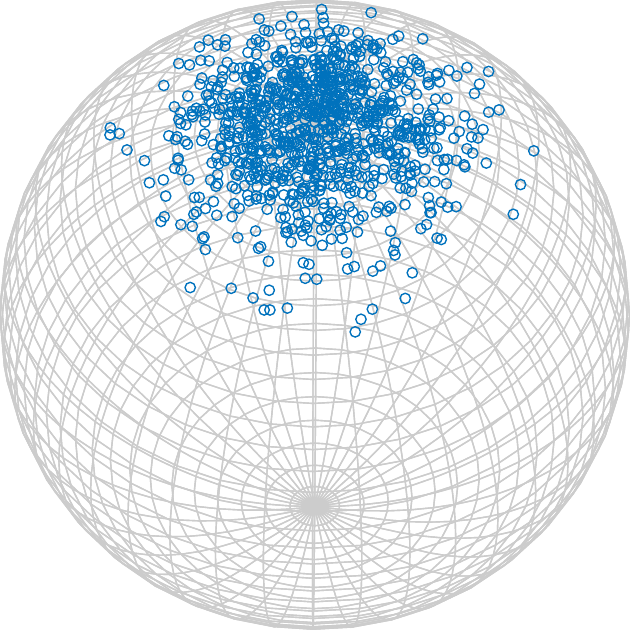}	
			\includegraphics[width = 0.3\textwidth]{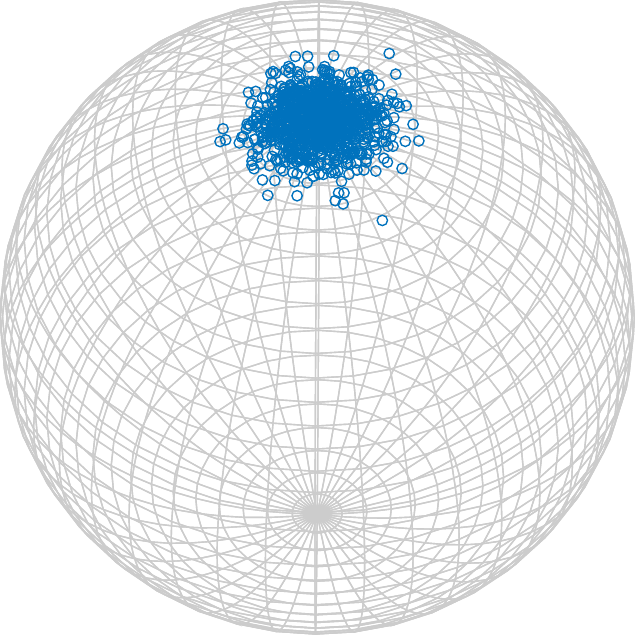}	
			\hspace{0.2cm}
			\includegraphics[width = 0.3\textwidth]{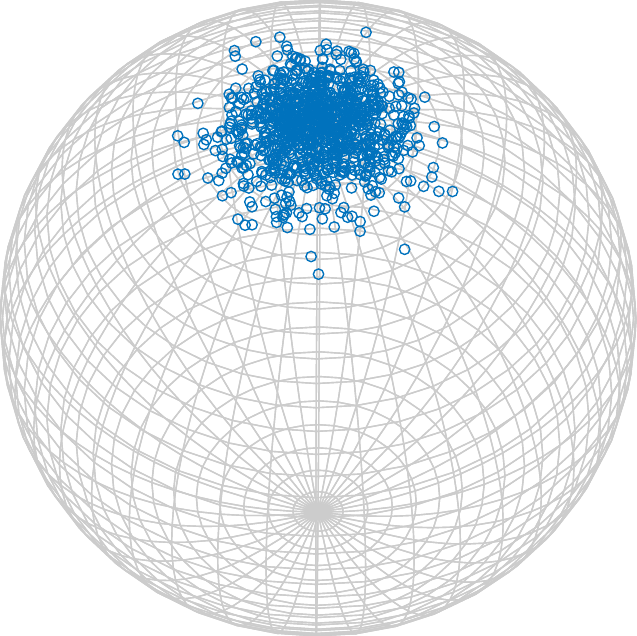}	
			\hspace{0.2cm}
			\includegraphics[width = 0.3\textwidth]{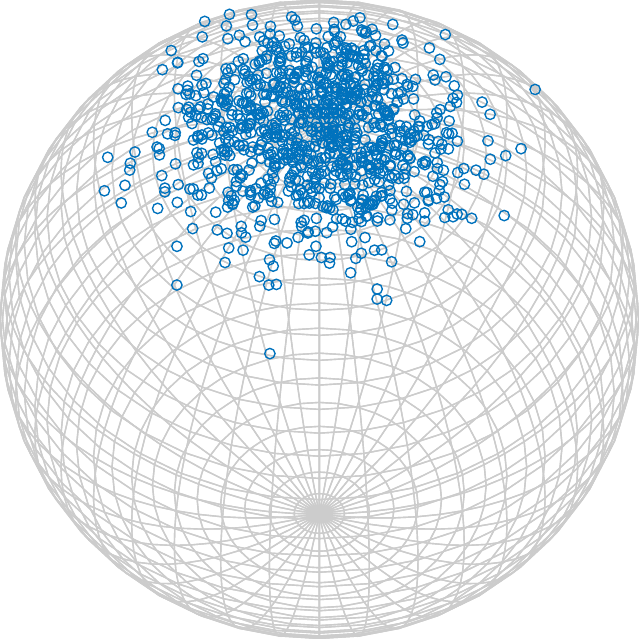}			
			\caption[]{1000 samples of Gaussian noise on $\SP^2$ for 
				$\mu = (0,0,1)$ and $\sigma\in \left\{\frac{1}{\sqrt{100}} ,\frac{1}{\sqrt{50}},\frac{1}{\sqrt{20}}\right\}$ 
				(top, from left to right) and of the von Mises-Fisher distribution on $\SP^2$ for $\mu = (0,0,1)$ 
				and $\kappa\in \{100,50,20\}$ (bottom, from left to right).}
			\label{Fig:noise_S2}
		\end{figure}
		\end{remark}
		\vspace{0.2cm}

		\subsection{NL-MMSE on Manifold-Valued Images} 
		
		Assume that $\bm Y = \exp_{\bm\mu}(Y)$ is a random point on $\MM$ 
		arising from a normally distributed 
		random point $\bm X = \exp_{\bm\mu}(X)  \sim {\cal N}_\MM(\bm\mu,\Sigma)$
		in the sense
		$$
		Y = X + \eta,\qquad X\sim \NN(0_{\bm \mu},\Sigma_X), \quad \eta\sim \NN(0,\sigma^2I_n),
		$$ 
		where by slight abuse of notation  we write
		$\exp_{\bm \mu}$ and $\log_{\bm \mu} $ instead of $ \exp_{\bm \mu} \circ h$ and
		$ h^{-1} \circ \log_{\bm \mu}$, respectively, as also done in Subsection \ref{gaussian_random_points}.
		In the following we propose an estimator for $\bm X$ based on $\bm Y = \exp_{\bm\mu} (Y)$, 
		which arises from a two-step estimation procedure and is motivated 
		by the Euclidean MMSE described in Section~\ref{Sec:MMSE}. 
		In order to avoid technical difficulties we restrict our attention to Hadamard manifolds $\MM$ such that the exponential 
		and logarithmic map are globally defined and the Riemannian center of mass exists and is uniquely determined. \\
		In the first step, we estimate the mean $\bm\mu = \E(\bm X)$ as
		\begin{align}\label{man_first}
		\bm \mu = \argmin_{\bm Z\in \sigma(\{\emptyset,\Omega\})} \E\bigl[\dist_\MM(\bm X,\bm Z)^2\bigr].
		\end{align}
		Note that this corresponds to the definition given in~\eqref{variance}, since those random variables $\bm Z$ 
		that are measurable with respect to the trivial $\sigma$-algebra $\{\emptyset,\Omega\}$ are exactly the constant random variables. 
		By construction we have $\E(\bm X) = \E(\bm Y)$.
		Once $\bm \mu$ is known, we estimate covariance matrix of $X$ by
		\begin{align}\label{man_sec}
		T_{\text{MMSE}}(\bm Y)&=	\argmin_{\log_{\bm \mu} (\bm Z) \in \sigma ( \log_{\bm \mu} (\bm Y))} 
		\E\bigl[\lVert \log_{\bm \mu}(\bm X) -\log_{\bm \mu}(\bm Z) \rVert_2^2\bigr]\\
		& = \argmin_{Z\in \sigma(Y)} \E\bigl[\lVert X-Z\rVert_2^2\bigr] = \E(X|Y).
		\end{align}
		In our specific Gaussian noise setting we are now in the same situation as described after Remark~\ref{rem:prop_gaussian}, 
		so that by combining \eqref{man_first} and \eqref{man_sec} we finally arrive at the estimator
		\begin{equation*}
		T(\bm Y) = \exp_{{\bm \mu}} 
		\bigl(
		( \Sigma_Y-\sigma^2 I_n)  \Sigma_Y^{-1} \, \log_{ {\bm \mu}}( \bm Y)
		\bigr).
		\end{equation*}
		Next, we describe how to estimate $\bm \mu$ and $\Sigma_Y$ based on samples. To this aim,  let ${x\colon \GG\to M}$ be a discrete image 
		defined on a grid $\GG = \{1,\ldots,N_1\}\times \{1,\ldots,N_2\}$ 
		with values in a $d$-dimensional manifold $M$. 
		As for real-valued images, we consider small $s \times s$ image patches
		centered at $i = (i_1,i_2)\in {\cal G}$.
		We assume that the patch  $\bm y_i$ corresponds to a realization of a  normally 
		distributed random point $\bm Y_i \sim \NN_{\MM}({\bm \mu}_i,\Sigma_i)$ on $\MM$, where
		$\MM = M^{s^2}$ is the product manifold of dimension $n  = s^2 d$ 
		equipped with the distance
		${\rm dist}_{\MM}^2 (\bm x, \bm y) = \sum_{j=1}^{s^2} \dist_M (\bm x_j,\bm y_j)^2$.
		We fix $K\in \N$ and take the $K$ nearest patches with respect to $\dist_{\MM}$
		in a  $w \times w$ search window around~$i$. 
		These patches are interpreted as other realizations of the same random point.
		Let $\SS(i)$ denote the set of centers of the patches similar to $\bm y_i$.
		Then the empirical estimates for the mean and the covariance 
		in   \eqref{karcher_disc}, respective~\eqref{cov_empir} read as
		\begin{align*}
		\hat{{\bm \mu}}_i  \in \argmin_{{\bm \mu} \in \MM} \sum_{j \in \SS(i)} \dist_\MM({\bm \mu}, {\bm y}_j)^2, 
		\qquad
		\hat{\Sigma}_i = \frac{1}{K}\sum_{j \in \SS(i)} \log_{\hat{{\bm \mu}}_i}(\bm y_j) \log _{\hat{{\bm \mu}}_i}({\bm y}_j)^\tT,
		\end{align*}
		and are used to restore the reference patch and all its similar patches by 
		\begin{equation} \label{MMSE_manifold}
		\hat{\bm y}_j = \exp_{\hat {\bm \mu}_i} 
		\bigl(
		(\hat \Sigma_i-\sigma^2 I_n) \hat \Sigma_i^{-1}(\log_{\hat {\bm \mu}_i}(\bm y_j))
		\bigr),
		\qquad j\in\SS(i). 
		\end{equation}
		This can be considered as the manifold counterpart to \eqref{3a}.

		With slight modifications the fine-tuning details listed in Remark \ref{details} can be generalized to manifolds. The treatment of patches at the boundaries, 
		the acceleration and the second step are analogously as in the real-valued case, only for flat areas and the aggregation step one has to replace the empirical variance 
		respective the mean by their manifold counterparts. The two steps of the algorithm are summarized in Algorithm~\ref{Alg:NL_M}.
		
		\begin{algorithm}[htp]
			\caption{Nonlocal MMSE Denoising Algorithm on $M$, Step 1}
			\label{Alg:NL_M}
			\begin{algorithmic}
				\State \textbf{Input:}
				noisy image $\bm y\in M^{N}$, variance $\sigma^2$ of noise
				\State \textbf{Output:}	first step denoised image $\hat{\bm y}$ and final image $\tilde{\bm y}$
				\State \textbf{Parameters:} $s_1,s_2$ sizes of patches, $K_1,K_2$ numbers of similar patches, $\gamma$ homogeneous area parameter, $w_1,w_2$ sizes of search area
				\State \textbf{Step 1:}
				\State Set $\mathcal{M} = M^{s_1^2}$
				\For{all patches $\bm y_i$ of the noisy image $y$ not considered before}	
				\State
				Determine the set $\SS_1(i)$ of centers of $K_1$ patches similar to $\bm y_i$ in a $w_1\times w_1$ window around $i$
				with respect to the distance measure on the manifold
				\State	Compute by a gradient descent algorithm the Karcher mean patch, $\hat{\mu}_i = (\hat{\mu}_{i,j})_{j=1}^{s_1^2}$,   
				\begin{align*}
				\hat{\bm \mu}_i &\in \argmin\limits_{\bm y\in \MM} \biggl\{\frac{1}{K_1} \sum_{j\in\SS_1(i)} \dist_{\MM}(\bm y,\bm y_j)^2\biggr\}
				\end{align*}
				\State \textbf{Homogeneous area test:} Compute by a gradient descent algorithm the Karcher mean value
				$
				\hat{\bm m}_i \in  \argmin\limits_{\bm y\in M} \biggl\{\frac{1}{K_1 s_1^2} \sum_{j\in\SS_1(i)}\sum_{k=1}^{s_1^2} \dist_{M}(\bm y,\bm y_{j,k})^2\biggr\}
				$
				and 		the empirical variance of the patches
				\State \begin{equation*}
				\hat{\sigma}^2_i = \frac{1}{d K_1 s_1^2}\sum_{j \in \SS_1(i)} \sum_{k=1}^{s_1^2}\dist_{M}(\hat{\bm m}_i,\bm y_{j,k})^2
				\end{equation*}
				\If{$\hat{\sigma}^2_i \leq \gamma \sigma^2$} 
				\State Compute the restored patches as $\hat{\bm y}_j = \mathbf{1}_{s_2^2}\otimes 	\hat{\bm m}_i$, $j\in \SS_1(i)$
				\Else
				\State Compute the empirical covariance matrix
				\begin{equation*}
				\hat{\Sigma}_i =\frac{1}{K_1}\sum_{j\in \SS_1(i)} \log_{\hat{\bm \mu}_i}(\bm y_j)\log_{\hat{\bm \mu}_i}(\bm y_j)^\tT 
				\end{equation*}
				\State Compute the restored patch  $\hat{\bm y}_j = \exp_{\hat{\bm \mu}_i}\bigl(
				( \hat{\Sigma}_i- \sigma^2 I_{s_1^2}) \hat{\Sigma}_i^{-1} \log_{\hat{\bm \mu}_i}(\bm y_j)\bigr)$, ${j\in \SS_1(i)}$
				\EndIf
				\State	\textbf{Aggregation:} Obtain the first estimate $\hat{\bm y}$ at each pixel by computing the Karcher mean over all restored patches containing the pixel	
				\EndFor	
				\algstore{manifoldalg}
			\end{algorithmic} 	
		\end{algorithm}

		\begin{algorithm}[htp]
			\ContinuedFloat
			\caption{Nonlocal MMSE Denoising Algorithm, Step 2}
			\begin{algorithmic}
				\algrestore{manifoldalg}
				\State \textbf{Step 2:}
				\State Set $\mathcal{M} = M^{s_2^2}$
				\For{all patches $\bm y_i$ of the noisy image $\bm y$ not considered before}	
				\State
				Determine the set $\SS_2(i)$ of centers of $K_2$ patches similar to the denoised image $\hat{y}_i$ in the first step 
				in a $w_2\times w_2$ window around $i$
				\State	Compute the Karcher mean patch, $\hat{\mu}_i = (\hat{\mu}_{i,j})_{j=1}^{s_1^2}$,  
				\begin{align*}
				\tilde{\mu}_i &\in \argmin\limits_{\bm y\in \MM} \biggl\{\frac{1}{K_2} \sum_{j\in\SS_2(i)} \dist_{\MM}(\bm y,\bm y_j)^2\biggr\} 
				\end{align*}
				\State \textbf{Homogeneous area test:} Compute the Karcher mean value 
				$
				\tilde{\bm m}_i \in  \argmin\limits_{\bm y\in M} \biggl\{\frac{1}{K_2 s_2^2} \sum_{j\in\SS_2(i)}\sum_{k=1}^{s_2^2} 
				\dist_{M}(\bm y,\bm y_{j,k})^2\biggr\}
				$
				and the empirical variance of the patches
				\State \begin{equation*}
				\tilde{\sigma}^2_i = \frac{1}{d K_2 s_2^2}\sum_{j \in \SS_2(i)} \sum_{k=1}^{s_2^2}\dist_{M}(\tilde{\bm m}_i,\bm y_{j,k})^2
				\end{equation*}
				\If{$\tilde{\sigma}^2_i \leq \gamma \sigma^2$} 
				\State Compute the restored patches  $\tilde{\bm y}_j = \mathbf{1}_{s_2^2}\otimes 	\tilde{\bm m}_i$, $j\in \SS_2(i)$
				\Else
				\State Compute the empirical covariance matrix
				\begin{equation*}
				\widetilde{\Sigma}_i =\frac{1}{K_2}\sum_{j\in \SS_1(i)} \log_{\tilde{\mu}_i}(\hat{\bm y}_j)\log_{\tilde{\mu}_i}(\hat{\bm y}_j)^\tT  + \sigma^2 I_{s_2^2}
				\end{equation*}
				\State Compute the restored patch $\tilde{\bm y}_j = \exp_{\tilde{\bm \mu}_i}\bigl((\widetilde{\Sigma}_i- \sigma^2 I_{s_2^2}) \widetilde{\Sigma}_i^{-1}  
				\log_{\tilde{\bm \mu}_i}(\bm y_j)\bigr)$, ${j\in \SS_2(i)}$
				\EndIf
				\State	\textbf{Aggregation:} Obtain the final estimate $\tilde{\bm y}$ at each pixel by computing the Karcher mean  over all restored patches containing the pixel	
				\EndFor
			\end{algorithmic} 	
		\end{algorithm}

		\FloatBarrier
		\section{Numerical Results}\label{sec:numerics}
		In this section we provide numerical examples to illustrate the   good performance of the
		NL-MMSE Algorithm~\ref{Alg:NL_M}.
		As manifolds we consider the circle $\SP^1$, the sphere $\SP^2$
		and the positive definite matrices $\SPD(r)$ for $r=2,3$. 
		While Algorithm~\ref{Alg:NL_M} is implemented in \textsc{Matlab}, the basic manifold functions, like logarithmic and exponential maps, 
		as well as the distance function are implemented as C++ functions in 
		\href{http://www.mathematik.uni-kl.de/imagepro/members/bergmann/mvirt/}{\glqq Manifold-valued 
			Image Restoration Toolbox\grqq (MVIRT)}\footnote{http://www.mathematik.uni-kl.de/imagepro/members/bergmann/mvirt/}
		and imported into \textsc{Matlab} using \lstinline!mex!-interfaces with the GCC 4.8.4 compiler.
		The experiments are carried out on a Dell Precision T1500 running 
		Ubuntu 14.04 LTS, Core i7, 2.93 GHz, and 8 GB RAM, using \textsc{Matlab} 2014b.
		
		To compare different methods we used as performance measure the mean squared error 
		\begin{equation}
		\epsilon = \frac{1}{N}\sum_{i\in\mathcal{G}} \dist_M(\hat{\bm x}_i,\bm x_i)^2,
		\end{equation}
		where $x$ denotes the original image and $\hat{x}$ is the restored one. 
		The parameters of all involved algorithms were optimized with respect to this error measure on the grids
		detailed below.
		We compared the following denoising methods:
		\begin{enumerate}[label = {\upshape(\roman*)}]
			\item {\bf NL-MMSE}: 
			we implemented Algorithm~\ref{Alg:NL_M} with parameters from the following grid search
			(in \textsc{Matlab} notation):
			patch size $s\in\{3:2:11\}$, window size $w\in\{9:2:127\}$, number of neighbors $K\in\{1:1:1200\}$, 
			and $\gamma\in\{0.1:0.1:2\}$. 
			We would like to mention that we started on coarser grids and refined them during the parameter search. 
			We briefly comment on general guidelines for the parameter in the following subsection.
			The final parameters for our experiments are listed in Table~\ref{tab:algparam}.
			Note that in the first three experiments we optimized only one set of parameters, 
			i.e., they are the same in both steps, 	while the parameters for both steps are optimized for
			the last three examples.
			
			\item {\bf NL-means}:
			we implemented a generalization of the NL-means algorithm~\cite{BCM2005,S2010}
			for manifold-valued images. Since this algorithm is not available in the general form
			required for our noisy images, we describe it in the next subsection. Concerning the grid search
			we used the same grids as in (i) for $s,w,K$. Further, $\delta $ is optimized on $\{0.5:0.5:50\}$, 
			and $\tau$ on $\{0.1:0.1:1\}$.
			
			\item {\bf TV approach}: 
			we applied the manifold version of the variational denoising approach with
			${\rm dist}_\MM^2$ as data fidelity term and anisotropic discrete total variation (TV) term as 
			proposed in~\cite{WDS2014}. 
			Furthermore, for cyclic data we also 
			added a second regularization term to (iii), called $\operatorname{TV}_2$, 
			which is a manifold version of second order differences.
			This method was proposed for the circle in~\cite{BLSW14} and for more general symmetric spaces in~\cite{BBSW2015} and
			we used the corresponding programs. Using the notation from~\cite{BBSW2015}, we did a grid search
			for the regularization parameter $\alpha$ of the $\operatorname{TV}$ term in $\{ 0.01:0.01:1\}$
			and for the regularization parameter $\beta$ of the $\operatorname{TV}_2$ term in $\{0.1:0.1:5\}$.
			The main drawback of the variational methods are their extensive running time compared to (i) and (ii).
		\end{enumerate}
		
		\begin{table}[tbp]
			\centering
			\begin{tabular}{l ccccccc}
				\toprule
				Figure & $s_1$ & $s_2$ & $w_1$ & $w_2$ & $K_1$ & $K_2$ & $\gamma$\\
				\midrule			
				Figure~\ref{fig:corals:hue_denoised} &  $7$ & $7$ & $81$ & $81$ & $70$ & $70$ & $1$ \\
				Figure~\ref{fig:s2coral:denoisedchroma} &  $5$ & $5$ & $37$ & $37$ & $110$ & $110$ & $1$\\
				Figure~\ref{fig:spd3} &  $5$ & $5$ & $59$ & $59$ & $415$ & $415$ & $.8$\\
				
				Figure~\ref{fig:spd2} &  $9$ & $9$ & $115$ & $115$ & $1038$ & $1038$ & $1$\\
				
				Figure~\ref{fig:arts1} & $9$ & $7$ & $119$ & $123$ & $186$ & $86$ & $1.1$\\
				
				Figure~\ref{fig:arts2} &  $3$ & $5$ & $127$ & $127$ & $65$ & $54$ & $0.8$\\		
				
				\bottomrule
			\end{tabular}
			\caption{Parameters for the NL-MMSE Algorithm \ref{Alg:NL_M} in the examples.}\label{tab:algparam}	
		\end{table}
		
		
		\subsection{Parameter Selection}\label{subsec:parameters}
		Our algorithm requires several input parameters. 
		Besides the variance $\sigma^2$ of the noise (which is assumed to be known or otherwise may be estimated in constant areas), 
		these are the size of the patches and of the search zone as well as the number of similar patches that 
		are kept and a parameter for the homogeneous area criterion. Even if it is not possible to state general parameter constellations 
		that are valid for the different manifolds there are some general principles how to choose good parameters. Based on these principles 
		we may obtain a first set of parameters, which may be fine-tuned by varying one of them while keeping the rest fixed.
		\begin{enumerate}[label = {\upshape(\roman*)}]
			\item patch size $s$: In general, the considered patches are rather small ($s\in \{3,5,7\}$), their exact value depends on the amount of noise, 
			measured in terms of the variance $\sigma^2$ of the noise. The higher the noise, the larger the patches, as they contain less information 
			when the noise level is high. 
			\item window size $w$: The size of the search  zone depends on the one hand on the patch size and on the other hand on the number of similar patches, 
			which depends itself on the dimension of the manifold. The larger those values are, the larger the search zone should be. 
			\item number of similar patches $K$: The number of similar patches has to be large enough to guarantee that the estimated covariance matrix 
			is invertible with high probability, which depends on the patch size and on the dimension $d$ of the manifold. On the other hand, 
			it should also not be too large as in this case also non-similar patches are chosen. 
			As a rule of thumb we observed that $K = 3 s^2 d$ yields good results in practice.
			\item homogeneous area parameter $\gamma$: 
			This value should be close to 1 and it should be the larger, the more constant areas an image contains.
		\end{enumerate}
		
		\subsection{Nonlocal Means on Manifolds}\label{subsec:nl_manifold}
		In this section we briefly discuss how to generalize 
		the NL-means approach introduced in~\cite{BCM2005} to manifolds. 
		The fundamental difference to the NL-MMSE lies, besides some details, 
		in the incorporation of second order information.
		
		Let $y\colon\mathcal{G}\rightarrow M$ be a noisy manifold-valued image. Consider a $s \times s$
		patch $y_i \in M^{s^2}$ centered at $i = (i_1,i_2)\in {\mathcal G}$. For each $i\in\mathcal{G}$ 
		we denote by $\mathcal{S}(i)$ the set of $K$ similar patches to $y_i$, 
		selected in a $w\times w$ search window around $y_i$. 
		Similar patches are found with respect to a weighted distance on the product manifold, i.e.,
		\begin{equation}
		\widetilde{\dist}_{M^{s^2}}(y_i,y_j)^2 
		\coloneqq \sum_{k_1=-\lfloor\frac{s-1}{2}\rfloor}^{\lfloor\frac{s-1}{2}\rfloor}
		\sum_{k_2=-\lfloor\frac{s-1}{2}\rfloor}^{\lfloor\frac{s-1}{2}\rfloor} \e^{-\frac{1}{2\delta^2}(k_1^2+k_2^2)}
		\dist_M(y_{i_1+k_1,i_2+k_2},y_{j_1+k_1,j_2+k_2})^2,
		\label{eq:nl:sim}
		\end{equation}
		where $\delta > 0$, $y_i\in M^{s^2}$ denotes the whole patch and $y_{i_1,i_2}\in M$ denotes a pixel value. \\
		The aggregation step is done by averaging the patch centers, weighted by the distance of the patches, i.e., let
		\begin{align}\label{eq:nlweights}
		\omega_{i,j} = \begin{cases}
		\e^{-\frac{1}{2\tau^2}\widetilde{\dist}_{M^{s^2}}(y_i,y_j)^2}&\  i\neq j,\\
		\max_{j\in\mathcal{S}(i),j\neq i}\{\omega_{i,j}\}&\  i = j,
		\end{cases}\quad\text{and}\qquad 
		W_i = \sum_{j\in\mathcal{S}(i)} \omega_{i,j}.
		\end{align}
		Then the restored pixel value is given by
		\begin{equation}
		\hat{y}_{i} = \argmin_{y\in M} \biggl\{\frac{1}{W_i}\sum_{j\in\mathcal{S}(i)}\omega_{i,j} \dist_M(y,y_{j})^2\biggr\}.\label{eq:nl:mean}
		\end{equation}
		For the weights in \eqref{eq:nlweights} we use the maximal weight approach for the center patch, 
		which was proposed in \cite{S2010} as the best choice without introducing an extra parameter.
		
		Let us briefly comment on two related approaches.
		A  NL-means denoising algorithm for DT-MRI images was given in~\cite{WPCMB07}. The
		authors use the affine invariant distance on $\SPD(r)$ as similarity measure in \eqref{eq:nl:sim}  
		and the log-Euclidean mean for computing the mean \eqref{eq:nl:mean},
		while we perform both steps  with the same affine invariant distance measure. 
		The authors of \cite{PBDW08} introduce a semi-local method for denoising manifold-valued data 
		motivated by the corresponding variational method for real valued data. 
		Their method performs an iterative averaging over circular shaped neighborhoods 
		with weights depending on the pixel similarity and distances between the pixels on the image grid. 
		In contrast to  \cite{PBDW08} we consider patches around the pixels for computing their similarity.
		
		\subsection{Noise on Color Channels} \label{subsec:color_channels}
\begin{figure}[tp]
			\centering
			\begin{subfigure}[t]{0.24\textwidth}
				\centering
				\includegraphics[width = 0.98\textwidth]{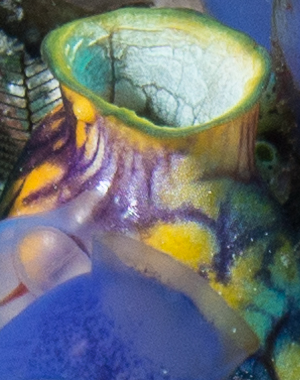}
				\caption[]{Original image}
			\end{subfigure}		
			\begin{subfigure}[t]{0.24\textwidth}
				\centering
				\includegraphics[width = 0.98\textwidth]{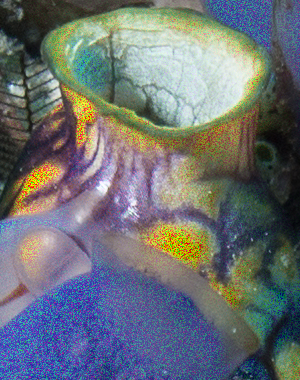}
				\caption[]{Noisy image}
			\end{subfigure}	
			\begin{subfigure}[t]{0.24\textwidth}
				\centering
				\includegraphics[width = 0.98\textwidth]{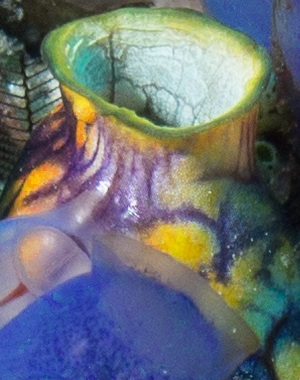}
				\caption[]{$\TV$ denoised image}	
			\end{subfigure}
			\begin{subfigure}[t]{0.24\textwidth}
				\centering
				\includegraphics[width = 0.98\textwidth]{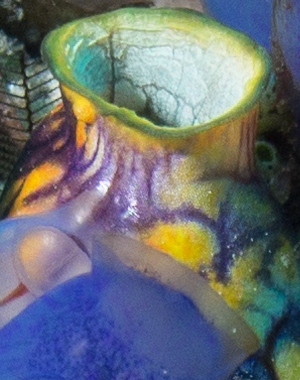}
				\caption[]{NL-MMSE denoised image}	
			\end{subfigure}		
			
			\begin{subfigure}[t]{0.24\textwidth}
				\centering
				\includegraphics[width = 0.98\textwidth]{orig_corals}
				\caption[]{Original image}
			\end{subfigure}
			\begin{subfigure}[t]{0.24\textwidth}
				\centering
				\includegraphics[width = 0.98\textwidth]{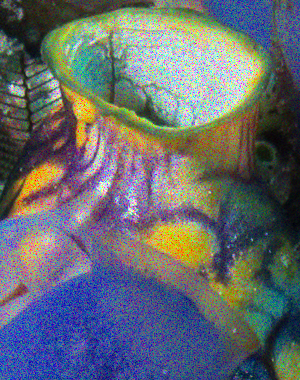}
				\caption[]{Noisy image}
			\end{subfigure}
			\begin{subfigure}[t]{0.24\textwidth}
				\centering
				\includegraphics[width = 0.98\textwidth]{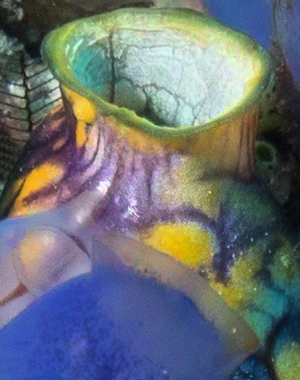}
				\caption[]{$\TV$ denoised image}
			\end{subfigure}
			\begin{subfigure}[t]{0.24\textwidth}
				\centering
				\includegraphics[width = 0.98\textwidth]{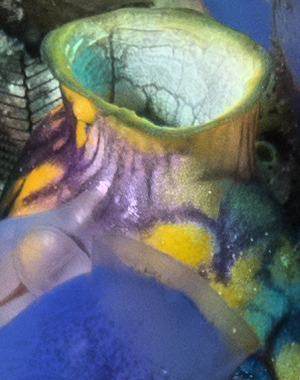}
				\caption[]{NL-MMSE denoised image}
			\end{subfigure}			
		\caption[]{Original (first column), noisy (second column), $\TV$ denoised (third column) and NL-MMSE denoised (fourth column) sponge image in the HSV (top row) and the CB (bottom row) color space.}\label{fig:coral_together}			
		\end{figure}

		Manifold-valued images naturally appear in various color image models different from the RGB model. 
		In the following, we consider the hue-saturation-value (HSV) and the chromaticity-brightness (CB)  color model. 
		We added Gaussian noise to the hue and the cromaticity  channels. 
		We are aware of the fact, that natural color images are in general not corrupted by Gaussian noise in only one color channel. 
		However, we provide these academical examples as a proof of concept. 
		How such  single channel noise affects the whole image can be seen in 
		Figure~\ref{fig:coral_together}.		
		
\begin{figure}[tp]
			\centering
			\begin{subfigure}[t]{0.24\textwidth}
				\centering
				\includegraphics[width = .98\textwidth]{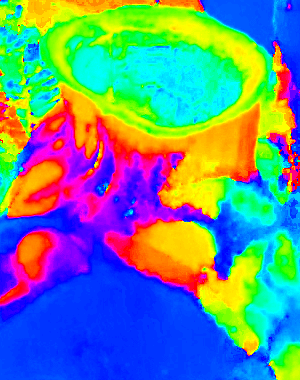}
				\caption[]{Original hue}\label{fig:corals:hue}
			\end{subfigure}	
			\begin{subfigure}[t]{0.24\textwidth}
				\centering
				\includegraphics[width = .98\textwidth]{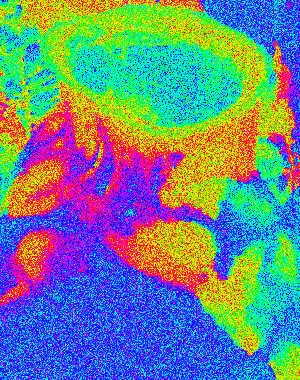}
				\caption[]{Noisy hue,\\ $\epsilon = 0.3609 $}\label{fig:corals:hue_noisy}
			\end{subfigure}
			\begin{subfigure}[t]{0.24\textwidth}
				\centering
				\includegraphics[width = .98\textwidth]{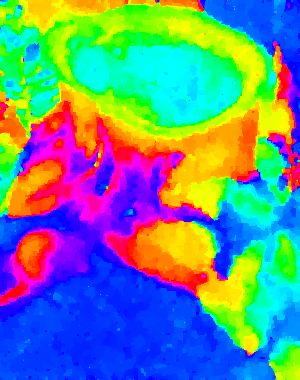}
				\caption[]{TV approach,\\ $\epsilon = 0.0263$}
				\label{fig:corals:hue_tv}
			\end{subfigure}
			\begin{subfigure}[t]{0.24\textwidth}
				\centering
				\includegraphics[width = .98\textwidth]{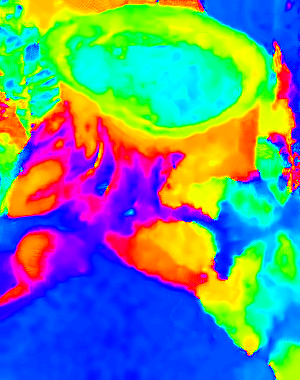}
				\caption[]{NL-MMSE,\\ $\epsilon = 0.0194$}
				\label{fig:corals:hue_denoised}
			\end{subfigure}
			
			\begin{subfigure}[t]{0.24\textwidth}
				\centering
				\includegraphics[width = 0.98\textwidth]{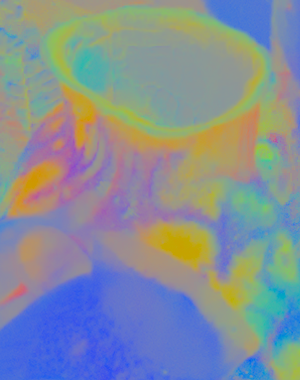}
				\caption[]{Original chromaticity}\label{fig:s2coral:chroma}
			\end{subfigure}
			\begin{subfigure}[t]{0.24\textwidth}
				\centering
				\includegraphics[width = 0.98\textwidth]{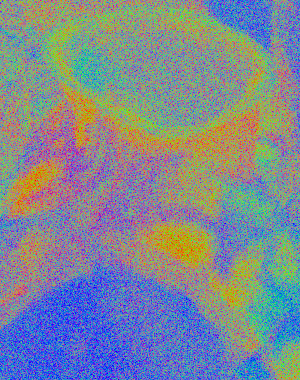}
				\caption[]{Noisy chromaticity,\\ $\epsilon =0.0798 $}\label{fig:s2coral:noisychroma}
			\end{subfigure}
			\begin{subfigure}[t]{0.24\textwidth}
				\centering
				\includegraphics[width = 0.98\textwidth]{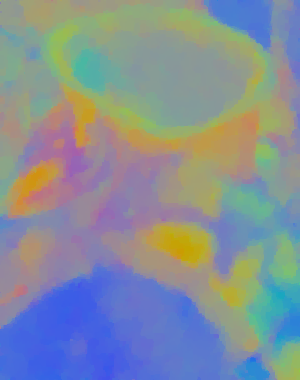}
				\caption[]{
					TV approach,\\ $\epsilon =0.0021 $}\label{fig:s2coral:tvchroma}
			\end{subfigure}
			\begin{subfigure}[t]{0.24\textwidth}
				\centering
				\includegraphics[width = 0.98\textwidth]{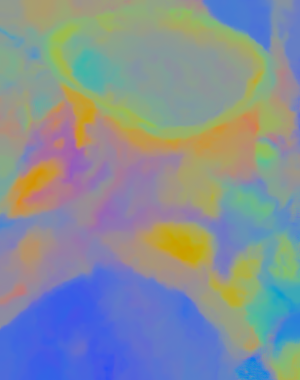}
				\caption[]{NL-MMSE,\\ $\epsilon =0.0017 $}\label{fig:s2coral:denoisedchroma}
			\end{subfigure}
			
			\caption[]{Denoising the hue and chromaticity of the color image \emph{sponge}.}
			\label{fig:coral}
			
		\end{figure}

		Cyclic data appears in the hue component of the HSV color model. 
		The hue component of the \emph{sponge} is considered in the  first row of Figure~\ref{fig:coral}.
		The second column~\ref{fig:corals:hue_noisy} shows the noisy hue 
		corrupted by wrapped Gaussian noise of standard deviation $\sigma = 0.6$. 
		Applying  the TV denoising method with optimized parameters $\alpha = 0.45,\ \lambda=\tfrac{\pi}{2}$, resp., NL-MMSE, 
		leads to the results in Figure~\ref{fig:corals:hue_tv}, resp. Figure~\ref{fig:corals:hue_denoised}.
		Despite rather flat areas in the image, the NL-MMSE approach outperforms the variational TV method.
		
		Spherical data occurs in the chromaticity component of the CB color space. 
		At this point, the chromaticity is defined as the direction of the RGB color vector and the brightness is given by its length. 
		We deal with the chromaticity of the \emph{sponge} image in the second row of Figure~\ref{fig:coral}. We corrupted it
		by Gaussian noise of standard deviation $\sigma = 0.2$, which yields the image shown in~Figure~\ref{fig:s2coral:noisychroma}. 
		Figure~\ref{fig:s2coral:tvchroma} gives the  result of the TV method with $\alpha = 0.21,\ \lambda=\tfrac{\pi}{2}$.
		Denoising with the NL-MMSE results in Figure~\ref{fig:s2coral:denoisedchroma} which is again better than the previous one.

		\subsection{Matrix-Valued Data} \label{subsec:spd}
		In this subsection, we provide two examples for images having values in  $\SPD(r)$ for $r=2,3$.
		A  matrix $\bm x\in\SPD(r),\ r = 2,3$, is depicted as an ellipse ($r = 2$) or an ellipsoid ($r=3$) 
		whose principal axis are determined by the spectral decomposition of $\bm x$. 
		
		\begin{figure}[tpb]
			\centering
			\begin{subfigure}[t]{0.45\textwidth}
				\centering
				\includegraphics[width = 0.98\textwidth]{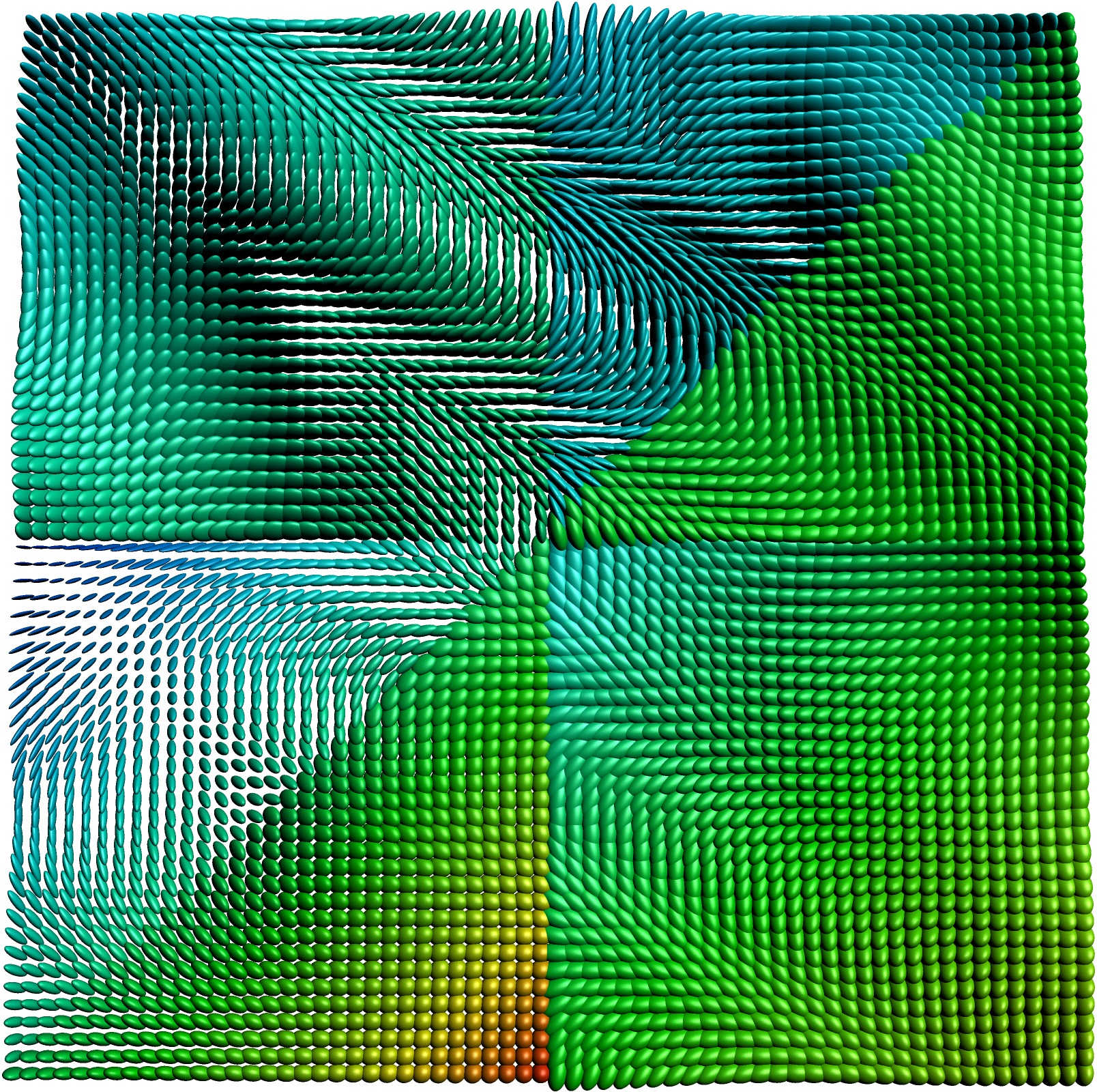}
				\caption[]{Original image}\label{fig:spd3:orig}
			\end{subfigure}
			\begin{subfigure}[t]{0.45\textwidth}
				\centering
				\includegraphics[width = 0.98\textwidth]{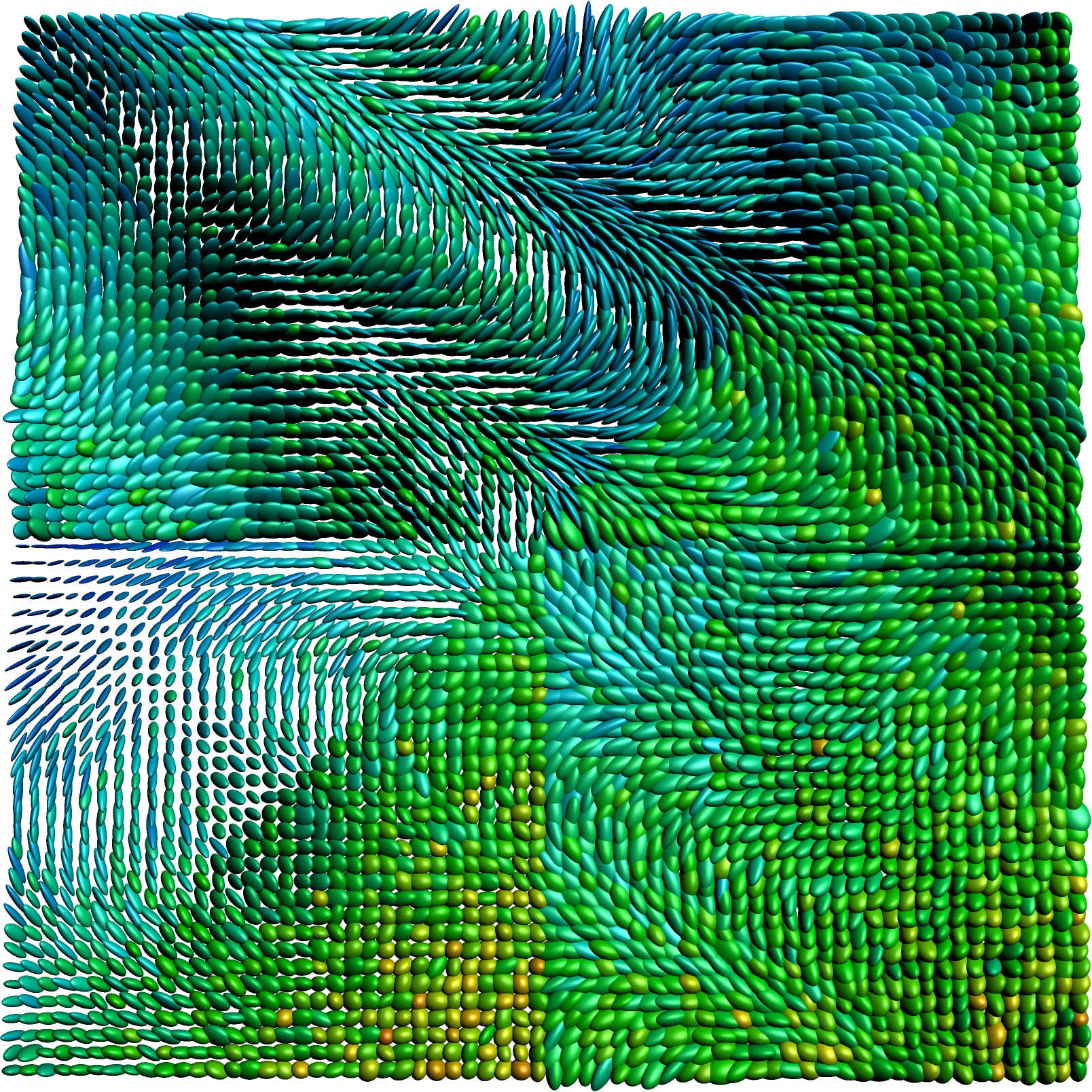}
				\caption[]{Noisy image, $\epsilon = 0.0926$}\label{fig:spd3:noisy}
			\end{subfigure}
			\begin{subfigure}[t]{0.45\textwidth}
				\centering
				\includegraphics[width = 0.98\textwidth]{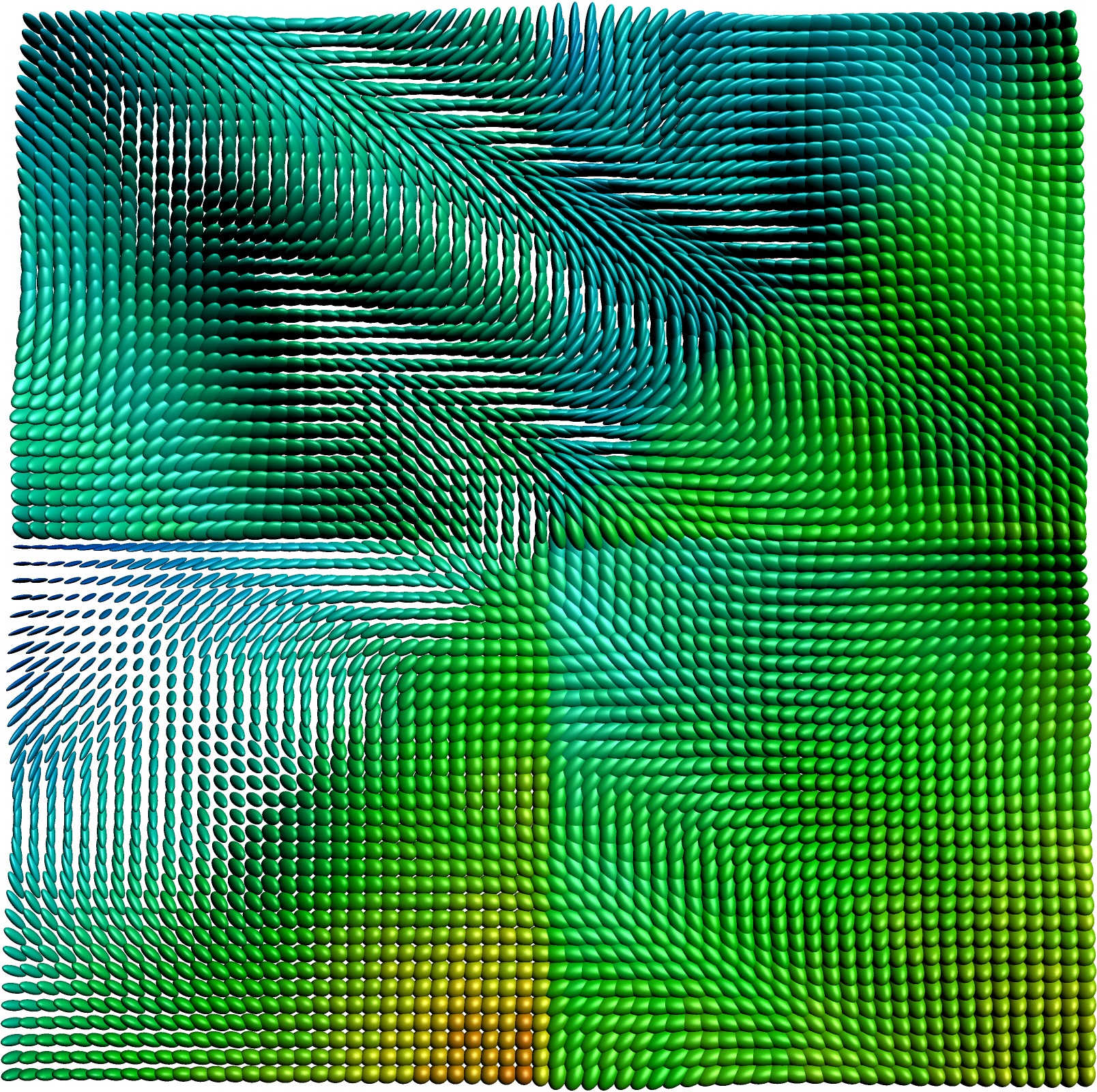}
				\caption[]{NL-MMSE, accelerated, $\epsilon = 0.0100$}\label{fig:spd3:pro}
			\end{subfigure}
			\begin{subfigure}[t]{0.45\textwidth}
				\centering
				\includegraphics[width = 0.98\textwidth]{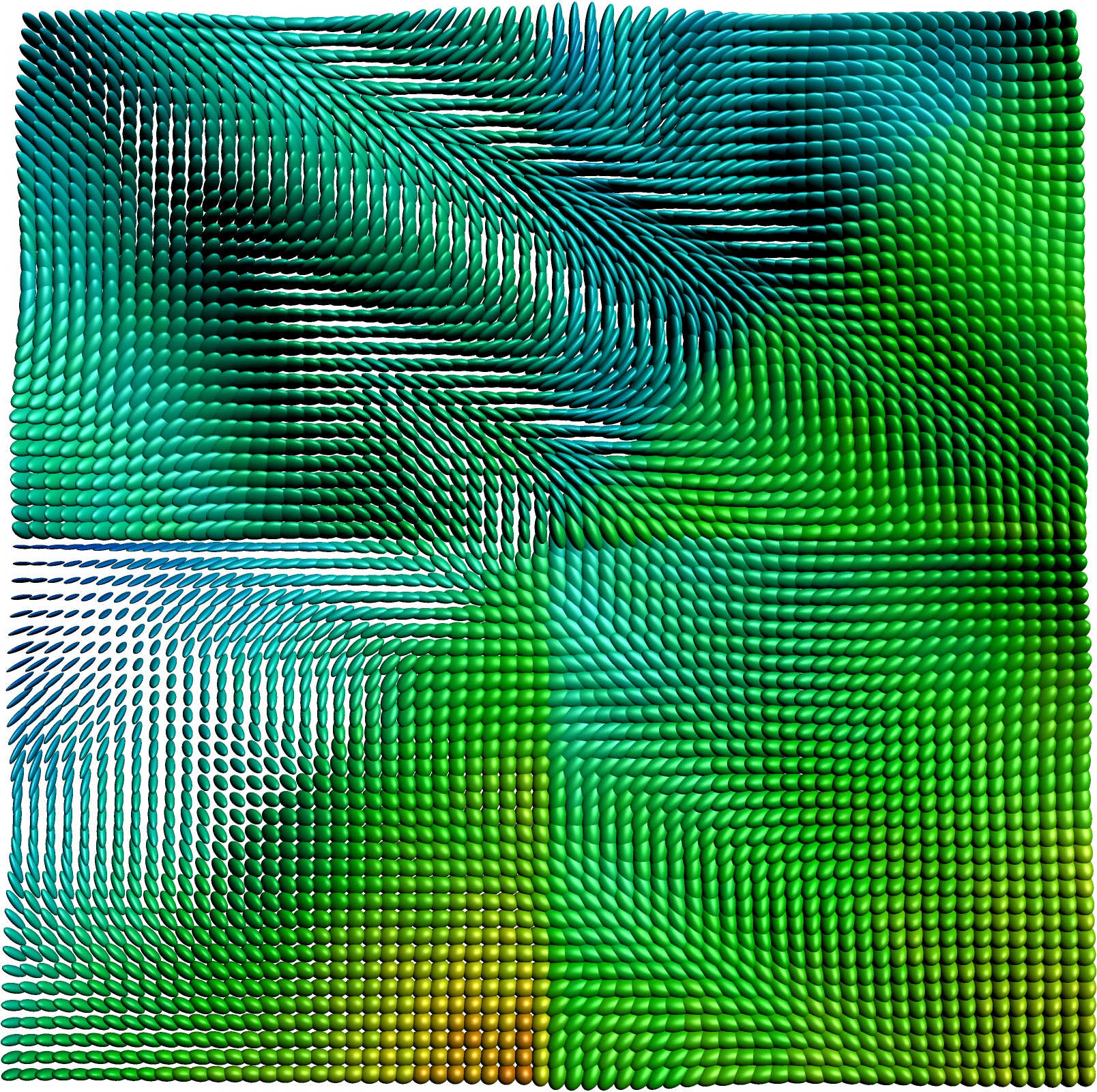}
				\caption[]{NL-MMSE, non accelerated, $\epsilon = 0.0101$}\label{fig:spd3:all}
			\end{subfigure}
			\caption[]{Denoising an image with values in $\SPD(3)$ 
				using NL-MMSE with and without acceleration.}\label{fig:spd3}
		\end{figure}
		
		First we examine the effect of the acceleration used in the NL-MMSE approach. 
		To this aim we consider the $64 \times 64$ image with $\SPD(3)$ values depicted in Figure~\ref{fig:spd3:orig}. 
		The image is corrupted by Gaussian noise of standard deviation $\sigma = 0.125$, see~Figure~\ref{fig:spd3:noisy}. 
		Using NL-MMSE yields Figure~\ref{fig:spd3:pro}.
		Taking all pixels as center of a reference patch, i.e., skipping the acceleration, 
		gives the result in Figure~\ref{fig:spd3:all}. Visually, there is nearly no difference between the two results, 
		and also the errors are roughly the same. However, having a  look at the running time 
		there is  a large difference between the two approaches. The accelerated algorithm needs $245$ seconds 
		and is about one hundred times faster as the non-accelerated version, which needs $40691$ seconds. 
		This justifies the acceleration step.
		
		\begin{figure}[tbp]
			\centering
			\begin{subfigure}[t]{0.32\textwidth}
				\centering
				\includegraphics[width = 0.98\textwidth]{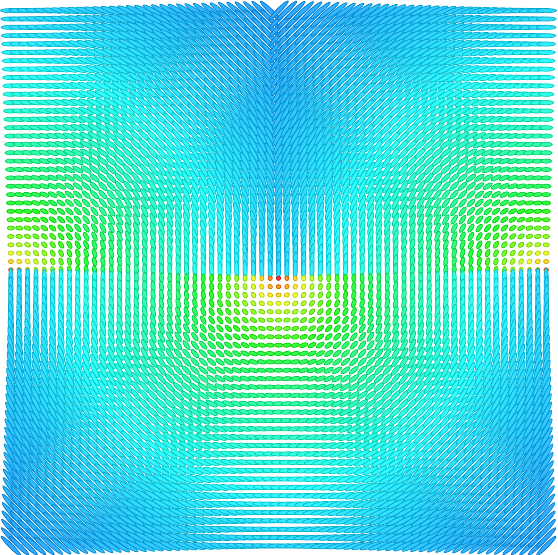}
				\caption[]{Original image}\label{fig:spd2:orig}
			\end{subfigure}
			\begin{subfigure}[t]{0.32\textwidth}
				\centering
				\includegraphics[width = 0.98\textwidth]{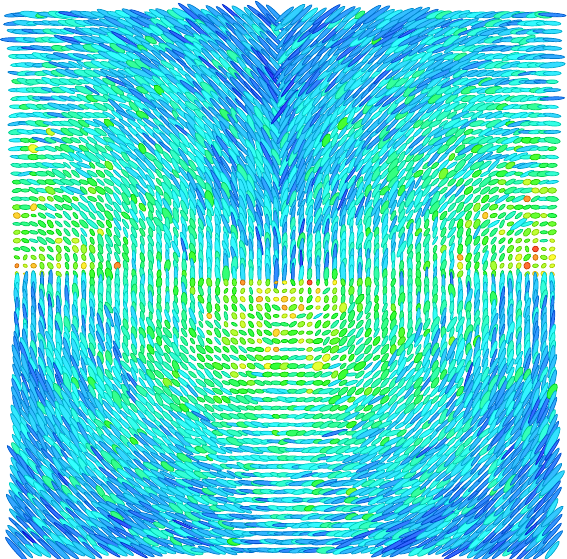}
				\caption[]{Noisy image,\\ $\epsilon = 0.1202$}\label{fig:spd2:noisy}
			\end{subfigure}

			\begin{subfigure}{0.32\textwidth}
				\centering
				\includegraphics[width = 0.98\textwidth]{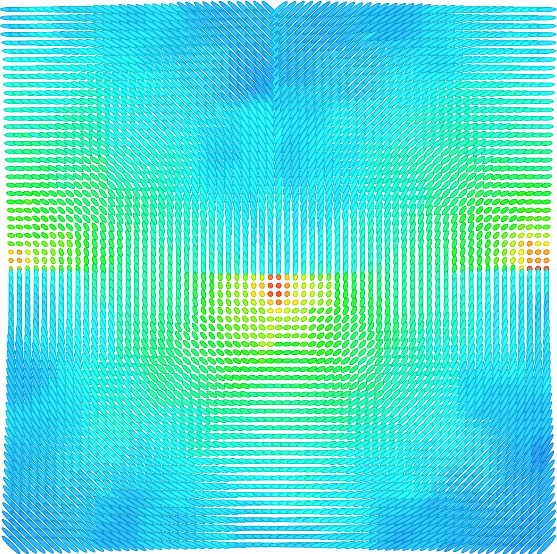}
				\caption[]{TV approach,\\ $\epsilon = 9.5\times 10^{-3}$}\label{fig:spd2:tv}
			\end{subfigure}
			\begin{subfigure}{0.32\textwidth}
				\centering
				\includegraphics[width = 0.98\textwidth]{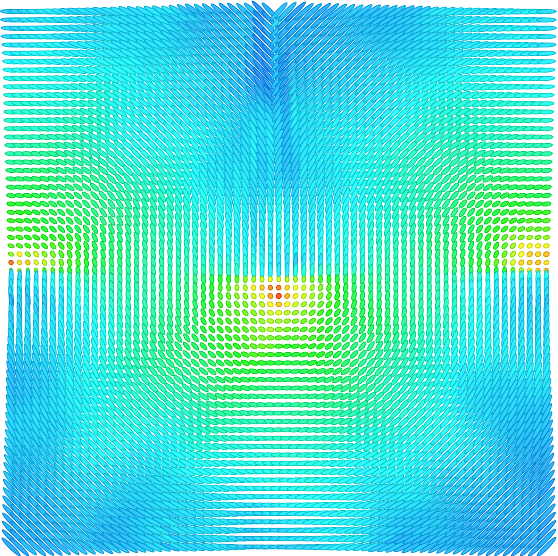}
				\caption[]{NL-means,\\ $\epsilon = 6\times 10^{-3}$}\label{fig:spd2:nl}
			\end{subfigure}
			\begin{subfigure}{0.32\textwidth}
				\centering
				\includegraphics[width = 0.98\textwidth]{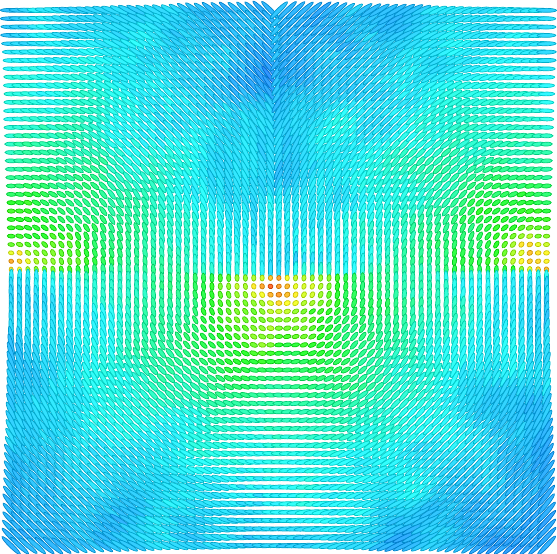}
				\caption[]{NL-MMSE,\\ $\epsilon = 4.3\times 10^{-3}$}\label{fig:spd2:denoised}
			\end{subfigure}
			\caption[]{Comparison of denoising methods for an image with values in $\SPD(2)$.}\label{fig:spd2}
		\end{figure}
		
		Next consider the artificial image of size $65 \times 65$
		consisting of $\SPD(2)$ matrices in~Figure~\ref{fig:spd2:orig} and its
		corrupted version with Gaussian noise of standard deviation $\sigma = 0.15$ in Figure~\ref{fig:spd2:noisy}. 
		In the denoising result with the TV method with parameters $\alpha = 0.25,\ \gamma = 1$ in Figure~\ref{fig:spd2:tv},
		the typical stair casing effect is visible. 
		Figure~\ref{fig:spd2:nl} depicts the result of NL-means using the optimized parameters 
		$s = 33, w = 9,\ \delta = 2,\ K = 81,\ \tau = 0.2$ which looks better than the previous one. 
		However, the NL-MMSE with the same parameters for both steps yields 
		a denoised image with error  $\epsilon = 0.0049$, and
		changing the parameters of the second step to 
		$s_2=7,\ K_2 = 193,\ w_2 = 41$ we finally obtain an error of $\epsilon = 0.0042$, 
		compare~Figure~\ref{fig:spd2:denoised}. This error is a lower than those of the TV and NL-means methods.
		Moreover, this example shows that different parameters in both steps allow a further improvement of the algorithm. 
		In the following examples we optimize the parameters of both steps separately. 

		\subsection{Cyclic Data} \label{subsec:cyclic}
\begin{figure}		
			\centering		
			\captionsetup{justification=centering}
			\begin{subfigure}{0.6\textwidth}
				\centering
				\includegraphics{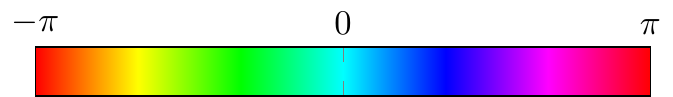}
			\end{subfigure}		
			
			\begin{subfigure}[t]{0.45\textwidth}
				\centering
				\includegraphics{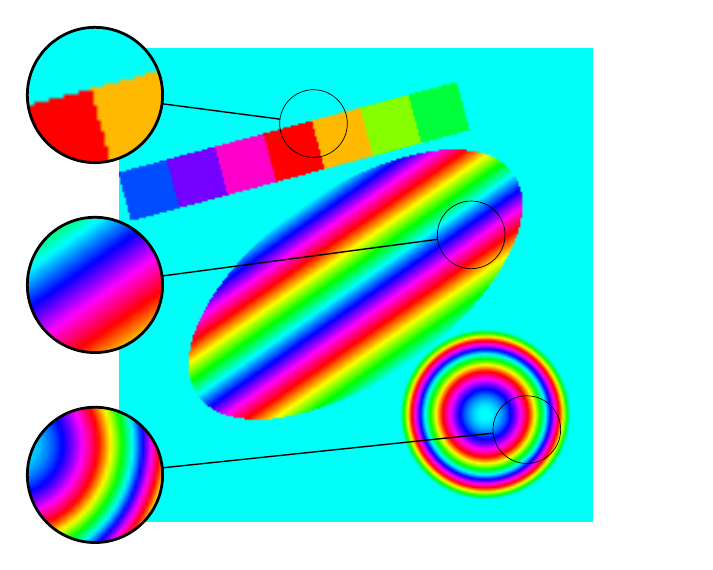}
				\caption[]{Original image}\label{fig:arts1:orig}
			\end{subfigure}	
			\begin{subfigure}[t]{0.45\textwidth}
				\centering
				\includegraphics{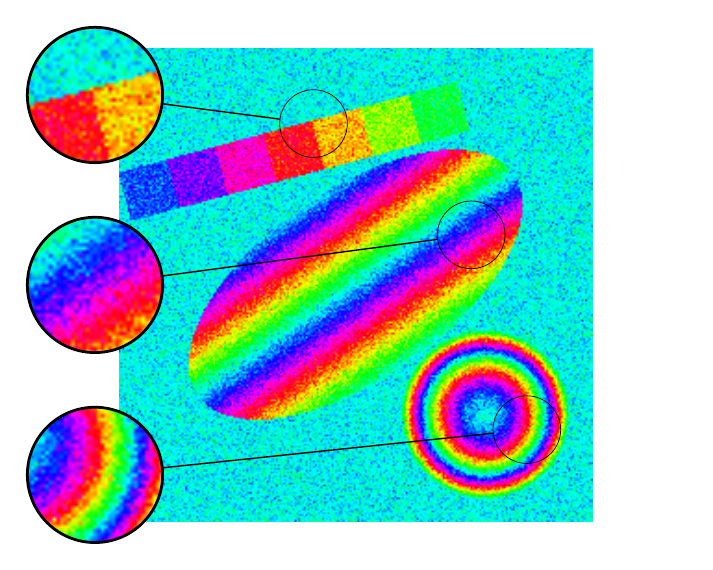}
				\caption[]{Noisy image,\\$\epsilon = 9\times10^{-2}$}\label{fig:arts1:noisy}
			\end{subfigure}
			\vspace{-\baselineskip}
			
			\begin{subfigure}{0.45\textwidth}
				\centering
				\includegraphics{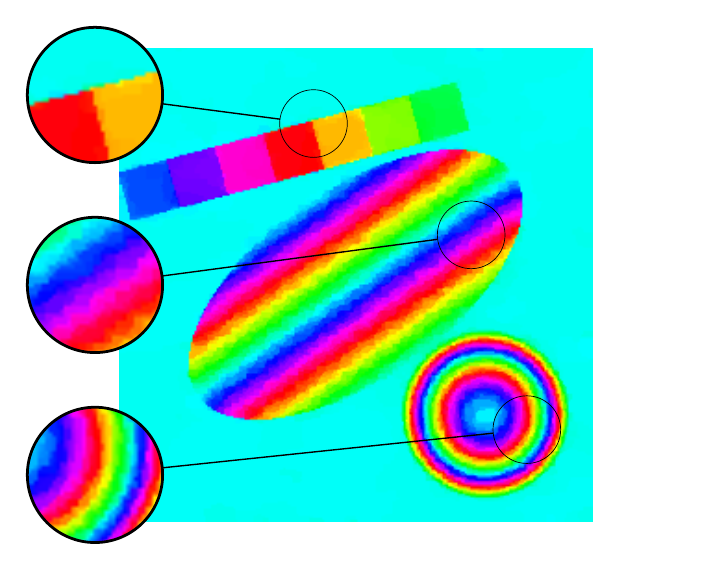}
				\caption[]{TV approach,\\$\epsilon = 7.2\times10^{-3}$}\label{fig:arts1:tv1}
			\end{subfigure}	
			\begin{subfigure}{0.45\textwidth}
				\centering
				\includegraphics{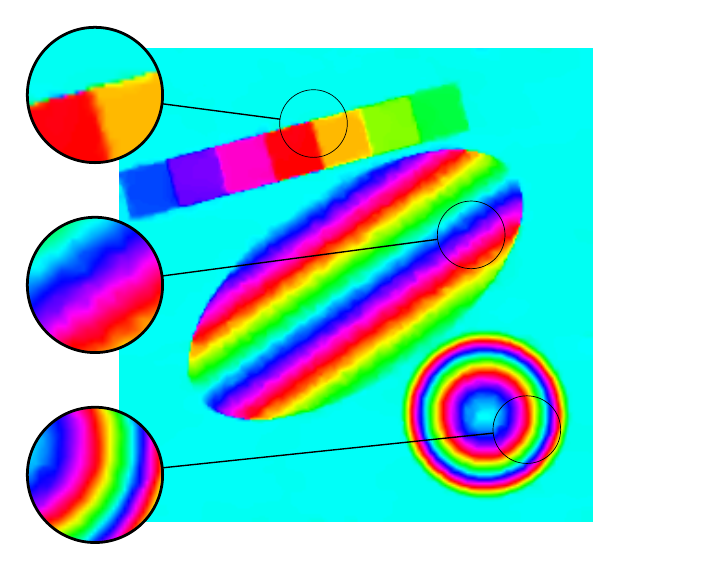}
				\caption[]{TV-$\operatorname{TV}_2$ approach,\\
					$\epsilon = 5.2\times10^{-3} $}\label{fig:arts1:tv12}
			\end{subfigure}		
			\vspace{-\baselineskip}
			
			\begin{subfigure}{0.45\textwidth}
				\centering
				\includegraphics{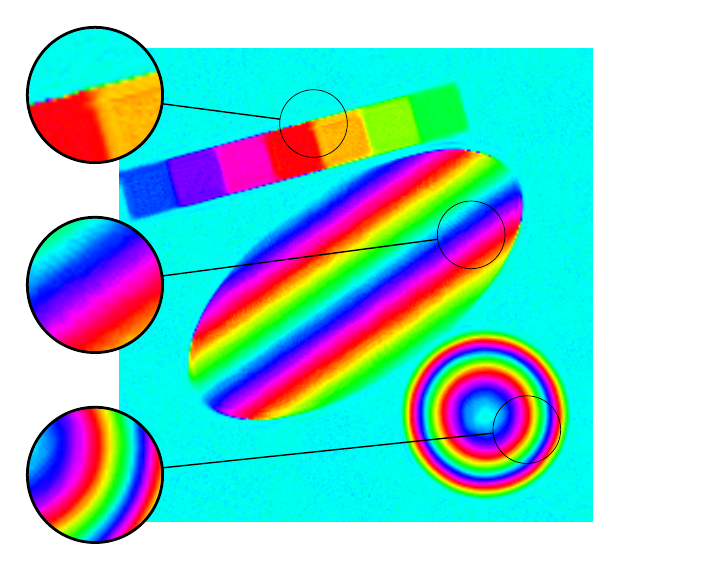}
				\caption[]{NL-means,\\ 
					$\epsilon = 8.1\times10^{-3}$} \label{fig:arts1:nlm}
			\end{subfigure}	
			\begin{subfigure}{0.45\textwidth}
				\centering
				\includegraphics{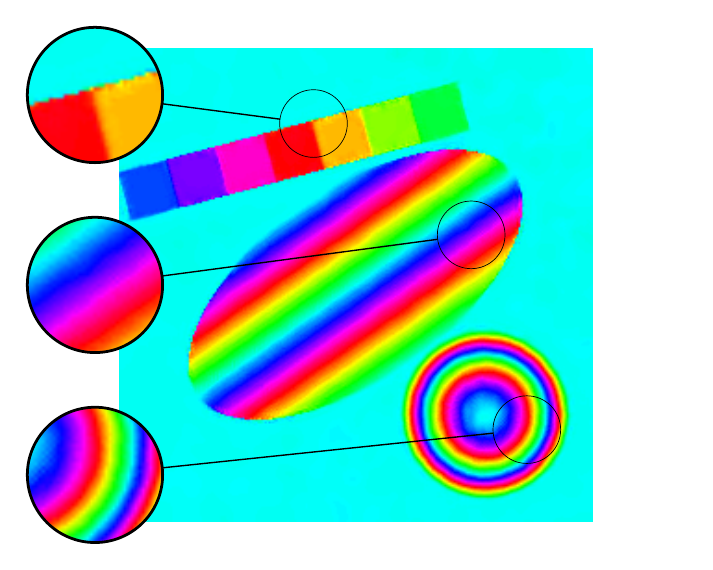}
				\caption[]{NL-MMSE,\\$\epsilon = 2.5\times10^{-3} $}\label{fig:arts1:mmse}
			\end{subfigure}

			\caption[]{Comparison of denoising methods for an image with values in $\mathbb S^1$.}\label{fig:arts1}	
		\end{figure}

		Next we compare the proposed NL-MMSE for the artificial image in~Figure~\ref{fig:arts1:orig}
		and its noisy version corrupt with wrapped Gaussian noise of standard deviation $\sigma = 0.3$ 
		in Figure~\ref{fig:arts1:noisy}. 
		These images as well as their 
		denoised versions via the $\operatorname{TV}$ approach and the
		$\operatorname{TV}$-$\operatorname{TV}_2$ method were taken from~\cite{BLSW14}. 
		The original image can be found in the toolbox
		\href{http://www.mathematik.uni-kl.de/imagepro/members/bergmann/mvirt/}{MVIRT}.
		The TV approach leads to the result in Figure~\ref{fig:arts1:tv1}.
		While the jumps between flat areas are preserved, the method suffers from stair casing. 
		The combined first and second order approach in Figure~\ref{fig:arts1:tv12} improves the results,
		but the edges between flat areas are smoothed. 
		Not surprisingly, the result obtained with the NL-means approach in Figure~\ref{fig:arts1:nlm} with parameters
		$s = 11, w = 23,\ \delta = 46,\ K = 33,\ \tau = 0.2$ has the worst error, even if the reconstructions 
		of the paraboloid in the bottom right and at the edges are pretty good. Here an extra fitting for 
		constant regions as incorporated in the fine tuning of  NL-MMSE, would be necessary. 
		The best result, shown in~Figure~\ref{fig:arts1:mmse}, is achieved with the NL-MMSE, see~Figure~\ref{fig:arts1:mmse}.
		On the one hand, sharp edges are preserved, while one the other hand also constant and linear parts are well reconstructed. 
		
		\subsection{Spherical Data}\label{subsec:spherical}
		\begin{figure}[tbp]
			\centering
			\begin{subfigure}{0.32\textwidth}
				\centering
				\includegraphics[width = 0.98\textwidth]{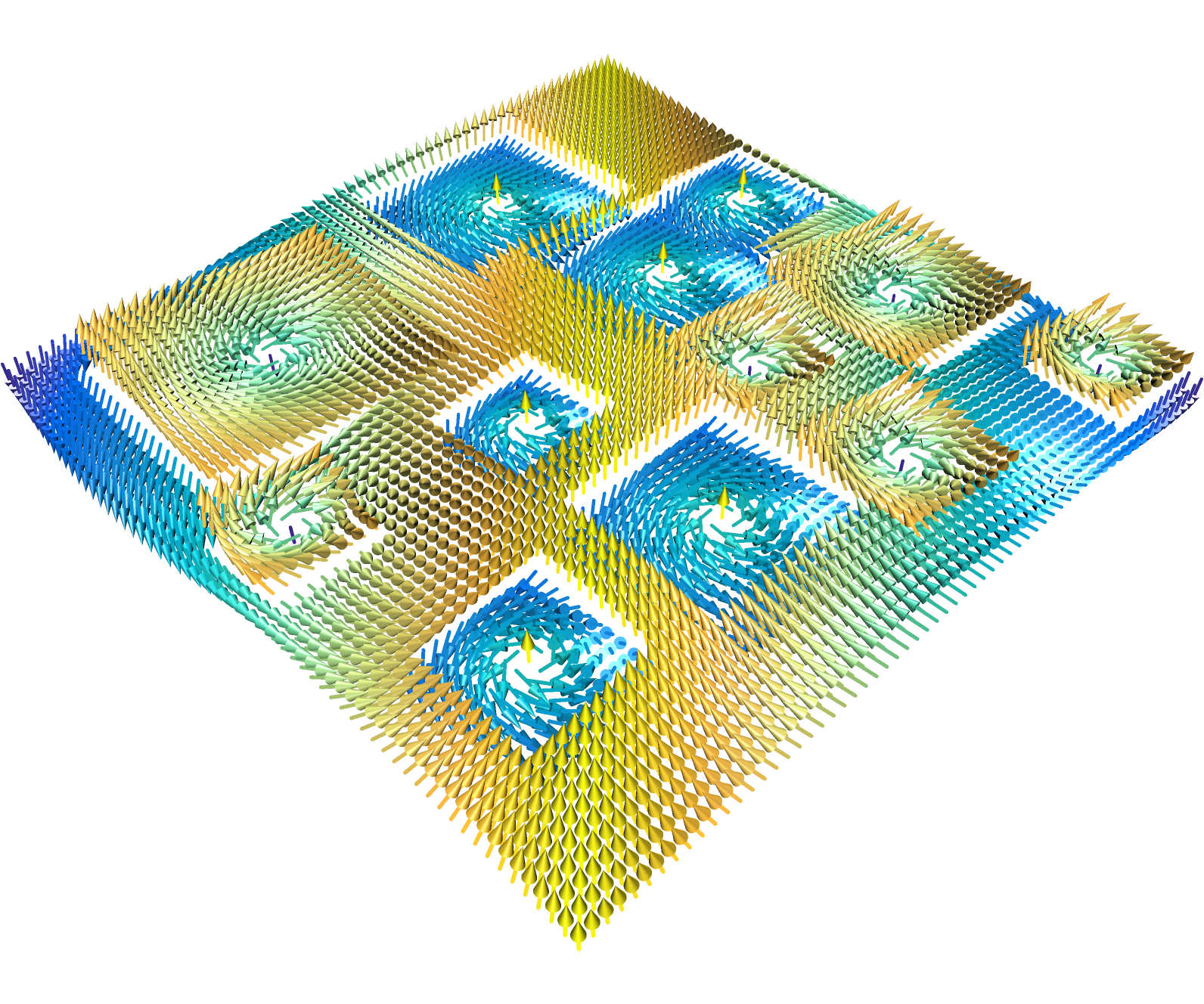}
				\caption[]{Original image}\label{fig:arts2:orig}
			\end{subfigure}	
			\begin{subfigure}{0.32\textwidth}
				\centering
				\includegraphics[width = 0.98\textwidth]{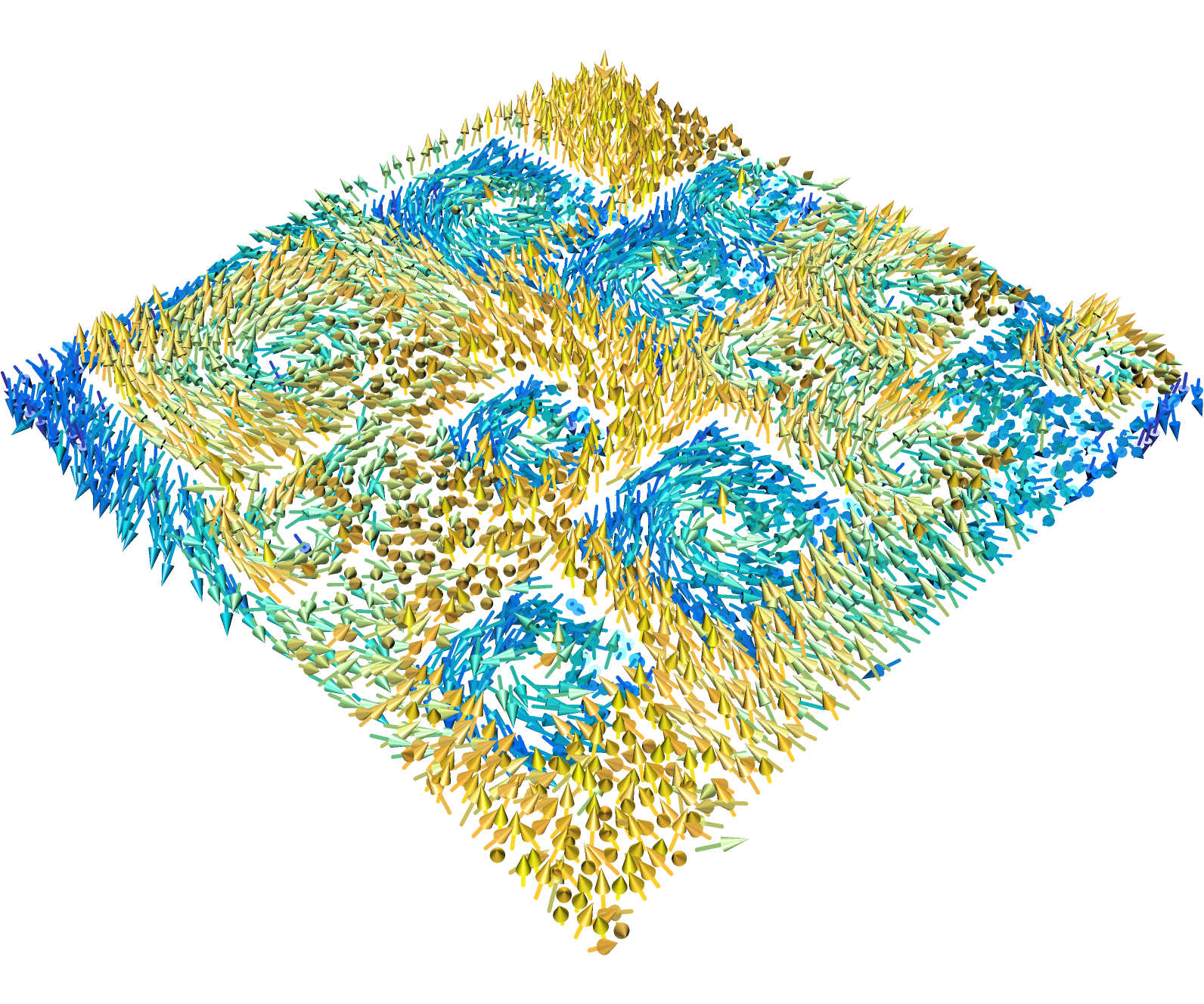}
				\caption[]{Noisy image, $\epsilon = 0.1767$}\label{fig:arts2:noisy}
			\end{subfigure}	
			
			\begin{subfigure}{0.32\textwidth}
				\centering
				\includegraphics[width = 0.98\textwidth]{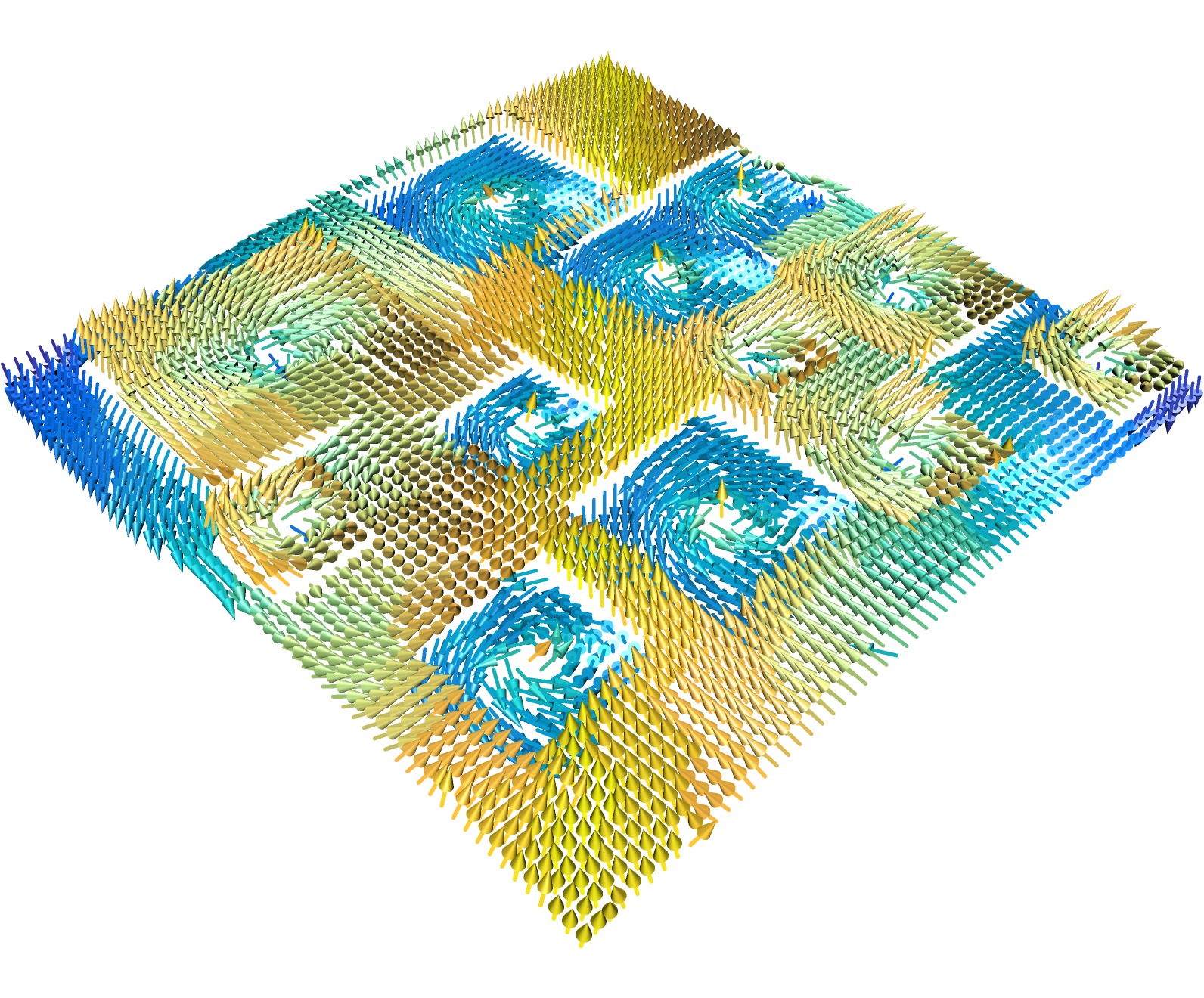}
				\caption[]{TV approach, $\epsilon = 0.0352$}\label{fig:arts2:tv1}
			\end{subfigure}	
			\begin{subfigure}{0.32\textwidth}
				\centering
				\includegraphics[width = 0.98\textwidth]{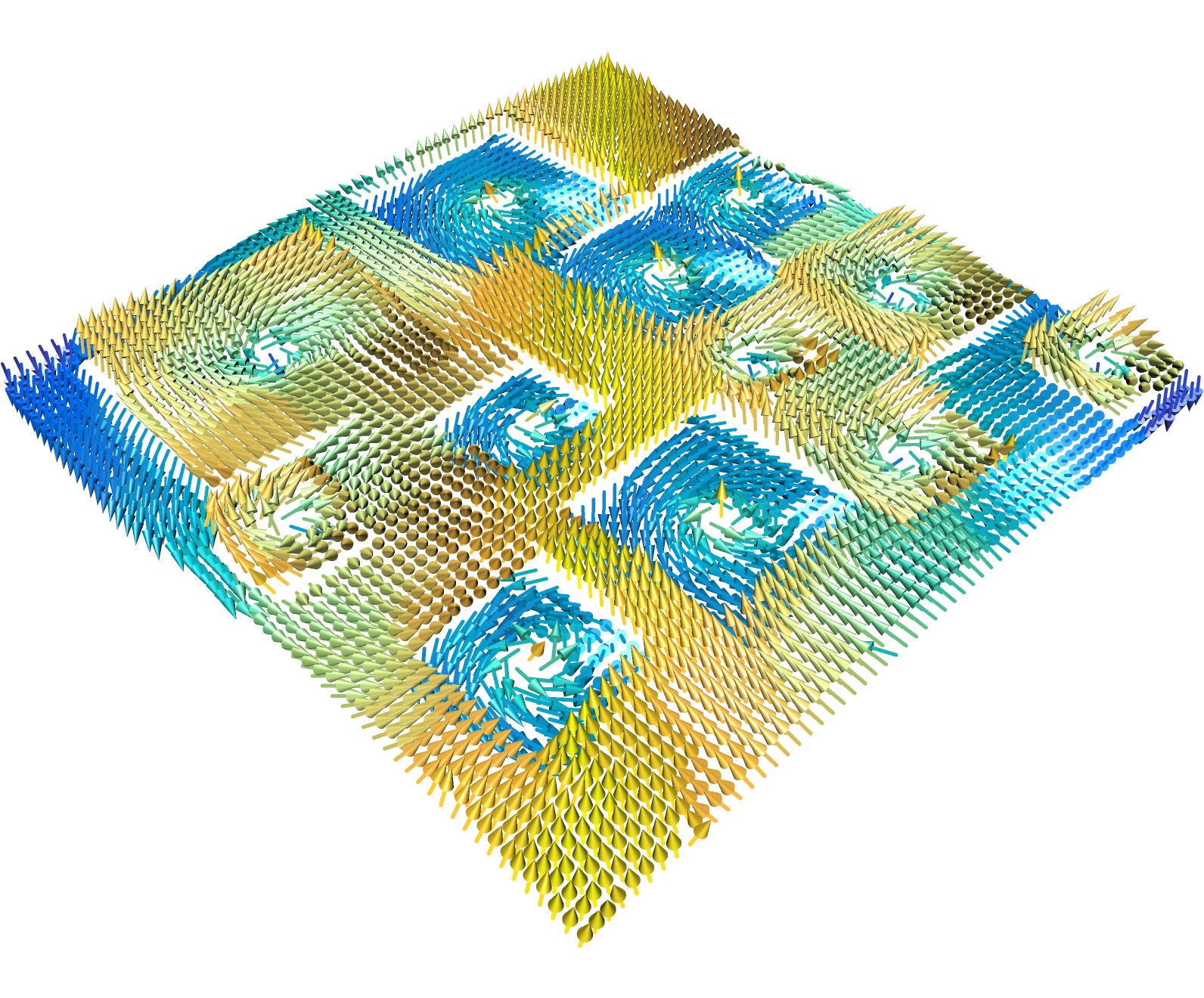}
				\caption[]{TV-$\operatorname{TV}_2$ approach, $\epsilon = 0.0338$}\label{fig:arts2:tv12}
			\end{subfigure}	
			\begin{subfigure}{0.32\textwidth}
				\centering
				\includegraphics[width = 0.98\textwidth]{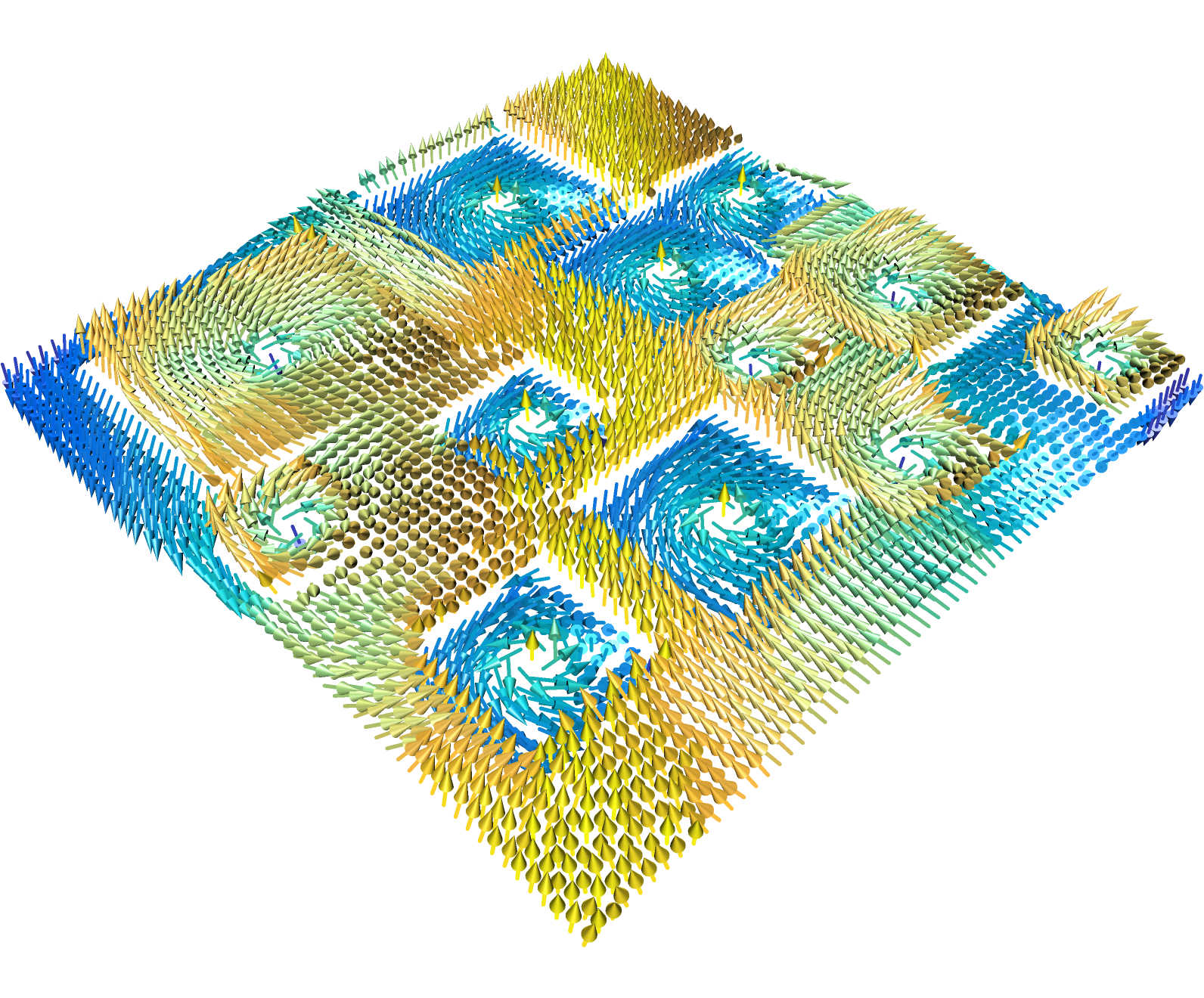}
				\caption[]{NL-means, $\epsilon = 0.0326$}\label{fig:arts2:nl_mean}
			\end{subfigure}
			
			\begin{subfigure}{0.32\textwidth}
				\centering
				\includegraphics[width = 0.98\textwidth]{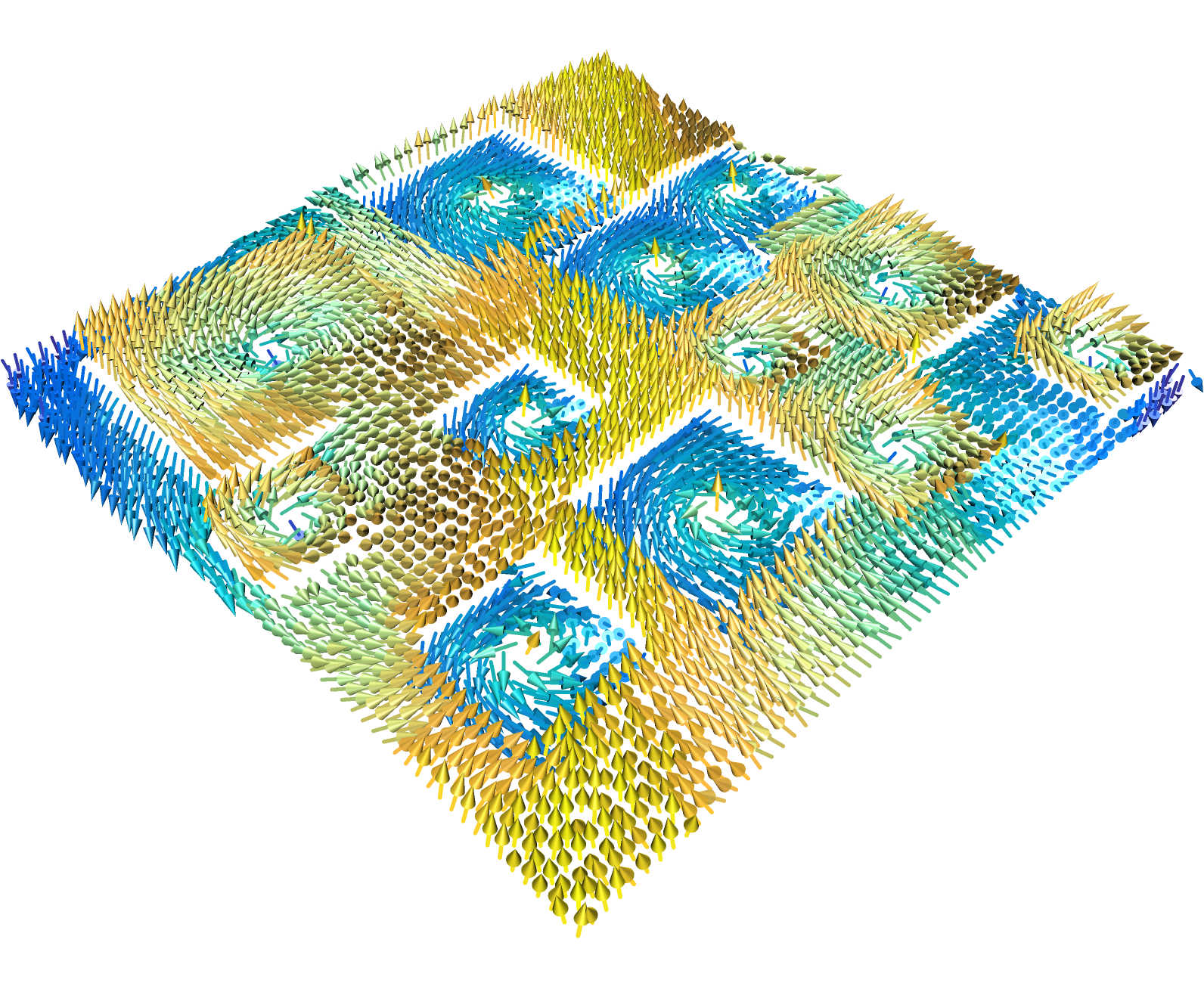}
				\caption[]{Oracle image of Algorithm~\ref{Alg:NL_M}, $\epsilon = 0.0312$}\label{fig:arts2:oracle}
			\end{subfigure}
			\begin{subfigure}{0.32\textwidth}
				\centering
				\includegraphics[width = 0.98\textwidth]{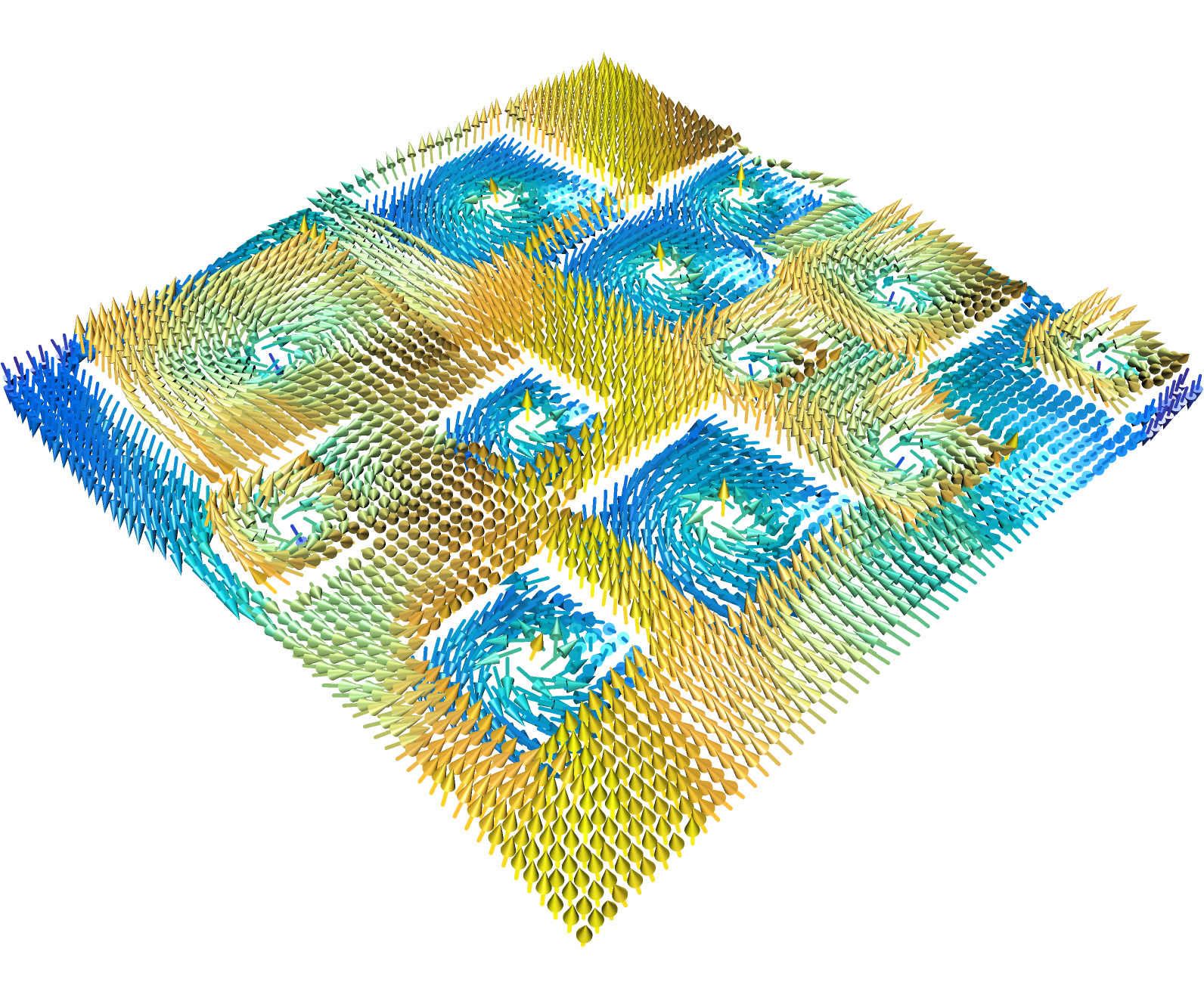}
				\caption[]{NL-MMSE, $\epsilon = 0.0258$}\label{fig:arts2:final}
			\end{subfigure}
			\hspace{0.2cm}
			\begin{subfigure}{0.32\textwidth}
				\centering
				\includegraphics[width = 0.98\textwidth]{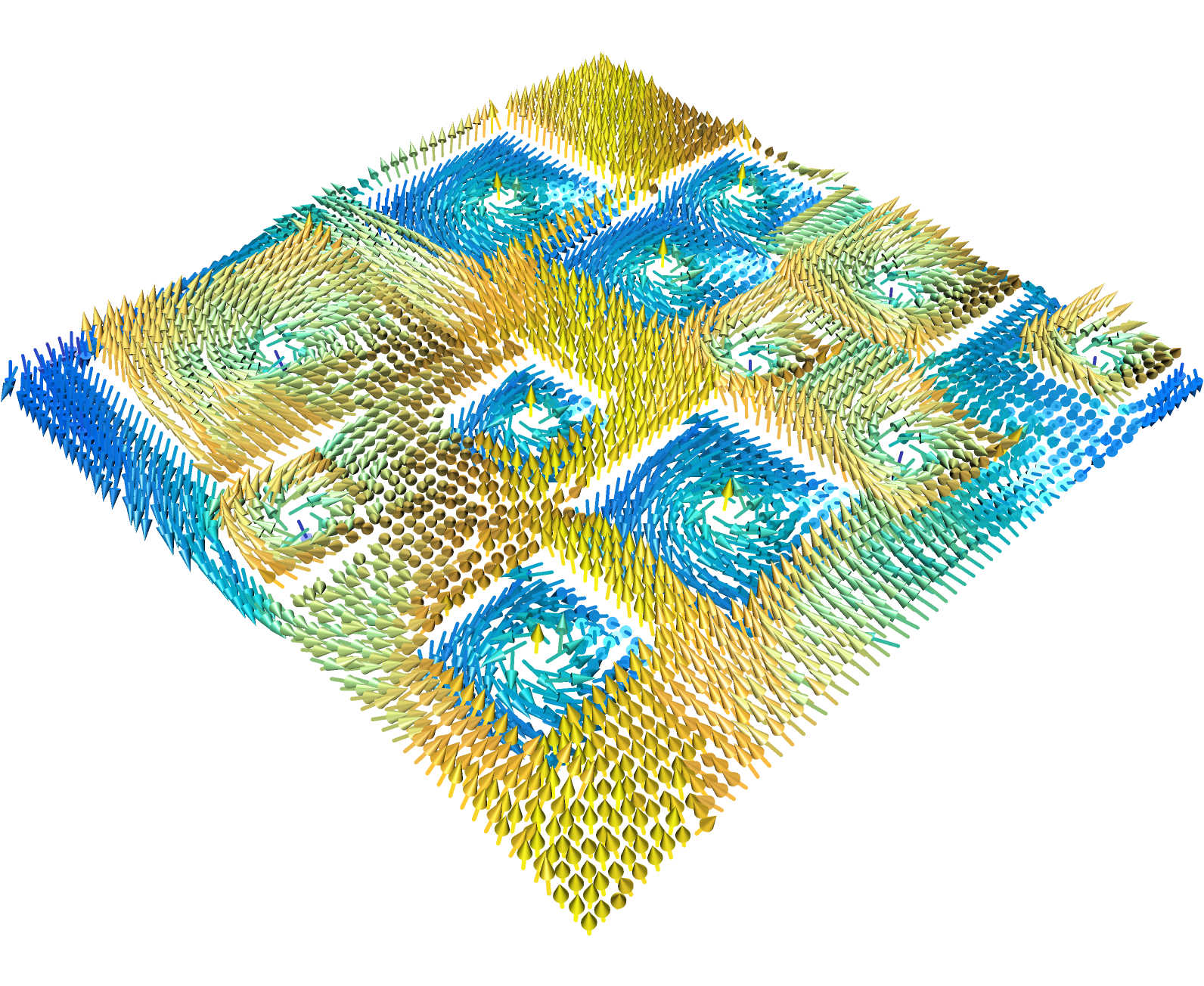}
				\caption[]{Denoised image without update, $\epsilon = 0.0335$}\label{fig:arts2:patch}
			\end{subfigure}		
			
			\caption[]{Comparison of denoising methods of an $\SP^2$-valued image.}\label{fig:arts2}
		\end{figure}
		
		Finally we consider the artificial image with values distributed over the whole sphere as shown
		in~Figure~\ref{fig:arts2:orig}. 
		The image consists of vortex like structures of different sizes and directions
		and a smoothly varying background. 
		It is affected by Gaussian noise with standard deviation $\sigma=0.3$, 
		see Figure~\ref{fig:arts2:noisy}. 
		We compare our method with the TV approach (parameters:  $\alpha=0.24,\ \gamma=\tfrac{\pi}{2}$) 
		in Figure~\ref{fig:arts2:tv1}, TV-$\operatorname{TV}_2$ (parameters:  $\alpha=0.18,\ \beta = 2.6,\ \gamma=\tfrac{\pi}{2}$) 
		in Figure~\ref{fig:arts2:tv12},
		and NL-means (parameters: $\epsilon$, 
		$s = 23, w = 127,\ \delta = 1.5,\ K = 104,\ \tau = 0.2$) in Figure~\ref{fig:arts2:nl_mean}.
		Note that the running time of the NL-mean is the same as the NL-MMSE in this example, i.e., 20 seconds, while the TV-$\operatorname{TV}_2$ 
		method needs around  minutes. The first order TV suffers from stair casing, which is removed with the second order term, 
		but the error is still the largest among all tested methods.
		Next we have a look at the oracle image after Step 1 of NL-MMSE in  Figure~\ref{fig:arts2:oracle}. 
		We see that this image has already a slightly smaller error than both TV and the  NL-means approaches. 
		However, it still contains  some noise in the background, 
		which is removed in the second step and leads to an improvement in the error, see~Figure~\ref{fig:arts2:final}.

		Figure~\ref{fig:arts2:patch}  shows the reconstruction 
		with Algorithm~\ref{Alg:NL_M} without the second order-update step, 
		i.e., we perform only Step 1, where we 
		replace $\hat{\bm y}_j = \exp_{\hat{\bm \mu}_i}\bigl(\hat{\Sigma}_i (\hat{\Sigma}_i + \sigma^2 I_{s_1^2})^{-1} \log_{\hat{\bm \mu}_i}(y_j)\bigr)
		$
		with
		$
		\hat{\bm y}_j = \hat{\bm \mu}_i
		$.
		The parameters are $s = 5,\ K = 6$ and $w,\gamma$ from before. 
		In comparison to the oracle image there are small visible differences, but the error is worse. 
		A disadvantage of this method is its running time. While the oracle image computation
		needs less then 10 seconds, about 30 seconds are required to get the result in Figure~\ref{fig:arts2:patch}.
		The time difference originates from the larger patch size which is needed to get a comparable result.
		
		
		This  experiment further shows that Algorithm~\ref{Alg:NL_M} is also able to handle data having values on the whole sphere
		which is a manifold with positive curvature. Here, we implicitly assume that the computed Karcher means are unique, 
		which is a reasonable assumption, since similar patches should be pointwise contained in regular balls. 
		Note that this does not prevent the patches to cover the whole sphere.
		
		
		\section{Conclusion and Future Work}\label{sec:conclusions}
		We  proposed a counterpart of the nonlocal Bayes'  denoising approach of Lebrun et al.~\cite{LBM13b,LBM13} for manifold-valued images. 
		The basic idea consists in translating the MMSE for similar image patches
		\eqref{3a} to the manifold-valued setting \eqref{MMSE_manifold}. To this aim, we used an intrinsic definition of a normal distribution  
		and in particular of white noise on Riemannian manifolds. 
		We demonstrated by various numerical experiments that our method performs very well when dealing with moderate noise variances. 
		
		Up to now all our examples were artificial ones. 
		In future work we want to apply our method to real-world data.
		In particular 
		we intend to examine whether our noise model covers specific applications.
		The close relation between different models of Gaussian noise for small $\sigma$ should be specified
		for the manifolds of interest.
		Moreover, it is well known that in various applications the variance~$\sigma^2$ is either not  known or not constant for the whole image.
		Therefore the noise estimation and the incorporation of spatially varying noise is an interesting research topic.
		
		Another issue is related to Remark~\ref{neg_cov}. 
		Even in the Euclidean setting the topic of negative eigenvalues 
		in the estimation of $\Sigma_X$ requires further discussion.
		Other directions of future work include other image restoration tasks as for instance inpainting. 
		This needs additional information, e.g., based on hyperpriors as in~\cite{AADGM2015} 
		or a fixed number of Gaussian models (covariance matrices), see e.g.~\cite{YSM2012}.
		
		\appendix
		\section{Proof of Proposition \ref{1d_manifolds} } \label{app:prop}
		\begin{proof} 
			For one-dimensional manifolds we have $x = x^1$ and
			$|G(x)| = |\tilde G(\bm x)| =1$. In the following we set $e_{\bm \mu} \coloneqq e_{\bm \mu,1}$.
			
			(i) With Appendix \ref{app:ex_mani} we obtain 
			$\dist_{{\rm SPD}(1)}({\bm \mu} , {\bm x}) = \left|\ln\left( \frac{\bm \mu}{\bm x} \right) \right|$
			so that with \eqref{stand_normal} we obtain the pdf stated in (i). 
			By \eqref{transform} we have
			$$
			\frac{1}{ \sqrt{2 \pi\sigma^2} } \int_{\R} \e^{-\frac{1}{2\sigma^2} x^2} \, \dx x
			=
			\frac{1}{ \sqrt{2 \pi\sigma^2} } \int_{\R_{>0}} \e^{-\frac{1}{2\sigma^2} ( \ln (\bm x) - \ln(\bm \mu))^2} \, \dx_{\R_{>0}} (\bm x)
			$$
			which  implies by the transformation theorem that $\dx_{\R_{>0}} (\bm x) = \frac{1}{\bm x} \, \dx  \bm x$.
			\\
			
			(ii)
			In  the given parameterization it holds that
			$e_{\bm \mu} = (-\sin (t_\mu), \cos (t_\mu))^\tT$, $t_\mu \in [-\pi,\pi)$
			and with Appendix \ref{app:ex_mani}
			further 
			$$
			\exp_{\bm \mu} (h(x)) = \exp_{\bm \mu} ( x e_{\bm \mu} )
			= \bm \mu \cos (|x|) +  \frac{x}{|x|} \sin (|x|)e_{\bm \mu} = 
			\begin{pmatrix}
			\cos (x+ t_\mu)\\
			\sin (x+ t_\mu)
			\end{pmatrix}.
			$$ 
			It holds
			${\cal D}_{\bm \mu} = (-\pi,\pi)$ 
			and we can choose 
			$\varphi_j\colon \left( (2j-1) \pi , (2j+1) \pi \right) \rightarrow (-\pi,\pi)$
			as $\varphi_j(x) \coloneqq x - 2j\pi$, $j \in \Z$ in \eqref{stand_normal_sphere}.
			Plugging this into \eqref{stand_normal} results in
			\begin{align}
			p_{\bm X} (\bm x (t)) 
			&= 
			\frac{1}{\sqrt{2 \pi \sigma^2}} \sum_{j \in \Z} \e^{-\frac{1}{2 \sigma^2} \left(  h^{-1}  (\log_{\bm \mu} (\bm x(t)))  + 2j\pi \right)^2}
			= 
			\frac{1}{\sqrt{2 \pi \sigma^2}} \sum_{j \in \Z} \e^{-\frac{1}{2 \sigma^2} \left(  {\rm d}_{\mathbb S^1} (\bm \mu , \bm x(t))  + 2j \pi \right)^2 }\\
			&=
			\frac{1}{\sqrt{2 \pi \sigma^2}} \sum_{j \in \Z} \e^{-\frac{1}{2 \sigma^2} \left( t - t_\mu  + 2j\pi \right)^2}.
			\end{align}
			
			(iii) 
			First note that $\Delta_1$ is not complete, but for any $\bm \mu$
			the function $\exp_{\bm \mu}$  is defined  a.e. on $T_{\bm \mu}\Delta_1$.
			More precisely, using Appendix \ref{app:ex_mani} we obtain 
			$
			e_{\bm \mu} =\tfrac{1}{2} \sin (t_\mu) (1,-1)^\tT
			$
			and
			$$
			\exp_{\bm \mu} (h(x)) 
			= \frac12 + \frac12 \begin{pmatrix} \cos(t_\mu)\\-\cos(t_\mu) \end{pmatrix} \cos (x) + 
			\frac12 \begin{pmatrix} \sin(t_\mu)\\-\sin(t_\mu) \end{pmatrix} \sin (x)
			= \frac12 + \frac12 \begin{pmatrix} 
			\cos(x- t_\mu)\\
			-\cos(x- t_\mu) \end{pmatrix},
			$$
			which is only in $\Delta_1$ if $ x \not \in \{t_\mu + j \pi: j \in \mathbb Z\}$. 
			Here have
			${\cal D}_{\bm \mu} = (t_\mu-\pi,t_\mu)$ and
			setting
			\begin{align}
			\varphi_{2j} (x) &\coloneqq x - 2 j \pi  \quad {\rm for} \quad x \in (t_\mu + (2j-1)\pi, t_\mu + 2j \pi),\\
			\varphi_{2j + 1} (x) &\coloneqq 2t_\mu- (x - 2j \pi ) \quad {\rm for} \quad x \in (t_\mu + 2j \pi, t_\mu + (2j+1)\pi)
			\end{align}
			we obtain by \eqref{stand_normal_sphere} that
			$$
			\tilde p_{X} (x) = \frac{1}{\sqrt{2 \pi \sigma^2}} \sum_{j \in \Z} 
			\left( 
			\e^{-\frac{1}{2 \sigma^2} \left( x + 2j\pi \right)^2}
			+
			\e^{-\frac{1}{2 \sigma^2} \left( 2t_\mu - x + 2j\pi \right)^2}
			\right).
			$$
			With \eqref{stand_normal} we obtain the assertion.
		\end{proof}
		\section{Example Manifolds}\label{app:ex_mani}
		\paragraph{Sphere $\SP^{d}$}
		Let $\SP^{d} = \bigl \{ \bm x\in \R^{d+1}\colon \norm{\bm x}{2}=1\bigr\}$.
		The geodesic distance is given by
		\begin{equation*}
		\dist_{\SP^{d}}(\bm x, \bm y) = \arccos(\langle \bm x, \bm y\rangle),
		\end{equation*}
		where $\langle\cdot,\cdot\rangle$ is the standard scalar product in $\R^{d+1}$.
		The tangential space at $\bm x\in\SP^{d}$ is given by 
		$T_{\bm x} \SP^{d}=\bigl\{v\in\R^{d+1}\vert \langle {\bm x} ,v\rangle=0\bigr\}$. 
		The Riemannian metric is the metric from the embedding space, i.e., the Euclidean
		inner product. 
		The exponential and logarithmic map read as
		\begin{align}
		\exp_{\bm x} (v) &= {\bm x} \cos\bigl(\lVert v\rVert\bigr)+\frac{v}{\lVert v\rVert}\sin\bigl(\lVert v\rVert\bigr),\\
		\log_{\bm x} (\bm y) &=  \dist_{\SP^{d}}(\bm x, \bm y) \, \frac{\bm y-\langle {\bm x},{\bm y} 
			\rangle \bm x}{\lVert \bm y-\langle \bm x, \bm y \rangle \bm x\rVert}, 
		\quad \bm x \not = - \bm y.
		\end{align}
		
		\paragraph{Positive Definite Matrices  ${\SPD}(r)$}
		The dimension of ${\SPD}(r)$ is $d = \frac{r(r+1)}{2}$.
		We denote by $\Exp$ and $\Log$ the matrix exponential and logarithm defined by
		\\
		\(
		{\Exp (x) \coloneqq \sum_{k=0}^\infty \frac{1}{k!} x^k}
		\)
		and
		$\Log (x) \coloneqq -\sum_{k=1}^\infty \frac{1}{k} (I-x)^k$, 
		$\rho(I-x) < 1$,
		where $\rho$ denotes the spectral radius.
		Then the affine invariant geodesic distance is given by
		\[
		\dist_{{\SPD}(r)}(\bm x,\bm y) = \bigl\lVert\Log(\bm x^{-\frac12} \bm y \bm x^{-\frac12} )\bigr\rVert_{\mathrm{F}},
		\]
		where $\lVert \cdot \rVert_{\mathrm{F}}$ denotes the Frobenius norm of matrices.
		The tangential space at $\bm x \in \MM$ is  $T_{\bm x}\MM = \{\bm x\} \times \operatorname{Sym}(r)$, 
		where  $\operatorname{Sym}$ denotes the space of
		symmetric $r \times r$ matrices. The Riemannian metric reads
		$\langle v_1,v_2 \rangle_{\bm x} = \tr (v_1 \bm x^{-1} v_2 \bm x^{-1})$.
		As orthogonal basis in $T_{\bm x} \MM$ we use
		$e_{\bm x,ij}  \coloneqq \bm x^\frac12 e_{ij} \bm x^\frac12$, $i,j \in \{1,\ldots,r\}, \, j\le i$,  where
		$$
		e_{ij} =
		\left\{
		\begin{array}{ll}
		e_i e_i^\tT&\mathrm{if} \; i=j,\\
		\frac{1}{\sqrt{2}} \bigl(e_i e_j^\tT+e_j e_i^\tT\bigr)&\mathrm{otherwise} 
		\end{array}
		\right.
		$$
		and $e_i \in \mathbb R^r$ are the $r$-dimensional unit vectors.
		Finally, the exponential and the logarithmic map read
		\begin{align} \label{exp_spd}
		\exp_{\bm x} (v) &= \bm x^{\frac12} \Exp\bigl( \bm x^{-\frac12} v \bm x^{-\frac12}\bigr) \bm x^{\frac12},\\
		\log_{\bm x}(\bm y) &= \bm x^{\frac{1}{2}}\Log\bigl(\bm x^{-\frac{1}{2}} \, \bm y \, \bm x^{-\frac{1}{2}}\bigr) 
		\bm x^{\frac{1}{2}}.
		\end{align}
		For more information on the affine invariant metric and its relation to the log-Euclidean metric we refer, e.g., to
		\cite{AFPA2005,pennec2006riemannian},
		
		\paragraph{Probability Simplex  $\Delta_d$}
		In the open probability simplex 
		$\Delta_{d} \coloneqq \{\bm x\in \R_{> 0}^{d+1}: \sum_{i=1}^{d+1} x_i= 1\}$ 
		equipped with the Fisher-Rao metric arising from the categorial distribution
		$
		\langle u,v\rangle_{\bm x} = \langle\tfrac{u}{\sqrt{\bm x}},\tfrac{v}{\sqrt{\bm x}}\rangle
		$
		the geodesic distance is given by
		\begin{equation}
		\dist_{\Delta_d} (\bm x,\bm y) = 2\arccos\bigl(\langle\sqrt{\bm x},\sqrt{\bm y}\rangle\bigr),
		\end{equation}
		where the square root is meant componentwise.
		Its tangential space is given by 
		$
		T_{\bm x} \MM = \{y\in\R^{d+1}: \langle y,\vecOne\rangle = 0\}
		$.
		The exponential map reads 
		\begin{equation}
		\exp_{\bm x} (v) = \frac{1}{2}\Bigl(\bm x+\frac{v_x^2}{\lVert v_x\rVert_2^2}\Bigr)+
		\frac{1}{2}\Bigl(\bm x-\frac{v_x^2}{\lVert v_x\rVert_2^2}\Bigr)\cos\bigl(\lVert v_x\rVert_2\bigr)
		+\frac{v}{\lVert v_x\rVert_2}\sin\bigl(\lVert v_x\rVert_2\bigr),
		\end{equation}
		where $v_x \coloneqq \tfrac{v}{\sqrt{\bm x}}$ and vector multiplications are meant componentwise.
		While the above function maps onto the closure of $\Delta_d$ we have to consider
		only the dense set in $T_{\bm x}\Delta_d$ with $\exp_{\bm x} (v) \in \Delta_d$.
		The logarithmic map is determined by
		\begin{equation}
		\log_{\bm x} (\bm  y )= \dist_{\Delta_d} (\bm x, \bm y)
		\,
		\frac{
			\sqrt{\bm x \bm y}-\langle \sqrt{\bm x},\sqrt{\bm y}\rangle {\bm x} }{ \sqrt{1-\langle\sqrt{\bm x},\sqrt{\bm y}\rangle^2}}.
		\end{equation}
		An orthonormal basis can be constructed by taking a basis of $T_{\bm x}\MM$, e.g.,
		\begin{equation}
		\bigl\{(1,-1,0,0,\dots)^\tT,(1,1,-2,0,\dots)^\tT,\dots,(1,1,\dots,1,-d)^\tT\bigr\}\subset\R^{d+1}.
		\end{equation}
		and applying Gram-Schmidt orthonormalization process w.r.t.~the inner product $\langle\cdot,\cdot\rangle_x$.
		\vspace{0.2cm}
		
		\section{Simulation of Gaussian Noise Model by Said et al.~\texorpdfstring{\cite{SBBM15}}{[46]}}\label{sec:said}
		In the following we explain how to generate samples from the normal distribution  $\NN_{\text{Said}}(\bm \mu,\sigma^2 I_n)$ on ${\SPD}(r)$ ($n =\dim(\SPD(r)) = \frac{r(r+1)}{2}$), which was only sketched in~\cite{SBBM15}.
		To do so, we parametrize $\bm x\in \SPD(r)$ by its eigenvalues and eigenvectors (spectral decomposition), given as
		\begin{equation*}
		\bm x(\rho,\bm u) = \bm u \diag(\e^\rho){\bm u^\tT},
		\end{equation*}
		where $\bm u\in \operatorname{O}(r)$ is an orthogonal matrix and $\diag(\e^\rho)$ is the diagonal matrix with diagonal $(\e^{\rho_1},\ldots,\e^{\rho_r})$.\\
		As it is shown in~\cite{SBBM15}, in order to sample from $\NN_{\text{Said}}(\bm \mu,\sigma^2 I_n)$ it suffices to generate samples from $\NN_{\text{Said}}(I_r,\sigma^2 I_n)$. Indeed, if $\bm x\sim \NN_{\text{Said}}(I_r,\sigma^2 I_n)$, 
		then $\bm \mu^{\frac{1}{2}}\bm x{\bigl(\bm \mu^{\frac{1}{2}}\bigr)^\tT}\sim \NN_{\text{Said}}(\bm \mu,\sigma^2 I_n)$. 
		Further, 
		for sampling from $\NN_{\text{Said}}(I_r,\sigma^2 I_n)$ it is enough to sample from the uniform distribution on $\operatorname{O}(r)$ to generate $\bm u$  and from the distribution with density
		\begin{equation} \label{other_dist}
		p(\rho) \propto \exp\biggl\{-\frac{\rho_1^2+\ldots+\rho_r^2}{2\sigma^2}\biggr\}\prod_{i<j} \sinh\biggl(\frac{\abs{\rho_i-\rho_j}}{2}\biggr)
		\end{equation}
		to generate $\rho$. Once these are obtained, they can be plugged into the spectral decomposition $\bm x=\bm x(\rho,\bm u)$ to obtain $\bm x\sim \NN(I_r,\sigma^2 I_n)$.\\
		Sampling from the uniform distribution on $\operatorname{O}(r)$ can be done using a matrix $A$ whose components are i.i.d.\ standard normally distributed. 
		Computing the QR-decomposition $\bm a=\bm u\bm r$ with $\bm u$ orthogonal and $\bm r$ upper triangular, $\bm u$ is uniformly distributed on $\operatorname{O}(r)$, see, e.g.,~\cite{Chi12}.\\
		Sampling from the multivariate density $f$ in \eqref{other_dist} can be achieved using the \emph{acceptance-rejection} method, see, e.g.,~\cite{RC13}. 
		As dominating density we choose the density of the Euclidean Gaussian distribution $\NN(0,\tilde{\sigma}^2 I_r)$, where $\tilde{\sigma}^2 = \frac{2\sigma^2}{2-2(r-1)\sigma^2}$. 
		As we need $\tilde{\sigma}^2>0$, this allows only to sample in the case where $\sigma^2< \frac{1}{r-1}$. 
		However, numerical experiments indicate that this completely suffices to generate realistic Gaussian noise matrices which might arise in applications. 
		In order to use the acceptance-rejection method we have to show that $\frac{f(\rho)}{g(\rho)}\leq C$ for some constant $C>0$, 
		where $f$ is proportional to the density we want to sample from and $g$ is proportional to the chosen dominating density. In our situation, we choose
		$$
		g(\rho) = \exp\biggl\{- \frac{\rho_1^2+\ldots + \rho_r^2}{2\tilde{\sigma}^2}\biggr\}\\
		\propto \frac{1}{(2\pi\tilde{\sigma}^2)^{\frac{r}{2}}} \exp\biggl\{- \frac{\rho_1^2+\ldots + \rho_r^2}{2\tilde{\sigma}^2}\biggr\}.
		$$
		To show $\frac{f(\rho)}{g(\rho)}\leq C$, we first estimate
		\begin{align*}
		\prod_{i<j} \sinh\Bigl(\tfrac{\abs{\rho_i-\rho_j}}{2}\Bigr) & = \prod_{i\neq j}\biggl[ \sinh\Bigl(\tfrac{\abs{\rho_i-\rho_j}}{2}\Bigr)\biggr]^{\frac{1}{2}}
		= \prod_{i\neq j}\biggl[ \frac{1}{2}\Bigl(\e^{\frac{\lvert\rho_i-\rho_j\rvert}{2}}-\e^{\frac{-\lvert\rho_i-\rho_j\rvert}{2}}\Bigr)\biggr]^{\frac{1}{2}}\\
		& \leq  2^{-\frac{r(r-1)}{2}}\prod_{i\neq j}\e^{\frac{\lvert\rho_i-\rho_j\rvert}{4}}
		= 2^{-\frac{r(r-1)}{2}}\exp\biggl\{\frac{1}{4}\sum_{i\neq j}\lvert\rho_i-\rho_j\rvert\biggr\}\\
		& \leq 2^{-\frac{r(r-1)}{2}}\exp\biggl\{\frac{1}{4}\sum_{i\neq j}\rvert\rho_i\lvert+\lvert\rho_j\rvert\biggr\}
		= 2^{-\frac{r(r-1)}{2}}\exp\biggl\{\frac{r-1}{2}\sum_{i=1}^r\lvert\rho_i\rvert\biggr\}.
		\end{align*}
		Using 
		\begin{align*}
		\exp\biggl\{-\frac{1}{2\tilde{\sigma}^2}\sum_{i=1}^r \rho_i^2\biggr\}  & = 	\exp\biggl\{-\frac{2-2(r-1)\sigma^2}{4\sigma^2}\sum_{i=1}^r \rho_i^2\biggr\}\\
		&= \exp\biggl\{-\frac{1}{2\sigma^2}\sum_{i=1}^r \rho_i^2\biggr\}\exp\biggl\{\frac{2(r-1)}{4}\sum_{i=1}^r \rho^2_i\biggr\}
		\end{align*}
		we finally obtain
		\begin{align*}
		\frac{f(\rho)}{g(\rho)}& \leq \frac{2^{-\frac{r(r-1)}{2}}\exp\bigl\{-\frac{1}{2\sigma^2}\sum_{i=1}^r \rho_i^2\bigr\}\exp\bigl\{\frac{r-1}{2}
			\sum_{i=1}^r\lvert\rho_i\rvert\bigr\}}{\exp\bigl\{-\frac{1}{2\sigma^2}\sum_{i=1}^r \rho_i^2\bigr\}\exp\bigl\{\frac{r-1}{2}\sum_{i=1}^r \rho^2_i\bigr\}}\\
		& =2^{-\frac{r(r-1)}{2}} \exp\biggl\{\frac{r-1}{2}\sum_{i=1}^r \underbrace{\lvert\rho_i\rvert-\rho^2_i}_{\leq \frac{1}{4}} \biggr\} \\
		&\leq C, 
		\end{align*}
		where $C = \e^{\frac{r(r-1)}{8}} \, 2^{-\frac{r(r-1)}{2}}>0$.

		\paragraph{Acknowledgments} 
		We would like to thank R. Bergmann for fruitful discussions and for providing the test image in Figure \ref{fig:arts1}.
		Funding by the German Research Foundation (DFG) within the project STE 571/13-1 is gratefully acknowledged.
		
		\bibliographystyle{siam}
		\bibliography{DR-ref}
	\end{document}